\documentclass[a4paper,11pt]{article}
\usepackage{paper}
\usepackage{fontawesome5}
\usepackage{pdflscape}

\AtEndPreamble{
    \usepackage[capitalize]{cleveref}
    \crefname{section}{Sec.}{Secs.}
    \Crefname{section}{Section}{Sections}
    \crefname{table}{Tab.}{Tabs.}
    \Crefname{table}{Table}{Tables}
}

\pdfoutput=1

\hypersetup{
    pdftitle={Visual Spatial Learning: Single-Field Spatial Interpolation Using Convolutional Neural Networks},
    pdfauthor={Daniel Tinoco, Raquel Menezes, Carlos Baquero, Alexandra Silva}
}

\newcommand{\bib}{bibliography.bib}

\begin{document}

% Enter title:
\title{Visual Spatial Learning: Single-Field Spatial Interpolation Using Convolutional Neural Networks}

% Enter authors:
\author{Daniel Tinoco, Raquel Menezes, Carlos Baquero, Alexandra Silva
%
% Enter affiliations and acknowledgements:
\thanks{D. Tinoco (\hspace{1px}\faEnvelope[regular]\hspace{1px})\\
Centro de Matemática (CMAT), Universidade do Minho, Guimarães, Portugal\\
DEI-FEUP \& INESC TEC, Universidade do Porto, Porto, Portugal\\
E-mail: daniel.b.tinoco@inesctec.pt\\\\
R. Menezes\\
Centro de Matemática (CMAT), Universidade do Minho, Guimarães, Portugal\\
E-mail: rmenezes@math.uminho.pt\\\\
C. Baquero\\
DEI-FEUP \& INESC TEC, Universidade do Porto, Porto, Portugal\\
E-mail: cbm@fe.up.pt\\\\
A. Silva\\
Instituto Português do Mar e da Atmosfera, I. P. (IPMA, I. P.), Lisboa, Portugal\\
Centro de Ciências do Mar e do Ambiente (MARE), Évora, Portugal\\
E-mail: asilva@ipma.pt
}
}

% Enter date:
\date{May 2026}

\begin{titlepage}
\maketitle

% Enter abstract:
Predicting a complete spatially correlated field from sparse observations is a fundamental challenge in spatial statistics and environmental modelling.
Classical interpolation methods such as Kriging rely on Gaussian process assumptions and variography, which can limit their effectiveness in non-stationary settings and require substantial domain expertise.
In this work, we leverage an architecture based on convolutional neural networks (CNNs) for spatial interpolation that is trained and applied on a single partially observed field, without access to external data or prior fields.
The model is supervised directly on the observed locations and learns to predict values at unobserved points on the user defined grid.
Unlike Kriging, our method does not require explicit covariance modelling or variogram estimation, and it can flexibly capture local spatial patterns in a data-driven manner.
This work demonstrates the potential of CNNs for single-instance spatial interpolation under sparse supervision, offering a practical alternative to classical geostatistical methods, and extending the use of CNNs to a new problem domain.

\end{titlepage}

% Enter main text:
\section{Introduction}
\label{sec:intro}

The problem of interpolating a spatially correlated field from partial observations arises frequently in geostatistics, environmental modelling, and more broadly in spatial analysis.
In many settings, observations are collected at a sparse subset of spatial locations, and the objective is to estimate the value of the field at unmeasured locations \citep{geostatistics2007}.
A classical solution to this problem is Kriging \citep{K51}, which provides a linear estimator derived from assumptions of second-order stationarity and a specified or estimated spatial covariance structure.
Kriging remains widely used due to its statistical interpretability and its capacity to yield unbiased estimates with minimum variance under idealized assumptions.

Despite its strengths, Kriging exhibits several limitations when applied to complex spatial phenomena.
Its reliance on stationary or locally stationary covariance models, as well as its dependence on variogram fitting and model selection, often imposes substantial modelling effort and restricts its applicability in heterogeneous or non-stationary environments \citep{Cressie2006, GK12}.
Moreover, numerical methods such as the inversion of covariance matrices required in Kriging methods may become computationally prohibitive on large grids.

In this work, we propose an alternative approach to spatial interpolation based on Convolutional Neural Networks (CNNs) \citep{Fukushima1980, LBDHHHJ1989}.
Motivated by analogies to image inpainting \citep{BSCB2000}, we treat interpolation as a spatial field completion problem.
A distinguishing aspect of our methodology is that the proposed model is trained and applied solely on a single partially observed field, without recourse to external datasets or prior model training.
The method may be described as posterior-only in the sense that the model is optimized entirely in response to a single realization of the field and the available observations within it.

The proposed framework is free from assumptions of stationarity or Gaussianity.
It does not require the construction or fitting of variograms and is not dependent upon any external source of training data.
The proposed model is trained by minimizing a loss function defined only over the observed grid points, and is subsequently used to predict values at unobserved locations.
We evaluate the method on both synthetic and real-world spatial fields, comparing its performance to that of Ordinary Kriging.
Empirical results indicate that the proposed model approach achieves comparable or, when subjected to appropriate hyperparameter optimization, superior accuracy in settings characterized by spatial heterogeneity or local structure not easily accommodated by classical models.

The remainder of this paper introduces the relevant spatial statistics background (\Cref{sec:background}), presents the proposed interpolation framework based on CNNs (\Cref{sec:methodology}), and evaluates it on both synthetic (\Cref{sec:num_exp}) and real-world spatial fields (\Cref{sec:ipma}).

\section{Background}
\label{sec:background}

\subsection{Classical Spatial Interpolation}

Spatial interpolation concerns the prediction of a spatial process $\{ Z(s) : s \in \mathcal{S} \subset \mathbb{R}^2 \}$ at unobserved locations, from observations $\{Z(s_i)\}_{i=1}^n$ at known coordinates $\{s_i\}_{i=1}^n$.
A central assumption in much of classical geostatistics is that $Z(\cdot)$ can be modelled as a realization of a stochastic process with prescribed dependence properties, typically formulated through a covariance function or variogram \citep{geostatistics2007}.

Kriging constitutes the canonical method in this framework.
It is derived as the best linear unbiased predictor (BLUP) of $Z(s_0)$ at a new location $s_0$ given the observations \citep{stein1999}, where ``best'' refers to minimum prediction variance under the assumed second-order structure, and ``linear'' restricts the predictor to be an affine combination of the data:

\begin{equation}
\hat{Z}(s_0) = \sum_{i=1}^n \lambda_i(s_0) \, Z(s_i)
\end{equation}

The Kriging weights $\lambda_i(s_0)$ are determined by solving a system of equations derived from the unbiasedness constraint and the known (or typically estimated) covariance function $C(h) = \text{Cov}\big(Z(s), Z(s+h)\big)$.

In practice, the true covariance function is unknown and must be inferred from the available data, most often via variogram estimation.
The theoretical variogram is defined as:

\begin{equation}
\gamma(h) = \frac{1}{2} \mathbb{E} \left[ (Z(s) - Z(s+h))^2 \right]
\end{equation}
and is first computed empirically from the observations.
This empirical variogram is subsequently approximated by a permissible parametric model (e.g., exponential, Gaussian, Matérn, spherical), subject to the constraint of positive semi-definiteness of the implied covariance matrix.
The process of fitting such a model requires subjective decisions such as binning strategies, lag tolerances, and selection of model family, often guided by domain expertise and obtained during the variography step.

The standard Kriging framework presumes second-order stationarity (covariance depends only on separation vector $h$) or intrinsic stationarity (variogram depends only on $h$).
These assumptions can be overly restrictive in many real-world applications, particularly when the spatial field exhibits non-stationary characteristics, such as spatially varying variance, localized anisotropies, or abrupt transitions, that are not well captured by a single global variogram model.

Variants such as Universal Kriging \citep{MPS2019-ch5} extend the standard Kriging framework by incorporating a deterministic trend model into the mean structure of the spatial process.
This trend component, also referred to as the drift, represents the low-frequency (large-scale) variation in the data and corresponds to the mathematical expectation of the random function $Z(\cdot)$.
Drift models may be specified in either parametric or non-parametric form to capture systematic spatial gradients or smooth variations across the study domain \citep{ID1998}.
In the parametric case, the drift is modelled at a global scale, applying a single functional representation to the entire study area.
In contrast, non-parametric approaches allow for a local drift, enabling the trend to vary spatially in a flexible, location-dependent manner.
By explicitly modelling this deterministic component, Universal Kriging separates large-scale trends from high-frequency (small-scale) stochastic fluctuations, thereby improving the estimation of the spatial correlation structure and enhancing prediction accuracy under non-stationary conditions.

Unlike stationary formulations, non-stationary Kriging relaxes the assumption of translation-invariant covariance and instead models location-dependent structures that allow spatial dependence to vary across the domain \citep{R2016}.
This flexibility is crucial when the process exhibits heterogeneity such as shifts in variability, anisotropy, or in correlation range, enabling improved predictive performance in settings with abrupt transitions, regional anisotropy, or varying smoothness that stationary models cannot adequately capture, while requiring careful parameterization to avoid overfitting \citep{FSLR2015}.

\subsection{Neural Networks for Spatial modelling}
\label{ssec:nn}

Neural networks \citep{Schmidhuber2015} have emerged as a flexible class of function approximators \citep{HSW1989, SZ2006} capable of representing complex, non-linear relationships in high-dimensional data.
When applied to spatial modelling, they offer an alternative to explicitly specifying correlation structures, instead learning spatial dependencies directly from data.

Convolutional neural networks (CNNs) \citep{Fukushima1980, LBDHHHJ1989}, in particular, have proven effective for tasks where the underlying data are naturally represented on regular grids, such as images.
In spatial settings, by translating the space to a regular grid, this architecture implicitly encodes the assumption that local patterns repeat across the domain, while allowing for hierarchical feature extraction as deeper layers integrate information over progressively larger receptive fields.

From a statistical perspective, a CNN layer can be viewed as a non-linear transformation of the field that captures localized correlation structures without requiring the specification of a parametric covariance function.

By stacking multiple convolutional layers interleaved with non-linear activation functions, the model can capture both short-range and longer-range dependencies, approximating complex, possibly non-stationary spatial processes. In this respect, CNNs bear conceptual similarities to kernel methods with learned, spatially localized kernels, but with the added capacity for non-linearity and data-driven adaptation.

Architectures based on CNNs are widely used in the field of computer vision, where they have demonstrated substantial success in tasks such as image inpainting \citep{BSCB2000} and completion \citep{SSI2017}.
In these applications, large missing regions of an image are reconstructed by leveraging contextual information from the surrounding pixel values \citep{PKDDE2016}.
These problems are conceptually analogous to spatial interpolation in that they require estimating unobserved field values based on partial spatial context.

However, the majority of existing image reconstruction methods employing CNNs are trained in a fully or semi-supervised setting, using large datasets containing numerous complete examples \cite{GSNSPSCIBHCW2023} of the underlying data.
This external data enables the network to learn a prior over plausible field configurations, which it then applies to new incomplete instances.
While effective, such an approach is incompatible with the more typical case in spatial statistics namely the single-instance interpolation problem, where only one partially observed field is available and no additional training data exists.

Advances in deep image prior \citep{UVL2018} methodologies suggest that the structure of a convolutional network itself can serve as an implicit prior, enabling useful reconstruction from a single instance by optimizing the network parameters directly on the available observations.
This idea motivates the present work, where a convolutional network is adapted to learn from sparse supervision within a single spatial field, bypassing the need for explicit covariance modelling or large training datasets while retaining the ability to capture complex local spatial patterns.

\section{Methodology}
\label{sec:methodology}
Let $Z : \mathcal{S} \rightarrow \mathbb{R}$ be a real-valued spatial field defined over a bounded spatial domain $\mathcal{S} \subset \mathbb{R}^2$.
In practice, the domain is discretized into a regular grid of size $H \times W$, yielding the discrete field $Z(i, j)$ for $(i, j) \in \mathcal{A}$ with $\mathcal{A} = \{1, \ldots, H\} \times \{1, \ldots, W\}$.
The observations consist of a subset of these grid locations, denoted $\mathcal{O} \subset \mathcal{A}$, together with the corresponding values $Z(i, j)$ for $(i, j) \in \mathcal{O}$.
The interpolation problem is to estimate the values of $Z(i,j)$ at locations $\mathcal{U}$ not contained in $\mathcal{O}$, using only the observed data.

Formally, the interpolation problem can be expressed as learning a mapping $\hat{f}: \mathcal{S} \to \mathbb{R}$ such that $\hat{f}(s_i) \approx y_i, \enspace \forall s_i \in \mathcal{O}$, and $\hat{f}(\cdot)$ generalizes to yield accurate predictions $\hat{f}(u)$ for all $u \in \mathcal{U}$.
In the classical geostatistical setting, $\hat{f}(\cdot)$ is derived analytically from an estimated covariance structure, whereas in the present work, it will be represented implicitly by a model based on CNNs whose parameters are optimized with respect to the available observations.

Unlike traditional learning settings, no collection of spatial fields from the same process is available for training, nor is any prior distribution over such fields assumed.
The task is to estimate a continuous extension of the partial field solely from the information present in $\mathcal{O}$.
This distinguishes our setting from typical supervised or self-supervised learning paradigms, in that both the training and inference stages are restricted to a single realization of the process.

\subsection{Spatial Interpolation Model}
We adopt a convolutional neural network architecture to perform single-field spatial interpolation on a discretized domain.
The input to the network is a tensor $\mathbf{X}^{H \times W \times 2}$, where the first channel contains the observed values and the second channel contains a binary mask $\mathbf{M} \in \{ {0, 1} \} ^{H \times W}$ indicating which values are observed.

Given that only a subset of the field values is observed, it is necessary to ensure that the loss function reflects this partial observability.
To that end, the binary mask $\mathbf{M}$ is introduced in the loss function $\mathcal{L}(\cdot)$ calculation to indicate the locations of observed entries, thereby constraining the model to learn exclusively from available data.
Accordingly, the network is trained by minimizing the mean squared error restricted to the observed locations:

\begin{equation}
\mathcal{L}(\theta) = \frac{1}{\sum_{i,j} \mathbf{M}_{i,j}} \sum_{i,j} \mathbf{M}_{i,j} \left( f_\theta(\mathbf{X})_{i,j} - Z_{i,j} \right)^2
\label{eq:mask_mse}
\end{equation}
where $f_\theta$ denotes the model parameterized by $\theta$.
The minimization is performed solely on the observed entries $\mathcal{O}$, with the objective of learning local and global spatial dependencies from the observed subset alone.
Once trained, the network produces a full field prediction $\hat{Z}(i,j) = f_\theta(\mathbf{X})_{i,j}$ for all $(i, j) \in \mathcal{A}$.
This loss function serves as a baseline, and alternative formulations will be explored in subsequent experiments.

The architecture employs an encoder-decoder design with convolutional layers and skip connections, similar in structure to U-Net \citep{RFB2015} like architectures and models used in image reconstruction tasks.
The convolutional structure imposes a spatial prior of locality and translation invariance, which is well-suited to structured grid data.
No explicit spatial covariance model is imposed or inferred; instead, dependencies are learned directly from the observed data through optimization.

This approach is purely data-driven and posterior-specific.
The model does not generalize across fields, nor does it rely on previously seen spatial patterns.
This property is of particular relevance in our case scenarios where only a single field is available or where field-to-field variability is too great to allow for meaningful generalization.

\subsection{Architecture}
\begin{figure}[t]
\centering
\makebox[\textwidth][c]{\includegraphics[width=1.2\textwidth]{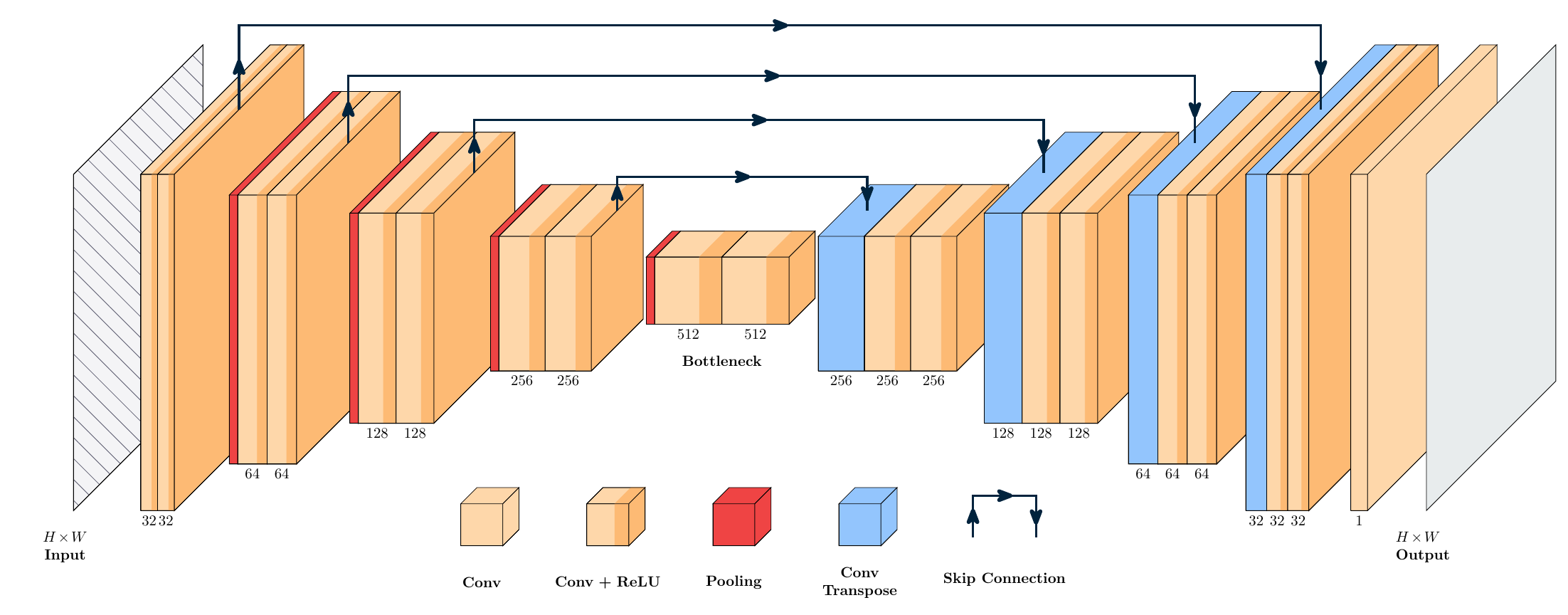}}
\caption{Diagram of an architecture, similar in structure to U-Net. The model exhibits the characteristic encoder-decoder structure with symmetric contracting and expanding paths connected through a set of skip connections. The left (encoding) arm applies successive convolutional and down-sampling operations to extract increasingly abstract feature representations, while the right (decoding) arm performs up-sampling and refinement to recover spatial information.}
\label{fig:unet_conv}
\end{figure}

At a high level, the spatial interpolation model (\Cref{fig:unet_conv}) is instantiated as an encoder-decoder \citep{SVL2014} convolutional architecture and operates on a two-dimensional grid representation of the spatial field, where each grid cell corresponds to a spatial location $s \in \mathcal{S} \subset \mathbb{R}^2$.
The input is a single-channel array representing observed values, with unobserved entries masked (binary mask $\mathbf{M}$).
The output has the same spatial dimensions and channel configuration, representing the predicted or reconstructed field values at all grid points.

The fundamental computational unit of the network is a double convolution block, composed of two consecutive convolutional layers with hyper-parametrized kernel size, each followed by a rectified linear unit (ReLU) activation.
These blocks serve as modular components across both encoder and decoder paths.
Unlike many conventional CNN architectures, normalization layers (e.g., batch normalization) are omitted, emphasizing direct learning from raw convolutional responses and preserving absolute value relationships that are often crucial in geostatistical data.

The encoder consists of a sequence of convolutional blocks interleaved with spatial downsampling operations, enabling the network to capture dependencies at increasing spatial scales.
As depth increases, the receptive field expands from local neighbourhoods to broader spatial context, allowing the model to integrate both fine-scale and large-scale spatial structure.

The decoder mirrors this structure through a sequence of upsampling operations followed by convolutional refinement.
Skip connections link corresponding encoder and decoder stages, allowing high-resolution spatial information to be directly propagated to later layers.
This design mitigates the loss of fine-scale details induced by downsampling and is particularly important for spatial interpolation, where localized structures and sharp transitions may be sparsely observed.

At the network's output, a convolution layer projects the final feature maps back to a single channel, restoring the original field dimensionality.
Depending on the value range of the spatial domain $\mathcal{S}$, the application of an output activation function may or may not be appropriate.
In the present setting, the interpolation task constitutes a continuous-valued regression problem, for which the output range is left unconstrained to permit unrestricted prediction of field values.

\subsection{Partial Convolutions}
\begin{figure}[t]
\centering
\includegraphics[width=\textwidth]{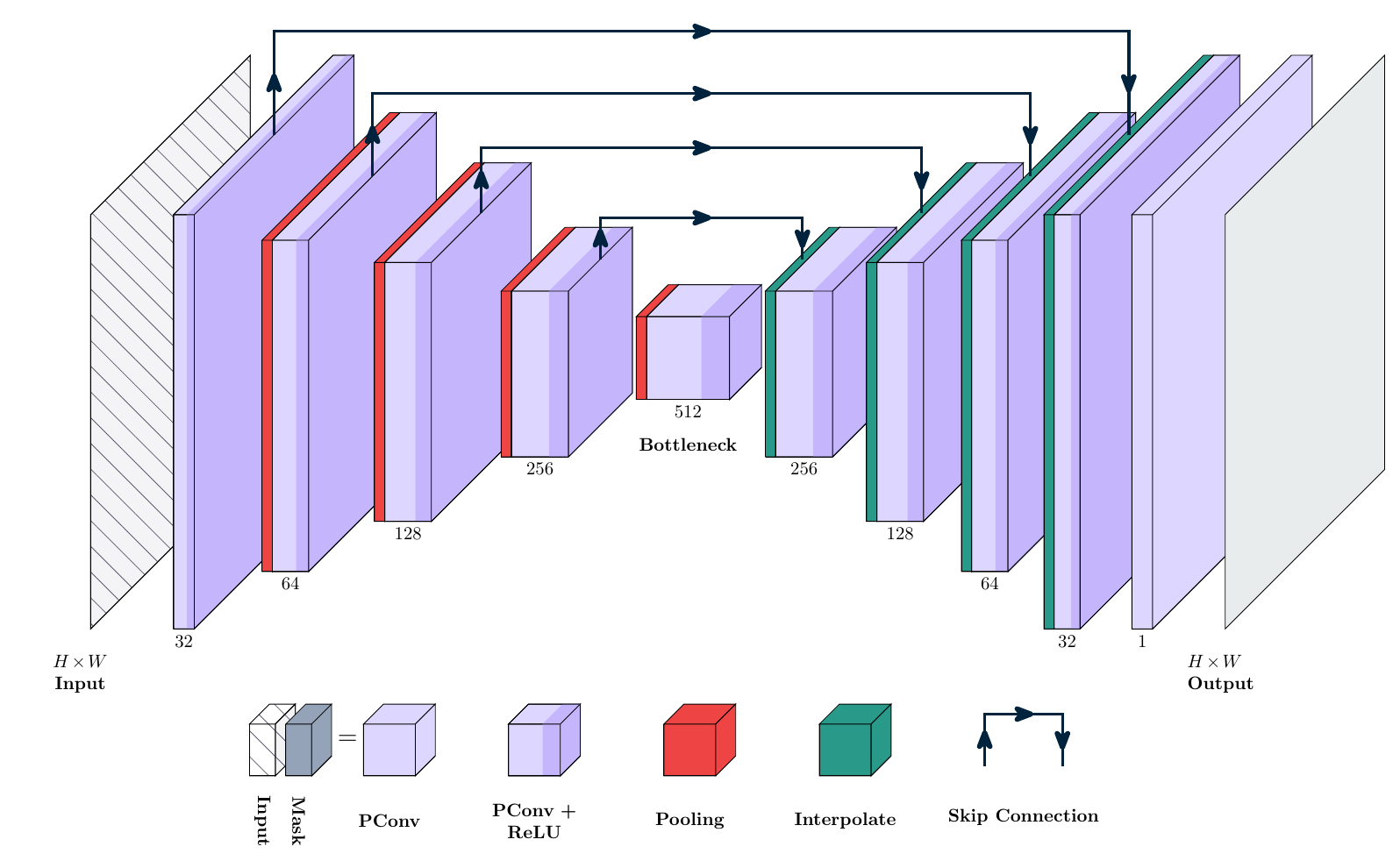}
\caption{Diagram of an encoder-decoder architecture, similar in structure to U-Net, in which all convolutional layers are replaced by partial convolution layers.}
\label{fig:unet_pconv}
\end{figure}

Sparse or irregularly sampled spatial data present a fundamental challenge for convolutional architectures, which conventionally assume complete and uniformly sampled inputs. In the context of spatial interpolation, large portions of the field may be unobserved, rendering standard convolutional operations ill-posed, as they aggregate information indiscriminately over both observed and missing regions. To address this issue, the present model replaces conventional convolutions with partial convolution layers, an operation explicitly designed to account for spatial incompleteness during feature extraction \citep{pconv2018}.

In the conventional convolution, a filter $\mathbf{W}$ is applied uniformly over all spatial locations, implicitly assuming that every element of the local receptive field contributes meaningfully to the output.
When inputs contain missing or corrupted values, the standard operation becomes agnostic to the distinction between valid and invalid entries, which can induce systematic bias and propagate artifacts.

Formally, for a convolution filter with weights $\mathbf{W}$ and bias $b$, and writing $\mathbf{X}$ for the local feature patch, the standard convolution at a given spatial location is expressed as:

\begin{equation}
\mathbf{Y}_{\text{Conv}} = \mathbf{W}^T \mathbf{X} + b
\end{equation}
where the inner product implicitly assumes that every element of $\mathbf{X}$ contributes equally to the computation, clearly illustrating its reliance on the full receptive field irrespective of data quality or completeness.

In contrast, the partial convolution operator \citep{HDK2017} introduces a binary mask $\mathbf{M}$, with $\mathbf{M}_{ij}=1$ indicating a valid (unmasked) entry and $\mathbf{M}_{ij}=0$ indicating a missing one.
The partial convolution at each spatial location is defined as:

\begin{equation}
\mathbf{Y}_{\text{PConv}} =
\begin{cases}
\mathbf{W}^T(\mathbf{X} \odot \mathbf{M}) \dfrac{\sum_{ij}\mathbf{1}_{ij}}{\sum_{ij}\mathbf{M}_{ij}} + b, & \text{if } \sum_{ij}\mathbf{M}_{ij} > 0 \\
0, & \text{otherwise}
\end{cases}
\label{eq:pconv}
\end{equation}
where $\odot$ denotes element-wise multiplication and $\mathbf{1}$ is an all-ones tensor of the same shape as $\mathbf{M}$.
The renormalization factor $\sum_{ij}\mathbf{1}_{ij}/\sum_{ij}\mathbf{M}_{ij}$ compensates for the variable number of valid inputs.
In effect, the operator computes a rescaled convolution over only the available data, maintaining consistent output magnitudes independent of the local pattern of absent entries.

Crucially, the partial convolution layer also propagates information about the evolving support of valid inputs through a deterministic mask update:

\begin{equation}
\mathbf{M'}_{ij} =
\begin{cases}
1, & \text{if } \sum_{ij}\mathbf{M}_{ij} > 0 \\
0, & \text{otherwise}
\end{cases}
\label{eq:maskupdate}
\end{equation}

Any location whose partial convolution involves at least one valid input becomes valid in subsequent layers.
Through repeated applications of this update rule, the region of validity expands until eventually all positions become valid, provided the initial mask contains at least one unmasked pixel in each connected component of interest.

Building on this operation, the overall architecture integrates partial convolutional blocks throughout both encoder and decoder paths (\Cref{fig:unet_pconv}), ensuring consistent handling of sparsity at all resolutions.
In the decoder, validity masks are upsampled and combined with skip-connected encoder masks, preserving coherence between feature propagation and spatial support.
This design prevents the uncontrolled diffusion of invalid information into unobserved regions and stabilizes training under severe observation gaps.
Predictions are restricted to previously unobserved regions, while original observations are preserved through the masking mechanism.
This yields an estimated field $\hat{Z}(s)$ that is consistent with known data and spatially coherent across missing regions.

Through the explicit incorporation of visibility masks into every convolutional operation, this architecture maintains a principled treatment of data sparsity.
By doing so, it allows the network to learn meaningful spatial representations even under severe observation gaps, which is a property of particular relevance in geostatistical and environmental applications where sampling is most often costly or spatially uneven.

\subsection{Adaptive Mixed Smooth Loss}
While the loss function presented at \cref{eq:mask_mse}, henceforth referred as $\mathcal{L}_{\text{masked}}$, provides a principled way to constrain supervision to observed locations, it does not incorporate information about the spatial structure of unobserved regions, nor does it explicitly encourage spatial coherence in the reconstructed field.
To address these limitations, we introduce an improved loss function that combines and augments the masked reconstruction term with a smoothness regularization component, adaptively weighted according to both the spatial proximity to known observations and the progression of training.
This formulation enables the network to balance fidelity to observed data with coherent interpolation in unobserved regions, ensuring more stable and physically plausible reconstructions.

\subsubsection{Smoothness Regularization}
To encourage spatial smoothness and control local variability, the loss incorporates two complementary operators, a Gaussian filter $G(\cdot)$ \citep{GW2017} and a Laplacian filter $L(\cdot)$ \citep{JRAA2000}.
The Gaussian operator acts as a local averaging filter that penalizes high-frequency fluctuations, while the Laplacian operator measures local curvature, thereby emphasizing edge-like structures and discontinuities.

Starting by implementing a two-dimensional Gaussian smoothing operation, the procedure first constructs a discrete one-dimensional Gaussian kernel of odd length $k$.
Writing $r = (k-1)/2$, the kernel is defined on the index set $\{-r, \ldots, r\}$ by:

\begin{equation}
g(h) = \exp\!\left( -\left(\frac{h}{2\sigma}\right)^{2} \right),
\qquad h \in \{-r,\ldots,r\}
\end{equation}
and subsequently normalized so that:

\begin{equation}
\tilde{g}(h) = \frac{g(h)}{\sum_{s=-r}^{r} g(s)}
\end{equation}

A separable \footnote{A 2-D function $G(x,y)$ is said to be separable if it can be written as the product of two 1-D functions, $g_1(x)$ and $g_2(y)$. Meaning $G(x,y) = g_1(x) g_2(y)$. In the same sense, a separable kernel is a matrix that can be expressed as the outer product of two vectors.} two-dimensional Gaussian filter is then obtained by forming the outer product:
\begin{equation}
G(u,v) = \tilde{g}(u)\,\tilde{g}(v),
\qquad u,v \in \{-r,\ldots,r\},
\end{equation}
yielding a $k \times k$ kernel.

As such, the Gaussian operator performs a weighted local averaging over a kernel, which is computed via the standard discrete convolution with symmetric zero-padding of radius $r$, where the Gaussian-smoothed field is defined as:

\begin{equation}
\mathbf{G}(f_\theta(\mathbf{X})_{ij}) = \sum_{u=-r}^{r} \sum_{v=-r}^{r} G(u,v) \, f_\theta(\mathbf{X})_{i+u, j+v}
\end{equation}
which penalizes high-frequency fluctuations and promotes gradual transitions.

Continuing with the Laplacian operator which captures local curvature via a discrete convolution with kernel $\mathbf{K}_L$:
\begin{equation}
\mathbf{K}_L =
\begin{bmatrix}
0 & 1 & 0 \\
1 & -4 & 1 \\
0 & 1 & 0
\end{bmatrix}
\end{equation}
emphasizing regions of sharp variation.
The Laplacian-filtered field is therefore given by:

\begin{equation}
\mathbf{L}(f_\theta(\mathbf{X})_{ij}) = \sum_{u=-1}^{1} \sum_{v=-1}^{1} \mathbf{K}_L(u,v) \, f_\theta(\mathbf{X})_{i+u, j+v}
\end{equation}

This operator emphasizes second-order spatial structure, by producing strong responses at locations of rapid intensity variation while suppressing regions of approximately constant intensity.

The corresponding smoothness terms are expressed as:

\begin{equation}
\mathcal{L}_G = \frac{1}{N} \sum_{i,j} (f_\theta(\mathbf{X})_{ij} - \mathbf{G}(f_\theta(\mathbf{X}))_{ij})^2,
\qquad
\mathcal{L}_L = \frac{1}{N} \sum_{i,j} (\mathbf{L}(f_\theta(\mathbf{X}))_{ij})^2
\end{equation}
where $N$ is the total number of grid cells.
These smoothness terms are then combined into a single smoothness penalty:

\begin{equation}
\mathcal{L}_{\text{smooth}} = \mathcal{L}_G + \lambda_L \, \mathcal{L}_L
\label{eq:l_smooth}
\end{equation}
where $\lambda_L > 0$ balances the relative contribution of the Laplacian term.

\subsubsection{Spatially Adaptive Regularization}
Since interpolation uncertainty increases with distance from observed data, the smoothness regularization is modulated spatially.
Let $\mathbf{M}$ denote the observation mask and $\mathbf{U} = \mathbf{1} - \mathbf{M}$ the unobserved region.
An approximate distance weighting $\mathbf{D}$ is obtained by locally averaging $\mathbf{U}$ with a blur kernel of size $k$:
\begin{equation}
\mathbf{B}_{i,j} = \frac{1}{k^2} \sum_{u=-k/2}^{k/2} \sum_{v=-k/2}^{k/2} \mathbf{U}_{i+u, j+v}
\end{equation}
followed by normalization to $[0,1]$:
\begin{equation}
\mathbf{D}_{i,j} =
\begin{cases}
0, & \mathbf{B}_{i,j} < 0 \\
\mathbf{B}_{i,j}, & 0 \le \mathbf{B}_{i,j} \le 1 \\
1, & \mathbf{B}_{i,j} > 1
\end{cases}
\end{equation}
where higher values of $\mathbf{D}_{i,j}$ correspond to locations farther from known data.
The spatial average $\mathbf{\overline{D}} = \frac{1}{N} \sum_{i,j} \mathbf{D}_{i,j}$ serves as a global scaling factor for the regularization term.

\subsubsection{Adaptive Weight Decay}
To prevent over-smoothing in later stages of training, the smoothness weight is further modulated over time through an adaptive decay schedule, defined as:

\begin{equation}
\omega_s = \omega_0 \left( 1 - \min\!\left(1, \frac{t}{T}\right) \right)
\end{equation}
where $\omega_0 $ is the initial smoothness weight, $t$ the current training iteration and $T$ the total number of iterations.
This schedule linearly reduces the influence of the smoothness term as the model converges.

\subsubsection{Final Adaptive Mixed Smooth Loss}
Combining the above components yields the total training objective:

\begin{equation}
\mathcal{L}_{\text{total}} = \mathcal{L}_{\text{masked}}
+ \omega_s \, \mathbf{\overline{D}} \, \mathcal{L}_{\text{smooth}}
\end{equation}

This composite loss enforces consistency with observed data through $\mathcal{L}_{\text{masked}}$ (\ref{eq:mask_mse}), while encouraging spatial regularity via $\mathcal{L}_{\text{smooth}}$ (\ref{eq:l_smooth}).
The adaptive weighting mechanism ensures that the degree of regularization diminishes both with proximity to known observations and with training progression.

Overall, this formulation provides a principled balance between data fidelity and spatial smoothness, enabling the network to reconstruct unobserved regions in a manner consistent with both local continuity and global field structure.

\section{Numerical Experiments}
\label{sec:num_exp}

\begin{figure}[t]
\centering
\includegraphics[width=\textwidth]{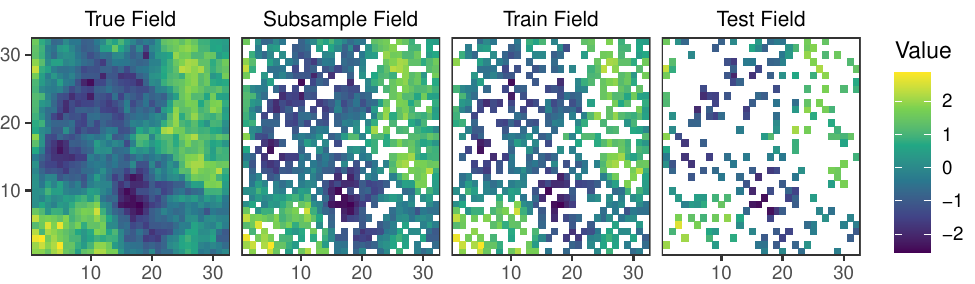}
\caption{Tile plot of a spatial correlated field discretized in a grid of size $32 \times 32$, a subsample of 70\% as available observations and a train/test split 70/30 percent of the subsample (49\%/21\%).}
\label{fig:true_subsample_train_test_048_plot}
\end{figure}

We assess the empirical performance of the proposed CNN interpolation method through a series of controlled numerical experiments and compare it against the classical Ordinary Kriging (OK) predictor.
All experiments, in this section, are conducted on synthetic spatial fields defined on a regular grid of size $32 \times 32$, where a subsample is taken from the true field in order to better simulate a real scenario, which is further split into the train and test fields, as illustrated in \Cref{fig:true_subsample_train_test_048_plot}.

Prediction accuracy is quantified using the root mean squared error (RMSE):
\begin{equation}
\text{RMSE} = \sqrt{{\frac {1}{N}}\sum _{i=1}^{N}\left(y_{i}-{\hat {y_{i}}}\right)^{2}}
\label{eq:rmse}
\end{equation}
and also the RMSE of the Moran's~\textit{I} (MI-RMSE), computed by comparing the global spatial autocorrelation of predicted and reference fields:
\begin{equation}
I = {\frac {N}{W}}{\frac {\sum _{i=1}^{N}\sum _{j=1}^{N}w_{ij}(x_{i}-{\bar {x}})(x_{j}-{\bar {x}})}{\sum _{i=1}^{N}(x_{i}-{\bar {x}})^{2}}}
\label{eq:morans_i}
\end{equation}
These metrics are used to assess complementary aspects of interpolation performance.
While RMSE provides a standard measure of pointwise predictive accuracy, it does not, by itself, indicate whether the predicted field adequately reproduces the spatial dependence structure of the underlying process.
To address this limitation, we additionally consider Moran's~\textit{I}, a global measure of spatial autocorrelation that evaluates the degree to which nearby locations exhibit similar values.
By comparing Moran's~\textit{I} computed on the predicted field to that of the reference field, we assess whether the interpolation method preserves the spatial coherence and correlation patterns inherent in the data.
The joint use of RMSE and the MI-RMSE therefore allows us to evaluate both local numerical accuracy and global spatial structure, which is particularly important in the context of spatial interpolation where reproducing dependence patterns is as critical as minimizing pointwise prediction error.

The RMSE is evaluated on the unobserved locations only, while the MI-RMSE is evaluated on the full interpolated field.
For Kriging, covariance model parameters are assumed known in order to isolate the effect of model assumptions rather than variogram estimation uncertainty.

A summary of the classical interpolation methods considered in this study, together with the benchmark CNN interpolation approach and the proposed model, is presented in \Cref{tab:models}.

\begin{table}[t]
\centering
\caption{Reference table of spatial interpolation model names considered in this study, together with brief descriptions of their underlying assumptions and methodological characteristics.}
\label{tab:models}
\begin{tabular}{lp{.7\textwidth}}
\toprule
\textbf{Model} & \textbf{Description}  \\
\midrule
\textbf{Kriging} & Ordinary Kriging. \\
\midrule
\textbf{Kriging NS} & Ordinary Kriging with non-stationary covariance model. \\
\midrule
\textbf{ML Base} & U-Net--like convolutional architecture, trained using a masked mean squared error loss. \\
\midrule
\textbf{ML VSL} & U-Net--like encoder-decoder architecture with partial convolution layers, as illustrated in \Cref{fig:unet_pconv}, trained using an adaptive mixed smooth loss function. \\
\bottomrule
\end{tabular}
\end{table}

\subsection{Stationary and Isotropic Gaussian Random Fields}
\label{subsec:grf_stationary}
We begin with the benchmark setting most favourable to classical geostatistics, namely synthetic spatial fields generated from stationary and isotropic Gaussian random fields \citep{Cressie1993, geostatistics2007}.
Spatial fields are generated from a zero-mean Gaussian process with a stationary isotropic Exponential covariance function (\ref{eq:cov_exp}) and with observations being sampled from a normal distribution, $\mathcal {N}(0,1)$, at random across the domain.

\begin{equation}
\mathcal{C}_{\text{Exponential}}(h;\sigma^2,\phi) = \sigma^2 \exp\left(-\frac{h}{\phi}\right), \quad \phi > 0
\label{eq:cov_exp}
\end{equation}
with $\sigma^2=1$ and varying range $\phi$.

Under these assumptions, Ordinary Kriging (OK) is theoretically optimal, being the best linear unbiased prediction (BLUP) at unobserved locations.
As expected, OK achieves slightly lower prediction error than the proposed visual spatial learning CNN method (\Cref{tab:gaussian-v2_10_phi}; \Cref{ap:subsec:grf_stationary} for the complete set of results).
The difference, however, is small, and both approaches produce visually and quantitatively similar reconstructions. The proposed CNN method successfully recovers the global smoothness and correlation structure of the field despite being trained on a single partially observed realization and without access to the true covariance model.
These results indicate that, even in settings that strictly adhere to Kriging assumptions, the proposed approach remains competitive, with only a marginal loss in accuracy relative to the optimal linear predictor.

\begin{table}[tb]
\centering
\caption{\textit{Exponential} --- Comparison of model performance across varying percentages of known points (20\%, 50\% and 80\%). Values represent the mean ($\pm$ standard deviation) computed over 100 independent runs. The spatial interpolation was conducted using an \textit{Exponential} covariogram with a range parameter of 10\% of the grid size. The underlying data follows a \textit{Gaussian} distribution.}
\label{tab:gaussian-v2_10_phi}
\begin{tabular}{llccc}
\toprule
\textbf{Model} & \textbf{Metric}  & \textbf{20\%} & \textbf{50\%} & \textbf{80\%}  \\
\midrule
\multirow{2}{*}{Kriging} & RMSE &  \textbf{0.658 $\pm$ 0.06} &  \textbf{0.585 $\pm$ 0.04} &  \textbf{0.544 $\pm$ 0.02} \\
 & MI-RMSE &  \textbf{0.193 $\pm$ 0.03} &  0.137 $\pm$ 0.02 &  0.088 $\pm$ 0.01 \\
\midrule
\multirow{2}{*}{ML Base} & RMSE &  0.961 $\pm$ 0.11 &  0.971 $\pm$ 0.09 &  0.949 $\pm$ 0.08 \\
 & MI-RMSE &  0.484 $\pm$ 0.08 &  0.405 $\pm$ 0.04 &  0.265 $\pm$ 0.03 \\
\midrule
\multirow{2}{*}{ML VSL} & RMSE &  0.664 $\pm$ 0.07 &  0.592 $\pm$ 0.04 &  0.550 $\pm$ 0.03 \\
 & MI-RMSE &  0.197 $\pm$ 0.03 &  \textbf{0.128 $\pm$ 0.02} &  \textbf{0.075 $\pm$ 0.01} \\
\bottomrule
\end{tabular}
\end{table}

To better approximate conditions encountered in applied spatial analysis, we extend the previous experiment by introducing a nugget effect into the data-generating process.
In geostatistical modelling, the nugget component represents microscale variability and measurement error that cannot be explained by spatial correlation alone, and is therefore common in real-world datasets appearing in many sources, such as environmental monitoring, and experimental measurements \citep{Cressie1993}.

The resulting covariance structure becomes:
\begin{equation}
\mathcal{C}_{\text{Exponential+Nugget}}(h;\sigma^2,\phi) =
\begin{cases}
\sigma^2 + \tau^2, & h = 0 \\
\sigma^2 \exp\left(-\dfrac{h}{\phi}\right), & h > 0
\end{cases}
\end{equation}
with $\sigma^2=1$, $\tau^2=0.3$ denotes the nugget variance, and $\phi > 0$ values are varied in order to assess the robustness of each interpolation method under increasing levels of spatial correlation.

The inclusion of a nugget effect degrades the effective signal-to-noise ratio and weakens short-range spatial dependence, thereby challenging interpolation methods that rely heavily on smoothness and local correlation.
In this setting, Ordinary Kriging remains the BLUP when the covariance parameters, including the nugget, are correctly specified.
However, in practice, estimation of the nugget parameter from sparse and irregularly spaced observations is notoriously difficult, often leading to biased variogram fits and suboptimal predictions.
Under this setting, the proposed visual spatial learning approach achieves lower prediction error than Ordinary Kriging when the available percentage points is higher than 20\% (\Cref{tab:gaussian-nugget-03-v2_10_phi}).

\begin{table}[tb]
\centering
\caption{\textit{Exponential$\,+$Nugget} --- Comparison of model performance across varying percentages of known points (20\%, 50\% and 80\%). Values represent the mean ($\pm$ standard deviation) computed over 100 independent runs. The spatial interpolation was conducted using an \textit{Exponential}, with \textit{Nugget} component, covariogram with a range parameter of 10\% of the grid size. The underlying data follows a \textit{Gaussian} distribution.}
\label{tab:gaussian-nugget-03-v2_10_phi}
\begin{tabular}{llccc}
\toprule
\textbf{Model} & \textbf{Metric}  & \textbf{20\%} & \textbf{50\%} & \textbf{80\%}  \\
\midrule
\multirow{2}{*}{Kriging} & RMSE &  \textbf{0.905 $\pm$ 0.09} &  0.854 $\pm$ 0.06 &  0.828 $\pm$ 0.03 \\
 & MI-RMSE &  \textbf{0.285 $\pm$ 0.08} &  \textbf{0.225 $\pm$ 0.02} &  0.148 $\pm$ 0.02 \\
\midrule
\multirow{2}{*}{ML Base} & RMSE &  1.119 $\pm$ 0.12 &  1.131 $\pm$ 0.09 &  1.124 $\pm$ 0.08 \\
 & MI-RMSE &  0.375 $\pm$ 0.08 &  0.321 $\pm$ 0.04 &  0.213 $\pm$ 0.03 \\
\midrule
\multirow{2}{*}{ML VSL} & RMSE &  0.909 $\pm$ 0.09 &  \textbf{0.851 $\pm$ 0.06} &  \textbf{0.825 $\pm$ 0.03} \\
 & MI-RMSE &  0.342 $\pm$ 0.04 &  0.226 $\pm$ 0.03 &  \textbf{0.134 $\pm$ 0.02} \\
\bottomrule
\end{tabular}
\end{table}

This performance difference can be explained by the distinct mechanisms through which the two approaches handle unstructured noise.
Even when the nugget parameter is fixed and correctly specified, Ordinary Kriging remains a linear predictor of the noisy observations and therefore propagates high-frequency variability into the predictions, particularly in the vicinity of observed locations.

In contrast, the proposed visual spatial learning method does not explicitly model the nugget as a separate variance component.
Instead, the partial convolutional architecture implicitly regularizes the field through spatial aggregation and non-linear feature extraction, attenuating uncorrelated noise while preserving the dominant spatial patterns of the latent process.
As a result, the network is better aligned with the objective of reconstructing the underlying smooth field rather than interpolating the noisy observations themselves.

These findings indicate that, even in a setting that remains close to the classical geostatistical assumptions, the introduction of a nugget component can favour data-driven interpolation strategies.
The proposed approach demonstrates increased robustness to noise contamination, highlighting its potential advantages in practical scenarios where microscale variability and measurement error are unavoidable and difficult to model accurately.

\subsection{Non-stationary and Anisotropic Fields}
\label{subsec:non-stationary}
We next consider spatial fields that depart from the assumptions of stationarity and isotropy.
In particular, we generate Gaussian random fields governed by a non-stationary Matérn covariance function \citep{PS2006, KN2012, BF2025}, with smoothness parameter $\nu = 1/2$ and marginal variance $\sigma^2 = 1$, while allowing the mean and range parameter $\phi$ to vary over space.
In addition, both anisotropy and directional tilt are modelled as smoothly varying functions of location, thereby inducing spatially heterogeneous dependence structures.
As a result, the simulated fields exhibit locally varying correlation lengths and direction-dependent behaviour across the spatial domain.

We use the covariance form:
\begin{equation}
\mathcal{C}_{\text{NS}}(s_i,s_j; \boldsymbol\psi_i,\boldsymbol\psi_j)= \sigma_i \sigma_j |\boldsymbol\Sigma_i|^{\frac{1}{4}}|\boldsymbol\Sigma_j|^{\frac{1}{4}} \Biggl|\frac{\boldsymbol\Sigma_i + \boldsymbol\Sigma_j}{2}\Biggr|^{-\frac{1}{2}} \mathcal{M}_{\sqrt{\nu_i + \nu_j}}\Bigl(\sqrt{Q_{ij}}\Bigr)
\label{eq:cov_ns}
\end{equation}
where $\boldsymbol\psi = (\sigma^2, \phi, \nu)$, $\sigma_{\ell} = \sigma(\boldsymbol{s}_{\ell})$ is a standard deviation process, $\nu_{\ell} = \nu(\boldsymbol{s}_{\ell})$ is a smoothness process, $\boldsymbol\Sigma_{\ell} = \boldsymbol\Sigma(\boldsymbol\psi_{\ell})$ is a $2\times2$ positive-definite covariance matrix process of the Gaussian kernel (i.e., the covariance kernel), $\mathcal{M}_{\nu}(\cdot)$ is the Matérn correlation function \citep{M1986} with smoothness $\nu$ and a deliberate valid scale parameter:

\begin{equation}
\mathcal{C}_{\text{Matérn}}(h; \boldsymbol\psi) = \sigma^2 \frac{1}{2^{\nu-1}\Gamma(\nu)}
\left(\frac{h}{\phi}\right)^{\nu}
\mathcal{K}_{\nu}\!\left(\frac{h}{\phi}\right)
\end{equation}
where $\sigma^2 > 0$, $\phi > 0$, $\nu > 0$, $\mathcal{K}_{\nu}(\cdot)$ is the modified Bessel function of the second kind of order $\nu$ \citep{W1995}, and $\Gamma(\cdot)$ is the gamma function \citep{D1959}.
Finally, $Q_{ij} = h_{(\boldsymbol\Sigma_{i} + \boldsymbol\Sigma_{j})/2}$ is a semi-metric distance function \citep{S1938} defined as:

\begin{equation}
Q_{ij}=(s_i - s_j)^\mathsf{T}\Bigl(\frac{\boldsymbol\Sigma_i + \boldsymbol\Sigma_j}{2}\Bigr)^{-1}(s_i - s_j) , \ \ s_i , s_j \in \mathcal{S}
\label{eq:qij}
\end{equation}
where $Q_{ij}$ reduces to the Euclidean distance when $\boldsymbol\Sigma_i = \boldsymbol\Sigma_j = \gamma \boldsymbol{I}$, with $\gamma > 0$.

For Kriging, prediction is performed both with a stationary Exponential covariance model (\ref{eq:cov_exp}) and also using a correctly specified non-stationary covariance model (\ref{eq:cov_ns}), again assuming known parameters.
Under these favourable conditions, non-stationary Kriging attains slightly better predictive performance than the proposed visual spatial learning method.
As in the stationary case, however, the performance gap remains small (\Cref{tab:ns-aniso-tilt-mean_10_phi}).
The proposed method captures the dominant spatial patterns and adapts locally to changes in correlation structure, despite lacking an explicit mechanism for encoding non-stationarity or anisotropy.
This suggests that the partial convolutional architecture, combined with the adaptive mixed smooth training objective, is able to implicitly learn spatially varying dependence structures from the single observed field.

\begin{table}[tb]
\centering
\caption{Non-stationary \textit{Matérn} --- Comparison of model performance across varying percentages of known points (20\%, 50\% and 80\%). Values represent the mean ($\pm$ standard deviation) computed over 100 independent runs. The spatial interpolation was conducted using a non-stationary \textit{Matérn} covariogram with a range parameter of 10\% of the grid size. The underlying data follows a \textit{Gaussian} distribution.}
\label{tab:ns-aniso-tilt-mean_10_phi}
\begin{tabular}{llccc}
\toprule
\textbf{Model} & \textbf{Metric}  & \textbf{20\%} & \textbf{50\%} & \textbf{80\%}  \\
\midrule
\multirow{2}{*}{Kriging} & RMSE &  \textbf{1.398 $\pm$ 0.14} &  1.270 $\pm$ 0.08 &  1.202 $\pm$ 0.05 \\
 & MI-RMSE &  0.239 $\pm$ 0.06 &  0.182 $\pm$ 0.02 &  0.119 $\pm$ 0.01 \\
\midrule
\multirow{2}{*}{Kriging NS} & RMSE &  1.426 $\pm$ 0.15 &  \textbf{1.256 $\pm$ 0.08} &  \textbf{1.177 $\pm$ 0.05} \\
 & MI-RMSE &  \textbf{0.227 $\pm$ 0.06} &  \textbf{0.178 $\pm$ 0.02} &  0.114 $\pm$ 0.02 \\
\midrule
\multirow{2}{*}{ML Base} & RMSE &  1.686 $\pm$ 0.15 &  1.697 $\pm$ 0.10 &  1.688 $\pm$ 0.10 \\
 & MI-RMSE &  0.407 $\pm$ 0.08 &  0.338 $\pm$ 0.04 &  0.224 $\pm$ 0.03 \\
\midrule
\multirow{2}{*}{ML VSL} & RMSE &  1.411 $\pm$ 0.14 &  1.283 $\pm$ 0.09 &  1.214 $\pm$ 0.06 \\
 & MI-RMSE &  0.298 $\pm$ 0.03 &  0.186 $\pm$ 0.02 &  \textbf{0.107 $\pm$ 0.02} \\
\bottomrule
\end{tabular}
\end{table}

These results highlight an important practical advantage of the proposed approach.
In settings where non-stationarity appears primarily through spatially varying mean behaviour and where reliable specification of a non-stationary and anisotropic model is difficult, the proposed visual spatial learning method provides a flexible alternative that avoids explicit covariance function modelling.
The ability to recover both global and local structure from a single realization, under sparse supervision, further supports the applicability of visual spatial learning for interpolation problems in complex and weakly specified environments.

\subsection{Composite Fields}
\label{subsec:composite_fields}

Finally, we consider a class of fields designed to challenge classical Kriging methods.
These fields are constructed by joining two distinct components: a smooth Gaussian random field similar to those used in \Cref{subsec:grf_stationary} with an Exponential covariance function (\ref{eq:cov_exp}), and a second field generated using a fixed Wave covariance function (\ref{eq:cov_wave}), inducing pronounced oscillatory behaviour and localized periodic structures.
The resulting fields exhibit sharp transitions between regions with qualitatively different spatial characteristics and are poorly described by a single global covariance model.

\begin{equation}
\mathcal{C}_{\text{Wave}}(h;\sigma^2,\phi) = \sigma^2 \frac{\phi}{h}\mathrm{sin}\left(\frac{h}{\phi}\right), \quad \phi > 0
\label{eq:cov_wave}
\end{equation}

In this setting, Ordinary Kriging struggles to accommodate the coexistence of smooth and oscillatory regimes, leading to over-smoothing in wave-dominated regions or spurious oscillations elsewhere, depending on the chosen covariance specification.
In contrast, the proposed visual spatial learning approach preserve both the smooth background structure and the localized oscillatory patterns more faithfully, with fewer boundary artifacts at the interfaces between regimes.
This is reflected in improved RMSE and MI-RMSE across most settings (\Cref{tab:exponential-times-wave_10_phi}).

\begin{table}[tb]
\centering
\caption{\textit{Exponential$\times$Wave} --- Comparison of model performance across varying percentages of known points (20\%, 50\% and 80\%). Values represent the mean ($\pm$ standard deviation) computed over 100 independent runs. The spatial interpolation was conducted using an \textit{Exponential$\:\times$Wave} covariogram with a range parameter of 10\% of the grid size. The underlying data follows a \textit{Gaussian} distribution.}
\label{tab:exponential-times-wave_10_phi}
\begin{tabular}{llccc}
\toprule
\textbf{Model} & \textbf{Metric}  & \textbf{20\%} & \textbf{50\%} & \textbf{80\%}  \\
\midrule
\multirow{2}{*}{Kriging} & RMSE &  0.734 $\pm$ 0.17 &  0.621 $\pm$ 0.11 &  \textbf{0.569 $\pm$ 0.08} \\
 & MI-RMSE &  \textbf{0.177 $\pm$ 0.05} &  0.126 $\pm$ 0.03 &  0.081 $\pm$ 0.02 \\
\midrule
\multirow{2}{*}{ML Base} & RMSE &  1.037 $\pm$ 0.31 &  1.007 $\pm$ 0.24 &  0.989 $\pm$ 0.21 \\
 & MI-RMSE &  0.490 $\pm$ 0.11 &  0.371 $\pm$ 0.07 &  0.252 $\pm$ 0.05 \\
\midrule
\multirow{2}{*}{ML VSL} & RMSE &  \textbf{0.729 $\pm$ 0.15} &  \textbf{0.621 $\pm$ 0.11} &  0.575 $\pm$ 0.08 \\
 & MI-RMSE &  0.189 $\pm$ 0.05 &  \textbf{0.117 $\pm$ 0.03} &  \textbf{0.070 $\pm$ 0.02} \\
\bottomrule
\end{tabular}
\end{table}

These results highlight a key advantage of the proposed method, its ability to adapt to complex, non-Gaussian, and composite spatial structures without requiring explicit covariance modelling or manual variography studies.
While Kriging remains superior or marginally better in regimes where its assumptions are satisfied and the covariance is correctly specified, the proposed visual spatial learning (CNN) architecture approach demonstrates clear benefits in settings with heterogeneous or oscillatory spatial behaviour that challenge classical geostatistical models.

Overall, the numerical experiments support the view that the proposed method provides a robust and flexible alternative to Kriging, achieving near-optimal performance in idealized Gaussian settings and superior performance in more complex and structurally diverse spatial interpolation problems.

\nocite{Rcitation}
\nocite{ggplot_W2016}
\nocite{spmodel_DHM2023}
\nocite{sf_PB2023}
\nocite{Ansel_PyTorch_2_Faster_2024}
\nocite{2020NumPy-Array}
\nocite{tange_2025_17692695}

\section{Case Study: IPMA PELAGO22}
\label{sec:ipma}
The proposed visual spatial learning (CNN) interpolation model was applied to real data collected during the PELAGO22 acoustic survey \citep{PELAGO2022} conducted by the \textit{Instituto Português do Mar e da Atmosfera} (IPMA).
Beginning in 1995, the PELAGO surveys, are designed to characterize the spatial distribution and abundance of small pelagic fish species.
The PELAGO22 survey focused mainly on sardine (\textit{Sardina pilchardus}), anchovy (\textit{Engraulis encrasicolus}), and mackerel (\textit{Scomber colias}), along the Portuguese continental shelf and the Bay of \textit{Cádiz}.

The PELAGO22 survey took place between 1 and 31 March 2022 and was executed in two parts aboard the research vessels \textit{Miguel Oliver} and \textit{Vizconde de Eza}, covering approximately 1120 nautical miles distributed across 71 acoustic transects.
The acoustic data were recorded using a SIMRAD EK80 echo sounder operating at multiple frequencies (18, 38, 70, 120, and 200 kHz), with the 38 kHz channel serving as the reference for energy integration.
Biological sampling was performed through 42 midwater trawl hauls and 29 purse-seine operations using the commercial vessel \textit{Deus Não Falta}, coordinated with the acoustic detections.
These operations provided ground-truth data for species identification, size, and age structure, which were later used to segment the total acoustic energy by species and length class.

\begin{figure}[t]
\centering
\makebox[\textwidth][c]{\includegraphics[width=1.2\textwidth]{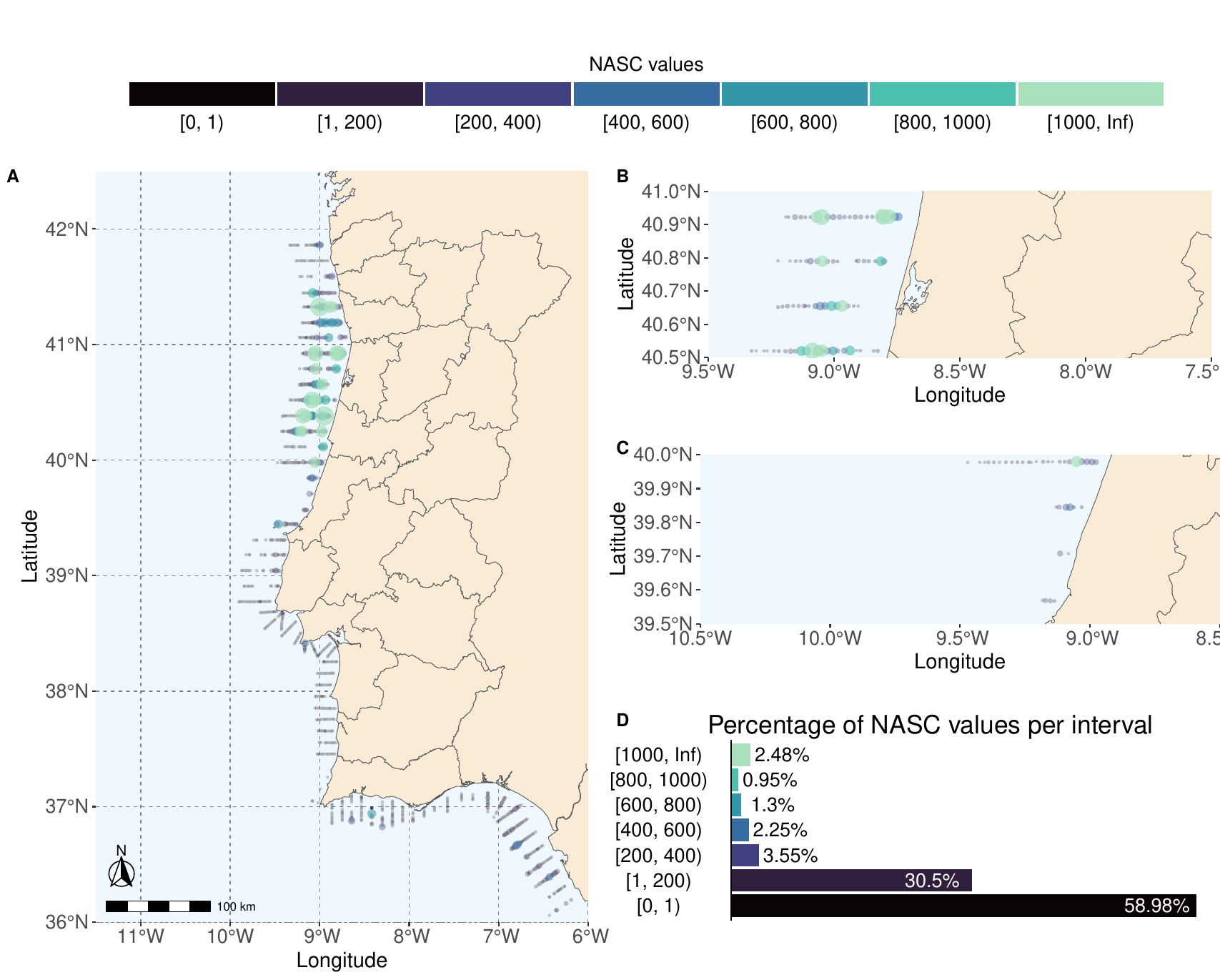}}
\caption{Spatial distribution of IPMA PELAGO22 Nautical Area Scattering Coefficient (NASC) values across the study domain.}
\label{fig:biq_plots}
\end{figure}

The resulting dataset represents a high-resolution spatial field of acoustic backscatter, measured through the Nautical Area Scattering Coefficient (NASC, $\text{m}^2\,\text{nmi}^{-2}$), irregularly distributed along transects, with strong local heterogeneity and pronounced regional contrasts in fish density.
In the present study, the focus will be on the anchovy population which, in particular, displayed spatially complex and non-stationary patterns, with high-density aggregations in the northern coast (between the \textit{Douro} and \textit{Mondego} rivers) and more fragmented distributions in the southern and \textit{Cádiz} sectors (\Cref{fig:biq_plots}).
Such patterns challenge the core assumptions of second-order stationarity and isotropy that underpin classical Kriging methods.

To evaluate the visual spatial learning interpolation model under these realistic conditions, the observed acoustic intensity values at trawl and auxiliary points were discretized into a regular grid as explained in \Cref{sec:methodology} with the point values inside each cell aggregated by summation.
This observer grid was then split into a train and test dataset of 80\% and 20\% respectably.
The train dataset served as ground truth for both methods, and predictions were assessed on the test dataset, which is a withheld subset of spatial locations sampled uniformly across the domain.
The model was then trained on the partially observed field defined by the survey transects, learning to infer a continuous surface of acoustic energy over the full spatial domain.
Unlike classical Kriging methods, the visual spatial learning model does not rely on explicit covariance modelling or parametric variogram fitting, but instead captures non-linear, location-dependent relationships through spatially localized convolutions and hierarchical feature aggregation.

\begin{table}[t]
\centering
\caption{Comparison of Ordinary Kriging with the proposed CNN Visual Spatial Learning model (CNN VSL) for a grid of $128 \times 128$ with a train/test split of 80\%/20\%.}
\label{tab:ipma_results}
\begin{tabular}{lcc}
\toprule
\textbf{Model} & \textbf{RMSE}  & \textbf{MAE}   \\
\midrule
\textbf{Kriging} & 563.661 & 305.376 \\
\textbf{CNN VSL} & 399.264 & 175.226 \\
\bottomrule
\end{tabular}
\end{table}

\Cref{tab:ipma_results} summarizes the RMSE (\ref{eq:rmse}) and the mean absolute error (MAE):
\begin{equation}
\text{MAE} = {\frac {1}{N}}\sum _{i=1}^{N}\left|y_{i}-{\hat {y_{i}}}\right|
\label{eq:mae}
\end{equation}
obtained by each method.
The MI-RMSE metric is omitted because it requires the true field, which is unavailable in this real-world setting.
The visual spatial learning model achieved the lowest prediction error, substantially improving upon the Ordinary Kriging baseline.
This improvement is consistent with the presence of pronounced local heterogeneity in the anchovy distribution, as documented in the survey, where high-density patches along the northern Portuguese shelf were interspersed with areas of low or diffuse signal, and the transition between the western and southern stock components showed no clear stationary structure.

\begin{figure}[t]
\centering
\makebox[\textwidth][c]{\includegraphics[width=1.1\textwidth]{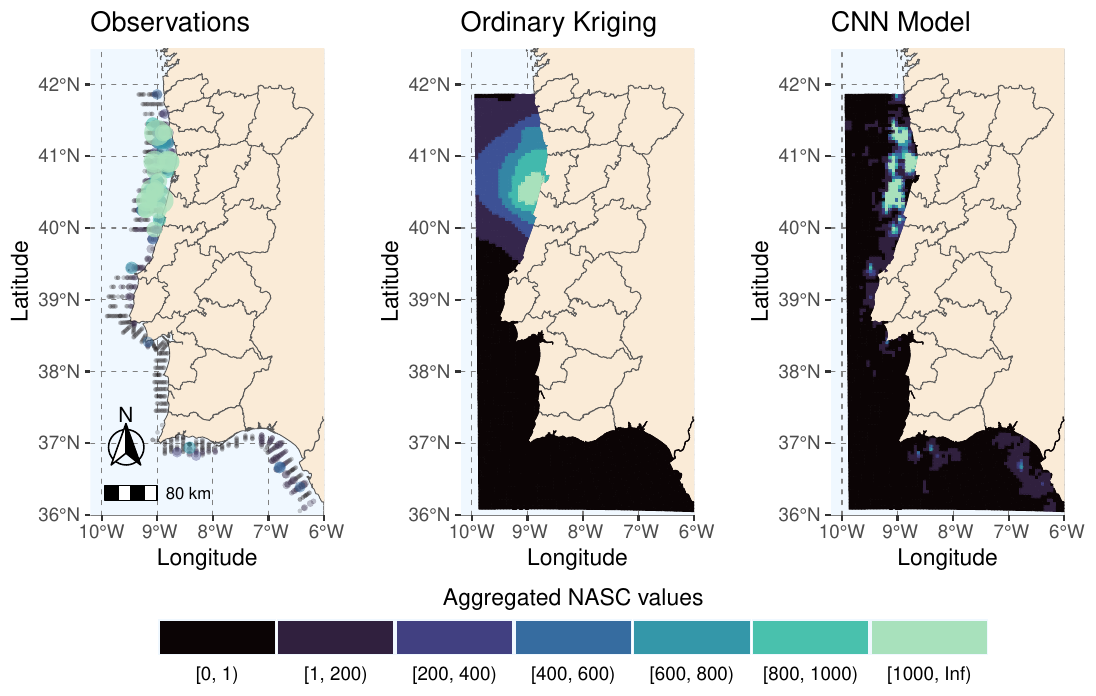}}
\caption{Full grid interpolation of Ordinary Kriging and the proposed CNN model for a grid of $128 \times 128$ with a train/test split of 80\%/20\%.}
\label{fig:biq_plots_ok_ml_slides}
\end{figure}

These differences are visually evident in \Cref{fig:biq_plots_ok_ml_slides}, which displays the interpolated acoustic fields produced by each method.
The visual spatial learning CNN model reconstruction preserves the geometry of the high-intensity anchovy aggregations identified in the survey's northern transects, reproducing their spatial extent and internal heterogeneity more faithfully than Ordinary Kriging.
The Ordinary Kriging prediction yields a smoother field with attenuated peak intensities and inflated values in surrounding low-density regions, a consequence of its reliance on a global variogram fit.
Moreover, the visual spatial learning CNN model captures secondary structures observed near the \textit{Cádiz} sector that are almost absent from the Kriging surface.

Overall, the empirical evidence demonstrates that the proposed visual spatial learning CNN model approach provides a more accurate and spatially coherent representation of the underlying acoustic field.
This supports the central premise of the study: that data-driven single-field spatial learning methods can offer a competitive and often superior alternative to classical geostatistical tools in challenging ecological settings.

\section{Conclusion}
This work examined the problem of reconstructing a spatially correlated field from sparse and irregularly distributed observations by leveraging a convolutional neural network specifically designed for single-field spatial interpolation.
Building upon the principles of U-Net style architectures, deep image prior methodologies, partial convolutions and an improved masked loss formulation, the proposed model provides a principled framework for learning spatial dependencies directly from data while accounting for uncertain or missing regions in the input domain.

More broadly, the studies illustrate that convolutional architectures, even in relatively simple forms, can effectively internalize spatial correlation structures without relying on explicit covariance modelling.
Future research may extend the present framework to multivariate fields or explore uncertainty quantification strategies consistent with the probabilistic foundations of spatial statistics.

\newpage
\paragraph{\textnormal{\textbf{Acknowledgements}}}
The first author received support from FCT (Foundation for Science and Technology) through the Individual PhD Scholarship 2024.06508.BDANA.
Computational resouces were provided by FCT 2025.09540.CPCA.A1.
The research of the first and second authors were partially financed by Portuguese Funds through FCT (Foundation for Science and Technology) within the Project \href{https://doi.org/10.54499/UID/00013/2025}{UID/00013/2025} --- Centre of Mathematics of the University of Minho (CMAT/UMinho) (\url{https://doi.org/10.54499/UID/00013/2025}).
The third author received support from national funds through FCT --- Foundation for Science and Technology, under the reference \href{https://doi.org/10.54499/UID/50014/2025}{UID/50014/2025} (\url{https://doi.org/10.54499/UID/50014/2025}).
This work was supported by Portuguese Funds through FCT (Foundation for Science and Technology) within the \href{https://doi.org/10.54499/2024.15617.PEX}{Project 2024.15617.PEX} (\url{https://doi.org/10.54499/2024.15617.PEX}).
The fourth author was funded by FCT through the strategic project UIDB/04292/2020 awarded to MARE and through project LA/P/0069/2020 granted to the Associate Laboratory ARNET.
The PELAGO survey is supported by PNAB/EU-DCF --- Programa Nacional de Amostragem Biológica/EU-Data Collection Framework [MAR-03.02.01-FEAMP-0015].

\clearpage
\bibliography{\bib}
\clearpage

\appendix
% Enter appendix text:
\section{Complete Numerical Experiments}
\label{sec:complete_num_exp}

\subsection{Stationary and Isotropic Gaussian Random Fields}
\label{ap:subsec:grf_stationary}

We begin with the benchmark setting most favourable to classical geostatistical methods, namely synthetic spatial fields generated from stationary and isotropic Gaussian random fields.
Spatial fields are simulated as single realizations of a zero-mean Gaussian process with a stationary isotropic Exponential covariance function (\ref{eq:cov_exp}), where the variance is fixed at $\sigma^2 = 1$ and the range parameter $\phi$ is varied across experiments.
Observations are sampled randomly over the spatial domain, with observed values drawn from $\mathcal{N}(0,1)$ according to the underlying Gaussian process.

\Crefrange{tab:complete_gaussian-v2_10_phi}{tab:complete_gaussian-v2_80_phi} present the full set of results for different values of the range parameter $\phi$.
While the numerical values vary slightly with $\phi$, the qualitative behaviour and relative performance of the methods remain consistent across all settings.
For this reason, the main text reports only a representative subset of results, while this appendix provides the complete tables for reference.

In this regime, Ordinary Kriging (OK) is theoretically optimal, as it coincides with the best linear unbiased predictor (BLUP) under the correct specification of the covariance structure.
As expected, OK consistently attains the lowest prediction error across all values of $\phi$.
Nevertheless, the proposed visual spatial learning CNN (ML VSL) approach achieves performance that is very close to that of OK, with only marginally higher error.
Importantly, the visual spatial learning method is trained on a single partially observed realization and does not rely on explicit covariance modelling or knowledge of the true generative process.

\begin{figure}[b]
\centering
\includegraphics[width=\textwidth]{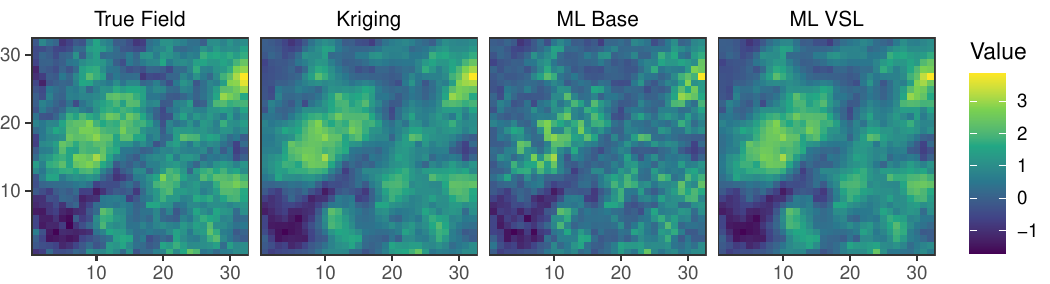}
\caption{\textit{Exponential} --- Visualisation of spatial interpolation methods for a representative realisation of a stationary isotropic Gaussian random field with \textit{Exponential} covariance. The panels show the True field alongside predictions from Ordinary Kriging, ML Base, and the proposed ML VSL method.}
\label{fig:gaussian-v2-20-phi-20-run_true_ok_ml-001_ml-004_002_plot}
\end{figure}

Across all range parameters considered, both methods yield visually smooth reconstructions (\Cref{fig:gaussian-v2-20-phi-20-run_true_ok_ml-001_ml-004_002_plot}) that faithfully recover the global correlation structure of the underlying field.
These results demonstrate that, even under conditions that strictly satisfy the assumptions of classical Kriging, the proposed visual spatial learning approach remains competitive, exhibiting only a minor loss in accuracy relative to the optimal linear predictor.
This robustness provides a useful baseline for interpreting the more challenging non-stationary and heterogeneous settings considered in \Cref{subsec:non-stationary,subsec:composite_fields} of the paper.

\paragraph{\textbf{Metrics}}
RMSE measures pointwise prediction error on unobserved locations, while MI-RMSE denotes the RMSE of Moran's~\textit{I} between the predicted and reference fields computed on the full interpolated field, thereby assessing preservation of global spatial dependence.
For Kriging, covariance parameters are assumed known to isolate modelling assumptions from variogram estimation effects.

\begin{landscape}
\begin{table}[tb]
\centering
\caption{\textit{Exponential} --- Comparison of model performance across varying percentages of known points (20\%, 30\%, 40\%, 50\%, 60\%, 70\% and 80\%). Values represent the mean ($\pm$ standard deviation) computed over 100 independent runs. The spatial interpolation was conducted using an \textit{Exponential} covariogram with a range parameter of 10\% of the grid size. The underlying data follows a \textit{Gaussian} distribution.}
\label{tab:complete_gaussian-v2_10_phi}
\begin{tabular}{llccccccc}
\toprule
\textbf{Model} & \textbf{Metric}  & \textbf{20\%} & \textbf{30\%} & \textbf{40\%} & \textbf{50\%} & \textbf{60\%} & \textbf{70\%} & \textbf{80\%}  \\
\midrule
\multirow{2}{*}{Kriging} & RMSE &  \textbf{0.658 $\pm$ 0.06} &  \textbf{0.632 $\pm$ 0.06} &  \textbf{0.618 $\pm$ 0.05} &  \textbf{0.585 $\pm$ 0.04} &  \textbf{0.564 $\pm$ 0.03} &  \textbf{0.554 $\pm$ 0.03} &  \textbf{0.544 $\pm$ 0.02} \\
 & MI-RMSE &  \textbf{0.193 $\pm$ 0.03} &  0.174 $\pm$ 0.02 &  0.155 $\pm$ 0.02 &  0.137 $\pm$ 0.02 &  0.119 $\pm$ 0.02 &  0.104 $\pm$ 0.01 &  0.088 $\pm$ 0.01 \\
\midrule
\multirow{2}{*}{ML Base} & RMSE &  0.961 $\pm$ 0.11 &  0.961 $\pm$ 0.11 &  0.965 $\pm$ 0.10 &  0.971 $\pm$ 0.09 &  0.959 $\pm$ 0.08 &  0.957 $\pm$ 0.08 &  0.949 $\pm$ 0.08 \\
 & MI-RMSE &  0.484 $\pm$ 0.08 &  0.483 $\pm$ 0.08 &  0.424 $\pm$ 0.07 &  0.405 $\pm$ 0.04 &  0.370 $\pm$ 0.04 &  0.323 $\pm$ 0.03 &  0.265 $\pm$ 0.03 \\
\midrule
\multirow{2}{*}{ML VSL} & RMSE &  0.664 $\pm$ 0.07 &  0.642 $\pm$ 0.06 &  0.628 $\pm$ 0.05 &  0.592 $\pm$ 0.04 &  0.571 $\pm$ 0.03 &  0.562 $\pm$ 0.03 &  0.550 $\pm$ 0.03 \\
 & MI-RMSE &  0.197 $\pm$ 0.03 &  \textbf{0.170 $\pm$ 0.03} &  \textbf{0.149 $\pm$ 0.02} &  \textbf{0.128 $\pm$ 0.02} &  \textbf{0.108 $\pm$ 0.02} &  \textbf{0.091 $\pm$ 0.01} &  \textbf{0.075 $\pm$ 0.01} \\
\bottomrule
\end{tabular}
\end{table}

\begin{table}[tb]
\centering
\caption{\textit{Exponential} --- Comparison of model performance across varying percentages of known points (20\%, 30\%, 40\%, 50\%, 60\%, 70\% and 80\%). Values represent the mean ($\pm$ standard deviation) computed over 100 independent runs. The spatial interpolation was conducted using an \textit{Exponential} covariogram with a range parameter of 20\% of the grid size. The underlying data follows a \textit{Gaussian} distribution.}
\label{tab:complete_gaussian-v2_20_phi}
\begin{tabular}{llccccccc}
\toprule
\textbf{Model} & \textbf{Metric}  & \textbf{20\%} & \textbf{30\%} & \textbf{40\%} & \textbf{50\%} & \textbf{60\%} & \textbf{70\%} & \textbf{80\%}  \\
\midrule
\multirow{2}{*}{Kriging} & RMSE &  \textbf{0.478 $\pm$ 0.05} &  \textbf{0.460 $\pm$ 0.04} &  \textbf{0.448 $\pm$ 0.03} &  \textbf{0.421 $\pm$ 0.03} &  \textbf{0.405 $\pm$ 0.03} &  \textbf{0.397 $\pm$ 0.02} &  \textbf{0.389 $\pm$ 0.02} \\
 & MI-RMSE &  0.127 $\pm$ 0.03 &  0.113 $\pm$ 0.02 &  0.101 $\pm$ 0.02 &  0.090 $\pm$ 0.02 &  0.078 $\pm$ 0.02 &  0.068 $\pm$ 0.01 &  0.058 $\pm$ 0.01 \\
\midrule
\multirow{2}{*}{ML Base} & RMSE &  0.889 $\pm$ 0.15 &  0.887 $\pm$ 0.14 &  0.876 $\pm$ 0.13 &  0.874 $\pm$ 0.12 &  0.850 $\pm$ 0.11 &  0.848 $\pm$ 0.13 &  0.855 $\pm$ 0.11 \\
 & MI-RMSE &  0.471 $\pm$ 0.14 &  0.466 $\pm$ 0.13 &  0.393 $\pm$ 0.13 &  0.373 $\pm$ 0.11 &  0.338 $\pm$ 0.11 &  0.300 $\pm$ 0.10 &  0.267 $\pm$ 0.07 \\
\midrule
\multirow{2}{*}{ML VSL} & RMSE &  0.488 $\pm$ 0.05 &  0.478 $\pm$ 0.05 &  0.466 $\pm$ 0.04 &  0.433 $\pm$ 0.03 &  0.419 $\pm$ 0.02 &  0.413 $\pm$ 0.02 &  0.400 $\pm$ 0.02 \\
 & MI-RMSE &  \textbf{0.121 $\pm$ 0.03} &  \textbf{0.105 $\pm$ 0.02} &  \textbf{0.094 $\pm$ 0.02} &  \textbf{0.081 $\pm$ 0.02} &  \textbf{0.069 $\pm$ 0.01} &  \textbf{0.059 $\pm$ 0.01} &  \textbf{0.048 $\pm$ 0.01} \\
\bottomrule
\end{tabular}
\end{table}

\begin{table}[tb]
\centering
\caption{\textit{Exponential} --- Comparison of model performance across varying percentages of known points (20\%, 30\%, 40\%, 50\%, 60\%, 70\% and 80\%). Values represent the mean ($\pm$ standard deviation) computed over 100 independent runs. The spatial interpolation was conducted using an \textit{Exponential} covariogram with a range parameter of 30\% of the grid size. The underlying data follows a \textit{Gaussian} distribution.}
\label{tab:complete_gaussian-v2_30_phi}
\begin{tabular}{llccccccc}
\toprule
\textbf{Model} & \textbf{Metric}  & \textbf{20\%} & \textbf{30\%} & \textbf{40\%} & \textbf{50\%} & \textbf{60\%} & \textbf{70\%} & \textbf{80\%}  \\
\midrule
\multirow{2}{*}{Kriging} & RMSE &  \textbf{0.392 $\pm$ 0.04} &  \textbf{0.380 $\pm$ 0.04} &  \textbf{0.367 $\pm$ 0.03} &  \textbf{0.345 $\pm$ 0.02} &  \textbf{0.330 $\pm$ 0.02} &  \textbf{0.326 $\pm$ 0.02} &  \textbf{0.318 $\pm$ 0.02} \\
 & MI-RMSE &  0.104 $\pm$ 0.02 &  0.092 $\pm$ 0.02 &  0.082 $\pm$ 0.02 &  0.073 $\pm$ 0.02 &  0.064 $\pm$ 0.01 &  0.055 $\pm$ 0.01 &  0.048 $\pm$ 0.01 \\
\midrule
\multirow{2}{*}{ML Base} & RMSE &  0.808 $\pm$ 0.16 &  0.781 $\pm$ 0.13 &  0.782 $\pm$ 0.14 &  0.736 $\pm$ 0.13 &  0.731 $\pm$ 0.12 &  0.715 $\pm$ 0.15 &  0.725 $\pm$ 0.13 \\
 & MI-RMSE &  0.431 $\pm$ 0.17 &  0.406 $\pm$ 0.18 &  0.331 $\pm$ 0.16 &  0.285 $\pm$ 0.16 &  0.279 $\pm$ 0.14 &  0.234 $\pm$ 0.14 &  0.220 $\pm$ 0.11 \\
\midrule
\multirow{2}{*}{ML VSL} & RMSE &  0.405 $\pm$ 0.04 &  0.405 $\pm$ 0.04 &  0.392 $\pm$ 0.04 &  0.361 $\pm$ 0.03 &  0.349 $\pm$ 0.02 &  0.347 $\pm$ 0.02 &  0.332 $\pm$ 0.02 \\
 & MI-RMSE &  \textbf{0.095 $\pm$ 0.03} &  \textbf{0.083 $\pm$ 0.02} &  \textbf{0.074 $\pm$ 0.02} &  \textbf{0.064 $\pm$ 0.02} &  \textbf{0.054 $\pm$ 0.01} &  \textbf{0.046 $\pm$ 0.01} &  \textbf{0.038 $\pm$ 0.01} \\
\bottomrule
\end{tabular}
\end{table}

\begin{table}[tb]
\centering
\caption{\textit{Exponential} --- Comparison of model performance across varying percentages of known points (20\%, 30\%, 40\%, 50\%, 60\%, 70\% and 80\%). Values represent the mean ($\pm$ standard deviation) computed over 100 independent runs. The spatial interpolation was conducted using an \textit{Exponential} covariogram with a range parameter of 40\% of the grid size. The underlying data follows a \textit{Gaussian} distribution.}
\label{tab:complete_gaussian-v2_40_phi}
\begin{tabular}{llccccccc}
\toprule
\textbf{Model} & \textbf{Metric}  & \textbf{20\%} & \textbf{30\%} & \textbf{40\%} & \textbf{50\%} & \textbf{60\%} & \textbf{70\%} & \textbf{80\%}  \\
\midrule
\multirow{2}{*}{Kriging} & RMSE &  \textbf{0.341 $\pm$ 0.03} &  \textbf{0.330 $\pm$ 0.03} &  \textbf{0.317 $\pm$ 0.02} &  \textbf{0.299 $\pm$ 0.02} &  \textbf{0.287 $\pm$ 0.02} &  \textbf{0.282 $\pm$ 0.01} &  \textbf{0.275 $\pm$ 0.01} \\
 & MI-RMSE &  0.092 $\pm$ 0.02 &  0.082 $\pm$ 0.02 &  0.073 $\pm$ 0.02 &  0.065 $\pm$ 0.02 &  0.057 $\pm$ 0.01 &  0.050 $\pm$ 0.01 &  0.042 $\pm$ 0.01 \\
\midrule
\multirow{2}{*}{ML Base} & RMSE &  0.748 $\pm$ 0.16 &  0.713 $\pm$ 0.14 &  0.693 $\pm$ 0.14 &  0.672 $\pm$ 0.12 &  0.671 $\pm$ 0.13 &  0.623 $\pm$ 0.15 &  0.638 $\pm$ 0.14 \\
 & MI-RMSE &  0.411 $\pm$ 0.18 &  0.374 $\pm$ 0.18 &  0.298 $\pm$ 0.18 &  0.271 $\pm$ 0.16 &  0.267 $\pm$ 0.14 &  0.205 $\pm$ 0.14 &  0.197 $\pm$ 0.12 \\
\midrule
\multirow{2}{*}{ML VSL} & RMSE &  0.355 $\pm$ 0.04 &  0.361 $\pm$ 0.04 &  0.348 $\pm$ 0.04 &  0.318 $\pm$ 0.02 &  0.309 $\pm$ 0.02 &  0.307 $\pm$ 0.03 &  0.293 $\pm$ 0.02 \\
 & MI-RMSE &  \textbf{0.082 $\pm$ 0.02} &  \textbf{0.072 $\pm$ 0.02} &  \textbf{0.064 $\pm$ 0.02} &  \textbf{0.055 $\pm$ 0.02} &  \textbf{0.047 $\pm$ 0.01} &  \textbf{0.039 $\pm$ 0.01} &  \textbf{0.032 $\pm$ 0.01} \\
\bottomrule
\end{tabular}
\end{table}

\begin{table}[tb]
\centering
\caption{\textit{Exponential} --- Comparison of model performance across varying percentages of known points (20\%, 30\%, 40\%, 50\%, 60\%, 70\% and 80\%). Values represent the mean ($\pm$ standard deviation) computed over 100 independent runs. The spatial interpolation was conducted using an \textit{Exponential} covariogram with a range parameter of 50\% of the grid size. The underlying data follows a \textit{Gaussian} distribution.}
\label{tab:complete_gaussian-v2_50_phi}
\begin{tabular}{llccccccc}
\toprule
\textbf{Model} & \textbf{Metric}  & \textbf{20\%} & \textbf{30\%} & \textbf{40\%} & \textbf{50\%} & \textbf{60\%} & \textbf{70\%} & \textbf{80\%}  \\
\midrule
\multirow{2}{*}{Kriging} & RMSE &  \textbf{0.306 $\pm$ 0.03} &  \textbf{0.296 $\pm$ 0.03} &  \textbf{0.284 $\pm$ 0.02} &  \textbf{0.268 $\pm$ 0.02} &  \textbf{0.256 $\pm$ 0.02} &  \textbf{0.253 $\pm$ 0.01} &  \textbf{0.246 $\pm$ 0.01} \\
 & MI-RMSE &  0.085 $\pm$ 0.02 &  0.076 $\pm$ 0.02 &  0.069 $\pm$ 0.02 &  0.060 $\pm$ 0.02 &  0.052 $\pm$ 0.01 &  0.047 $\pm$ 0.01 &  0.039 $\pm$ 0.01 \\
\midrule
\multirow{2}{*}{ML Base} & RMSE &  0.692 $\pm$ 0.15 &  0.660 $\pm$ 0.13 &  0.626 $\pm$ 0.13 &  0.610 $\pm$ 0.13 &  0.587 $\pm$ 0.13 &  0.581 $\pm$ 0.14 &  0.587 $\pm$ 0.14 \\
 & MI-RMSE &  0.385 $\pm$ 0.18 &  0.360 $\pm$ 0.18 &  0.265 $\pm$ 0.18 &  0.244 $\pm$ 0.16 &  0.213 $\pm$ 0.14 &  0.201 $\pm$ 0.14 &  0.187 $\pm$ 0.12 \\
\midrule
\multirow{2}{*}{ML VSL} & RMSE &  0.322 $\pm$ 0.04 &  0.331 $\pm$ 0.04 &  0.317 $\pm$ 0.03 &  0.290 $\pm$ 0.02 &  0.282 $\pm$ 0.02 &  0.280 $\pm$ 0.02 &  0.266 $\pm$ 0.02 \\
 & MI-RMSE &  \textbf{0.074 $\pm$ 0.02} &  \textbf{0.065 $\pm$ 0.02} &  \textbf{0.058 $\pm$ 0.02} &  \textbf{0.049 $\pm$ 0.02} &  \textbf{0.042 $\pm$ 0.01} &  \textbf{0.035 $\pm$ 0.01} &  \textbf{0.028 $\pm$ 0.01} \\
\bottomrule
\end{tabular}
\end{table}

\begin{table}[tb]
\centering
\caption{\textit{Exponential} --- Comparison of model performance across varying percentages of known points (20\%, 30\%, 40\%, 50\%, 60\%, 70\% and 80\%). Values represent the mean ($\pm$ standard deviation) computed over 100 independent runs. The spatial interpolation was conducted using an \textit{Exponential} covariogram with a range parameter of 60\% of the grid size. The underlying data follows a \textit{Gaussian} distribution.}
\label{tab:complete_gaussian-v2_60_phi}
\begin{tabular}{llccccccc}
\toprule
\textbf{Model} & \textbf{Metric}  & \textbf{20\%} & \textbf{30\%} & \textbf{40\%} & \textbf{50\%} & \textbf{60\%} & \textbf{70\%} & \textbf{80\%}  \\
\midrule
\multirow{2}{*}{Kriging} & RMSE &  \textbf{0.279 $\pm$ 0.03} &  \textbf{0.270 $\pm$ 0.03} &  \textbf{0.260 $\pm$ 0.02} &  \textbf{0.244 $\pm$ 0.02} &  \textbf{0.234 $\pm$ 0.02} &  \textbf{0.230 $\pm$ 0.01} &  \textbf{0.225 $\pm$ 0.01} \\
 & MI-RMSE &  0.081 $\pm$ 0.02 &  0.073 $\pm$ 0.02 &  0.066 $\pm$ 0.02 &  0.057 $\pm$ 0.02 &  0.050 $\pm$ 0.01 &  0.044 $\pm$ 0.01 &  0.038 $\pm$ 0.01 \\
\midrule
\multirow{2}{*}{ML Base} & RMSE &  0.651 $\pm$ 0.14 &  0.611 $\pm$ 0.12 &  0.586 $\pm$ 0.13 &  0.559 $\pm$ 0.13 &  0.540 $\pm$ 0.13 &  0.520 $\pm$ 0.14 &  0.514 $\pm$ 0.14 \\
 & MI-RMSE &  0.375 $\pm$ 0.17 &  0.338 $\pm$ 0.18 &  0.254 $\pm$ 0.18 &  0.224 $\pm$ 0.16 &  0.204 $\pm$ 0.15 &  0.173 $\pm$ 0.13 &  0.155 $\pm$ 0.12 \\
\midrule
\multirow{2}{*}{ML VSL} & RMSE &  0.296 $\pm$ 0.03 &  0.310 $\pm$ 0.04 &  0.297 $\pm$ 0.03 &  0.268 $\pm$ 0.02 &  0.262 $\pm$ 0.02 &  0.260 $\pm$ 0.03 &  0.245 $\pm$ 0.02 \\
 & MI-RMSE &  \textbf{0.068 $\pm$ 0.02} &  \textbf{0.060 $\pm$ 0.02} &  \textbf{0.054 $\pm$ 0.02} &  \textbf{0.045 $\pm$ 0.02} &  \textbf{0.039 $\pm$ 0.01} &  \textbf{0.032 $\pm$ 0.01} &  \textbf{0.026 $\pm$ 0.01} \\
\bottomrule
\end{tabular}
\end{table}

\begin{table}[tb]
\centering
\caption{\textit{Exponential} --- Comparison of model performance across varying percentages of known points (20\%, 30\%, 40\%, 50\%, 60\%, 70\% and 80\%). Values represent the mean ($\pm$ standard deviation) computed over 100 independent runs. The spatial interpolation was conducted using an \textit{Exponential} covariogram with a range parameter of 70\% of the grid size. The underlying data follows a \textit{Gaussian} distribution.}
\label{tab:complete_gaussian-v2_70_phi}
\begin{tabular}{llccccccc}
\toprule
\textbf{Model} & \textbf{Metric}  & \textbf{20\%} & \textbf{30\%} & \textbf{40\%} & \textbf{50\%} & \textbf{60\%} & \textbf{70\%} & \textbf{80\%}  \\
\midrule
\multirow{2}{*}{Kriging} & RMSE &  \textbf{0.258 $\pm$ 0.03} &  \textbf{0.250 $\pm$ 0.02} &  \textbf{0.241 $\pm$ 0.02} &  \textbf{0.226 $\pm$ 0.01} &  \textbf{0.217 $\pm$ 0.01} &  \textbf{0.214 $\pm$ 0.01} &  \textbf{0.209 $\pm$ 0.01} \\
 & MI-RMSE &  0.078 $\pm$ 0.02 &  0.070 $\pm$ 0.02 &  0.063 $\pm$ 0.02 &  0.055 $\pm$ 0.02 &  0.049 $\pm$ 0.01 &  0.043 $\pm$ 0.01 &  0.037 $\pm$ 0.01 \\
\midrule
\multirow{2}{*}{ML Base} & RMSE &  0.627 $\pm$ 0.15 &  0.565 $\pm$ 0.13 &  0.545 $\pm$ 0.13 &  0.519 $\pm$ 0.12 &  0.502 $\pm$ 0.13 &  0.478 $\pm$ 0.13 &  0.475 $\pm$ 0.13 \\
 & MI-RMSE &  0.367 $\pm$ 0.17 &  0.316 $\pm$ 0.20 &  0.248 $\pm$ 0.18 &  0.202 $\pm$ 0.16 &  0.195 $\pm$ 0.15 &  0.166 $\pm$ 0.14 &  0.143 $\pm$ 0.11 \\
\midrule
\multirow{2}{*}{ML VSL} & RMSE &  0.276 $\pm$ 0.03 &  0.293 $\pm$ 0.04 &  0.280 $\pm$ 0.04 &  0.252 $\pm$ 0.02 &  0.247 $\pm$ 0.02 &  0.246 $\pm$ 0.03 &  0.230 $\pm$ 0.02 \\
 & MI-RMSE &  \textbf{0.064 $\pm$ 0.02} &  \textbf{0.057 $\pm$ 0.02} &  \textbf{0.051 $\pm$ 0.02} &  \textbf{0.042 $\pm$ 0.02} &  \textbf{0.036 $\pm$ 0.01} &  \textbf{0.031 $\pm$ 0.01} &  \textbf{0.024 $\pm$ 0.01} \\
\bottomrule
\end{tabular}
\end{table}

\begin{table}[tb]
\centering
\caption{\textit{Exponential} --- Comparison of model performance across varying percentages of known points (20\%, 30\%, 40\%, 50\%, 60\%, 70\% and 80\%). Values represent the mean ($\pm$ standard deviation) computed over 100 independent runs. The spatial interpolation was conducted using an \textit{Exponential} covariogram with a range parameter of 80\% of the grid size. The underlying data follows a \textit{Gaussian} distribution.}
\label{tab:complete_gaussian-v2_80_phi}
\begin{tabular}{llccccccc}
\toprule
\textbf{Model} & \textbf{Metric}  & \textbf{20\%} & \textbf{30\%} & \textbf{40\%} & \textbf{50\%} & \textbf{60\%} & \textbf{70\%} & \textbf{80\%}  \\
\midrule
\multirow{2}{*}{Kriging} & RMSE &  \textbf{0.241 $\pm$ 0.02} &  \textbf{0.235 $\pm$ 0.02} &  \textbf{0.225 $\pm$ 0.02} &  \textbf{0.211 $\pm$ 0.01} &  \textbf{0.203 $\pm$ 0.01} &  \textbf{0.200 $\pm$ 0.01} &  \textbf{0.195 $\pm$ 0.01} \\
 & MI-RMSE &  0.076 $\pm$ 0.02 &  0.068 $\pm$ 0.02 &  0.062 $\pm$ 0.02 &  0.054 $\pm$ 0.02 &  0.047 $\pm$ 0.01 &  0.042 $\pm$ 0.01 &  0.036 $\pm$ 0.01 \\
\midrule
\multirow{2}{*}{ML Base} & RMSE &  0.572 $\pm$ 0.13 &  0.536 $\pm$ 0.12 &  0.510 $\pm$ 0.12 &  0.497 $\pm$ 0.13 &  0.476 $\pm$ 0.13 &  0.440 $\pm$ 0.12 &  0.435 $\pm$ 0.12 \\
 & MI-RMSE &  0.346 $\pm$ 0.19 &  0.303 $\pm$ 0.20 &  0.231 $\pm$ 0.18 &  0.219 $\pm$ 0.16 &  0.191 $\pm$ 0.15 &  0.150 $\pm$ 0.14 &  0.133 $\pm$ 0.11 \\
\midrule
\multirow{2}{*}{ML VSL} & RMSE &  0.261 $\pm$ 0.03 &  0.280 $\pm$ 0.04 &  0.267 $\pm$ 0.04 &  0.239 $\pm$ 0.02 &  0.235 $\pm$ 0.03 &  0.233 $\pm$ 0.03 &  0.217 $\pm$ 0.02 \\
 & MI-RMSE &  \textbf{0.061 $\pm$ 0.02} &  \textbf{0.054 $\pm$ 0.02} &  \textbf{0.048 $\pm$ 0.02} &  \textbf{0.040 $\pm$ 0.02} &  \textbf{0.034 $\pm$ 0.01} &  \textbf{0.028 $\pm$ 0.01} &  \textbf{0.023 $\pm$ 0.01} \\
\bottomrule
\end{tabular}
\end{table}

\end{landscape}

Following the previous experiment, we now report the complete results for the setting incorporating a nugget component into the data-generating process, as described in the main text.
Parameters are fixed at $\sigma^2=1$ and $\tau^2=0.3$, with the range parameter $\phi > 0$ varied across experiments.
\Crefrange{tab:complete_gaussian-nugget-03-v2_10_phi}{tab:complete_gaussian-nugget-03-v2_80_phi} report the full results for all considered values of $\phi$.

\begin{figure}[t]
\centering
\includegraphics[width=\textwidth]{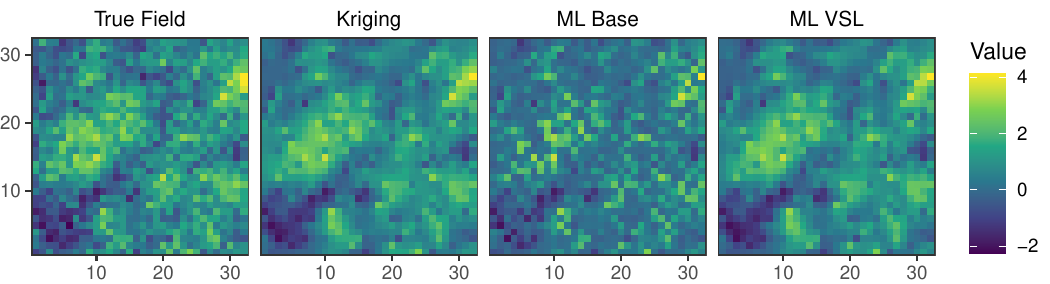}
\caption{\textit{Exponential$\,+$Nugget} --- Visualisation of spatial interpolation methods for a representative realisation of a stationary isotropic Gaussian random field with \textit{Exponential}, with \textit{Nugget} component, covariance. The panels show the True field alongside predictions from Ordinary Kriging, ML Base, and the proposed ML VSL method.}
\label{fig:gaussian-nugget-03-v2-20-phi-20-run_true_ok_ml-001_ml-004_002_plot}
\end{figure}

The results across all range parameters are broadly consistent with the findings discussed in the main text. At low observed percentages (20\%), Ordinary Kriging attains marginally lower RMSE, while ML VSL achieves competitive or superior performance from moderate observed percentages onward, with the crossover occurring earlier for shorter range parameters.
This advantage is more pronounced under the MI-RMSE metric, where ML VSL outperforms Kriging across the majority of sparsity levels and $\phi$ values, suggesting that the visual spatial learning approach reconstructs the unobserved regions more accurately even when global RMSE differences are small (\Cref{fig:gaussian-nugget-03-v2-20-phi-20-run_true_ok_ml-001_ml-004_002_plot}).

As the range parameter $\phi$ increases, both methods benefit from the stronger spatial correlation structure, yielding lower absolute errors.
However, the relative advantage of ML VSL over Kriging diminishes slightly at larger $\phi$, consistent with the expectation that smoother, longer-range fields are inherently easier for linear predictors to exploit.
Nonetheless, across all settings, ML Base performs substantially worse than both competing approaches, confirming that the architectural inductive biases of the partial convolutional network and the adaptive mixed smooth loss function are essential for achieving competitive spatial interpolation under nugget-contaminated observations.

\begin{landscape}
\begin{table}[tb]
\centering
\caption{\textit{Exponential$\,+$Nugget} --- Comparison of model performance across varying percentages of known points (20\%, 30\%, 40\%, 50\%, 60\%, 70\% and 80\%). Values represent the mean ($\pm$ standard deviation) computed over 100 independent runs. The spatial interpolation was conducted using an \textit{Exponential}, with \textit{Nugget} component, covariogram with a range parameter of 10\% of the grid size. The underlying data follows a \textit{Gaussian} distribution.}
\label{tab:complete_gaussian-nugget-03-v2_10_phi}
\begin{tabular}{llccccccc}
\toprule
\textbf{Model} & \textbf{Metric}  & \textbf{20\%} & \textbf{30\%} & \textbf{40\%} & \textbf{50\%} & \textbf{60\%} & \textbf{70\%} & \textbf{80\%}  \\
\midrule
\multirow{2}{*}{Kriging} & RMSE &  \textbf{0.905 $\pm$ 0.09} &  \textbf{0.886 $\pm$ 0.08} &  0.879 $\pm$ 0.06 &  0.854 $\pm$ 0.06 &  0.837 $\pm$ 0.05 &  0.836 $\pm$ 0.04 &  0.828 $\pm$ 0.03 \\
 & MI-RMSE &  \textbf{0.285 $\pm$ 0.08} &  \textbf{0.275 $\pm$ 0.04} &  \textbf{0.248 $\pm$ 0.03} &  \textbf{0.225 $\pm$ 0.02} &  0.199 $\pm$ 0.02 &  0.174 $\pm$ 0.02 &  0.148 $\pm$ 0.02 \\
\midrule
\multirow{2}{*}{ML Base} & RMSE &  1.119 $\pm$ 0.12 &  1.124 $\pm$ 0.12 &  1.132 $\pm$ 0.11 &  1.131 $\pm$ 0.09 &  1.125 $\pm$ 0.08 &  1.128 $\pm$ 0.09 &  1.124 $\pm$ 0.08 \\
 & MI-RMSE &  0.375 $\pm$ 0.08 &  0.383 $\pm$ 0.06 &  0.346 $\pm$ 0.04 &  0.321 $\pm$ 0.04 &  0.298 $\pm$ 0.04 &  0.260 $\pm$ 0.04 &  0.213 $\pm$ 0.03 \\
\midrule
\multirow{2}{*}{ML VSL} & RMSE &  0.909 $\pm$ 0.09 &  0.889 $\pm$ 0.08 &  \textbf{0.878 $\pm$ 0.06} &  \textbf{0.851 $\pm$ 0.06} &  \textbf{0.834 $\pm$ 0.05} &  \textbf{0.832 $\pm$ 0.04} &  \textbf{0.825 $\pm$ 0.03} \\
 & MI-RMSE &  0.342 $\pm$ 0.04 &  0.298 $\pm$ 0.03 &  0.260 $\pm$ 0.03 &  0.226 $\pm$ 0.03 &  \textbf{0.193 $\pm$ 0.02} &  \textbf{0.163 $\pm$ 0.02} &  \textbf{0.134 $\pm$ 0.02} \\
\bottomrule
\end{tabular}
\end{table}

\begin{table}[tb]
\centering
\caption{\textit{Exponential$\,+$Nugget} --- Comparison of model performance across varying percentages of known points (20\%, 30\%, 40\%, 50\%, 60\%, 70\% and 80\%). Values represent the mean ($\pm$ standard deviation) computed over 100 independent runs. The spatial interpolation was conducted using an \textit{Exponential}, with \textit{Nugget} component, covariogram with a range parameter of 20\% of the grid size. The underlying data follows a \textit{Gaussian} distribution.}
\label{tab:complete_gaussian-nugget-03-v2_20_phi}
\begin{tabular}{llccccccc}
\toprule
\textbf{Model} & \textbf{Metric}  & \textbf{20\%} & \textbf{30\%} & \textbf{40\%} & \textbf{50\%} & \textbf{60\%} & \textbf{70\%} & \textbf{80\%}  \\
\midrule
\multirow{2}{*}{Kriging} & RMSE &  0.791 $\pm$ 0.08 &  \textbf{0.781 $\pm$ 0.07} &  0.774 $\pm$ 0.06 &  0.755 $\pm$ 0.05 &  0.744 $\pm$ 0.04 &  0.745 $\pm$ 0.04 &  0.739 $\pm$ 0.03 \\
 & MI-RMSE &  \textbf{0.276 $\pm$ 0.07} &  \textbf{0.265 $\pm$ 0.04} &  \textbf{0.240 $\pm$ 0.03} &  0.218 $\pm$ 0.03 &  0.193 $\pm$ 0.03 &  0.169 $\pm$ 0.02 &  0.144 $\pm$ 0.02 \\
\midrule
\multirow{2}{*}{ML Base} & RMSE &  1.099 $\pm$ 0.15 &  1.095 $\pm$ 0.15 &  1.101 $\pm$ 0.14 &  1.099 $\pm$ 0.11 &  1.099 $\pm$ 0.11 &  1.102 $\pm$ 0.12 &  1.097 $\pm$ 0.11 \\
 & MI-RMSE &  0.390 $\pm$ 0.10 &  0.405 $\pm$ 0.09 &  0.358 $\pm$ 0.07 &  0.340 $\pm$ 0.06 &  0.327 $\pm$ 0.05 &  0.284 $\pm$ 0.05 &  0.238 $\pm$ 0.04 \\
\midrule
\multirow{2}{*}{ML VSL} & RMSE &  \textbf{0.790 $\pm$ 0.08} &  0.782 $\pm$ 0.07 &  \textbf{0.771 $\pm$ 0.06} &  \textbf{0.751 $\pm$ 0.05} &  \textbf{0.738 $\pm$ 0.04} &  \textbf{0.741 $\pm$ 0.04} &  \textbf{0.735 $\pm$ 0.03} \\
 & MI-RMSE &  0.308 $\pm$ 0.05 &  0.273 $\pm$ 0.04 &  0.243 $\pm$ 0.04 &  \textbf{0.212 $\pm$ 0.03} &  \textbf{0.182 $\pm$ 0.03} &  \textbf{0.155 $\pm$ 0.02} &  \textbf{0.127 $\pm$ 0.02} \\
\bottomrule
\end{tabular}
\end{table}

\begin{table}[tb]
\centering
\caption{\textit{Exponential$\,+$Nugget} --- Comparison of model performance across varying percentages of known points (20\%, 30\%, 40\%, 50\%, 60\%, 70\% and 80\%). Values represent the mean ($\pm$ standard deviation) computed over 100 independent runs. The spatial interpolation was conducted using an \textit{Exponential}, with \textit{Nugget} component, covariogram with a range parameter of 30\% of the grid size. The underlying data follows a \textit{Gaussian} distribution.}
\label{tab:complete_gaussian-nugget-03-v2_30_phi}
\begin{tabular}{llccccccc}
\toprule
\textbf{Model} & \textbf{Metric}  & \textbf{20\%} & \textbf{30\%} & \textbf{40\%} & \textbf{50\%} & \textbf{60\%} & \textbf{70\%} & \textbf{80\%}  \\
\midrule
\multirow{2}{*}{Kriging} & RMSE &  0.745 $\pm$ 0.08 &  \textbf{0.737 $\pm$ 0.07} &  0.731 $\pm$ 0.05 &  0.716 $\pm$ 0.05 &  0.707 $\pm$ 0.04 &  0.710 $\pm$ 0.03 &  0.704 $\pm$ 0.03 \\
 & MI-RMSE &  \textbf{0.274 $\pm$ 0.08} &  \textbf{0.267 $\pm$ 0.05} &  \textbf{0.243 $\pm$ 0.04} &  0.222 $\pm$ 0.03 &  0.196 $\pm$ 0.03 &  0.173 $\pm$ 0.03 &  0.148 $\pm$ 0.02 \\
\midrule
\multirow{2}{*}{ML Base} & RMSE &  1.066 $\pm$ 0.17 &  1.060 $\pm$ 0.17 &  1.071 $\pm$ 0.15 &  1.066 $\pm$ 0.14 &  1.070 $\pm$ 0.13 &  1.078 $\pm$ 0.14 &  1.080 $\pm$ 0.15 \\
 & MI-RMSE &  0.344 $\pm$ 0.13 &  0.373 $\pm$ 0.11 &  0.336 $\pm$ 0.09 &  0.319 $\pm$ 0.08 &  0.316 $\pm$ 0.06 &  0.283 $\pm$ 0.06 &  0.242 $\pm$ 0.05 \\
\midrule
\multirow{2}{*}{ML VSL} & RMSE &  \textbf{0.741 $\pm$ 0.08} &  0.739 $\pm$ 0.07 &  \textbf{0.730 $\pm$ 0.05} &  \textbf{0.711 $\pm$ 0.05} &  \textbf{0.703 $\pm$ 0.04} &  \textbf{0.707 $\pm$ 0.04} &  \textbf{0.702 $\pm$ 0.04} \\
 & MI-RMSE &  0.311 $\pm$ 0.06 &  0.277 $\pm$ 0.05 &  0.247 $\pm$ 0.04 &  \textbf{0.216 $\pm$ 0.04} &  \textbf{0.187 $\pm$ 0.03} &  \textbf{0.158 $\pm$ 0.03} &  \textbf{0.130 $\pm$ 0.02} \\
\bottomrule
\end{tabular}
\end{table}

\begin{table}[tb]
\centering
\caption{\textit{Exponential$\,+$Nugget} --- Comparison of model performance across varying percentages of known points (20\%, 30\%, 40\%, 50\%, 60\%, 70\% and 80\%). Values represent the mean ($\pm$ standard deviation) computed over 100 independent runs. The spatial interpolation was conducted using an \textit{Exponential}, with \textit{Nugget} component, covariogram with a range parameter of 40\% of the grid size. The underlying data follows a \textit{Gaussian} distribution.}
\label{tab:complete_gaussian-nugget-03-v2_40_phi}
\begin{tabular}{llccccccc}
\toprule
\textbf{Model} & \textbf{Metric}  & \textbf{20\%} & \textbf{30\%} & \textbf{40\%} & \textbf{50\%} & \textbf{60\%} & \textbf{70\%} & \textbf{80\%}  \\
\midrule
\multirow{2}{*}{Kriging} & RMSE &  0.721 $\pm$ 0.08 &  \textbf{0.713 $\pm$ 0.06} &  \textbf{0.708 $\pm$ 0.05} &  0.694 $\pm$ 0.04 &  0.686 $\pm$ 0.04 &  \textbf{0.690 $\pm$ 0.03} &  0.685 $\pm$ 0.03 \\
 & MI-RMSE &  \textbf{0.279 $\pm$ 0.08} &  \textbf{0.271 $\pm$ 0.06} &  \textbf{0.248 $\pm$ 0.04} &  0.228 $\pm$ 0.04 &  0.202 $\pm$ 0.03 &  0.178 $\pm$ 0.03 &  0.152 $\pm$ 0.02 \\
\midrule
\multirow{2}{*}{ML Base} & RMSE &  1.035 $\pm$ 0.17 &  1.035 $\pm$ 0.17 &  1.043 $\pm$ 0.16 &  1.044 $\pm$ 0.15 &  1.049 $\pm$ 0.15 &  1.045 $\pm$ 0.14 &  1.054 $\pm$ 0.15 \\
 & MI-RMSE &  0.320 $\pm$ 0.14 &  0.350 $\pm$ 0.12 &  0.315 $\pm$ 0.10 &  0.306 $\pm$ 0.09 &  0.302 $\pm$ 0.08 &  0.270 $\pm$ 0.07 &  0.235 $\pm$ 0.06 \\
\midrule
\multirow{2}{*}{ML VSL} & RMSE &  \textbf{0.715 $\pm$ 0.08} &  0.717 $\pm$ 0.07 &  0.708 $\pm$ 0.05 &  \textbf{0.690 $\pm$ 0.04} &  \textbf{0.683 $\pm$ 0.04} &  0.690 $\pm$ 0.04 &  \textbf{0.684 $\pm$ 0.03} \\
 & MI-RMSE &  0.323 $\pm$ 0.06 &  0.288 $\pm$ 0.06 &  0.258 $\pm$ 0.05 &  \textbf{0.225 $\pm$ 0.04} &  \textbf{0.195 $\pm$ 0.04} &  \textbf{0.163 $\pm$ 0.03} &  \textbf{0.136 $\pm$ 0.02} \\
\bottomrule
\end{tabular}
\end{table}

\begin{table}[tb]
\centering
\caption{\textit{Exponential$\,+$Nugget} --- Comparison of model performance across varying percentages of known points (20\%, 30\%, 40\%, 50\%, 60\%, 70\% and 80\%). Values represent the mean ($\pm$ standard deviation) computed over 100 independent runs. The spatial interpolation was conducted using an \textit{Exponential}, with \textit{Nugget} component, covariogram with a range parameter of 50\% of the grid size. The underlying data follows a \textit{Gaussian} distribution.}
\label{tab:complete_gaussian-nugget-03-v2_50_phi}
\begin{tabular}{llccccccc}
\toprule
\textbf{Model} & \textbf{Metric}  & \textbf{20\%} & \textbf{30\%} & \textbf{40\%} & \textbf{50\%} & \textbf{60\%} & \textbf{70\%} & \textbf{80\%}  \\
\midrule
\multirow{2}{*}{Kriging} & RMSE &  0.708 $\pm$ 0.08 &  \textbf{0.698 $\pm$ 0.06} &  \textbf{0.693 $\pm$ 0.05} &  0.680 $\pm$ 0.04 &  \textbf{0.673 $\pm$ 0.04} &  \textbf{0.677 $\pm$ 0.03} &  0.673 $\pm$ 0.03 \\
 & MI-RMSE &  \textbf{0.281 $\pm$ 0.09} &  \textbf{0.275 $\pm$ 0.06} &  \textbf{0.252 $\pm$ 0.05} &  \textbf{0.233 $\pm$ 0.04} &  0.207 $\pm$ 0.03 &  0.183 $\pm$ 0.03 &  0.156 $\pm$ 0.03 \\
\midrule
\multirow{2}{*}{ML Base} & RMSE &  1.018 $\pm$ 0.19 &  1.007 $\pm$ 0.17 &  1.026 $\pm$ 0.18 &  1.027 $\pm$ 0.16 &  1.024 $\pm$ 0.14 &  1.019 $\pm$ 0.16 &  1.039 $\pm$ 0.16 \\
 & MI-RMSE &  0.298 $\pm$ 0.14 &  0.318 $\pm$ 0.13 &  0.300 $\pm$ 0.10 &  0.301 $\pm$ 0.09 &  0.291 $\pm$ 0.09 &  0.264 $\pm$ 0.07 &  0.233 $\pm$ 0.06 \\
\midrule
\multirow{2}{*}{ML VSL} & RMSE &  \textbf{0.701 $\pm$ 0.07} &  0.702 $\pm$ 0.06 &  0.694 $\pm$ 0.05 &  \textbf{0.677 $\pm$ 0.04} &  0.673 $\pm$ 0.04 &  0.679 $\pm$ 0.04 &  \textbf{0.672 $\pm$ 0.03} \\
 & MI-RMSE &  0.336 $\pm$ 0.07 &  0.302 $\pm$ 0.06 &  0.269 $\pm$ 0.05 &  0.235 $\pm$ 0.04 &  \textbf{0.203 $\pm$ 0.04} &  \textbf{0.171 $\pm$ 0.03} &  \textbf{0.140 $\pm$ 0.03} \\
\bottomrule
\end{tabular}
\end{table}

\begin{table}[tb]
\centering
\caption{\textit{Exponential$\,+$Nugget} --- Comparison of model performance across varying percentages of known points (20\%, 30\%, 40\%, 50\%, 60\%, 70\% and 80\%). Values represent the mean ($\pm$ standard deviation) computed over 100 independent runs. The spatial interpolation was conducted using an \textit{Exponential}, with \textit{Nugget} component, covariogram with a range parameter of 60\% of the grid size. The underlying data follows a \textit{Gaussian} distribution.}
\label{tab:complete_gaussian-nugget-03-v2_60_phi}
\begin{tabular}{llccccccc}
\toprule
\textbf{Model} & \textbf{Metric}  & \textbf{20\%} & \textbf{30\%} & \textbf{40\%} & \textbf{50\%} & \textbf{60\%} & \textbf{70\%} & \textbf{80\%}  \\
\midrule
\multirow{2}{*}{Kriging} & RMSE &  0.699 $\pm$ 0.08 &  \textbf{0.688 $\pm$ 0.07} &  \textbf{0.683 $\pm$ 0.05} &  0.669 $\pm$ 0.04 &  \textbf{0.663 $\pm$ 0.04} &  \textbf{0.667 $\pm$ 0.03} &  \textbf{0.663 $\pm$ 0.03} \\
 & MI-RMSE &  \textbf{0.279 $\pm$ 0.10} &  \textbf{0.276 $\pm$ 0.06} &  \textbf{0.253 $\pm$ 0.06} &  \textbf{0.237 $\pm$ 0.04} &  \textbf{0.211 $\pm$ 0.04} &  0.186 $\pm$ 0.03 &  0.159 $\pm$ 0.03 \\
\midrule
\multirow{2}{*}{ML Base} & RMSE &  1.005 $\pm$ 0.19 &  0.987 $\pm$ 0.17 &  1.002 $\pm$ 0.17 &  1.018 $\pm$ 0.17 &  1.004 $\pm$ 0.15 &  1.011 $\pm$ 0.16 &  1.001 $\pm$ 0.16 \\
 & MI-RMSE &  0.280 $\pm$ 0.14 &  0.309 $\pm$ 0.13 &  0.283 $\pm$ 0.11 &  0.288 $\pm$ 0.09 &  0.284 $\pm$ 0.08 &  0.261 $\pm$ 0.07 &  0.216 $\pm$ 0.07 \\
\midrule
\multirow{2}{*}{ML VSL} & RMSE &  \textbf{0.690 $\pm$ 0.07} &  0.692 $\pm$ 0.06 &  0.685 $\pm$ 0.05 &  \textbf{0.668 $\pm$ 0.04} &  0.665 $\pm$ 0.04 &  0.673 $\pm$ 0.04 &  0.668 $\pm$ 0.04 \\
 & MI-RMSE &  0.350 $\pm$ 0.07 &  0.315 $\pm$ 0.06 &  0.280 $\pm$ 0.05 &  0.244 $\pm$ 0.05 &  0.211 $\pm$ 0.04 &  \textbf{0.178 $\pm$ 0.03} &  \textbf{0.144 $\pm$ 0.03} \\
\bottomrule
\end{tabular}
\end{table}

\begin{table}[tb]
\centering
\caption{\textit{Exponential$\,+$Nugget} --- Comparison of model performance across varying percentages of known points (20\%, 30\%, 40\%, 50\%, 60\%, 70\% and 80\%). Values represent the mean ($\pm$ standard deviation) computed over 100 independent runs. The spatial interpolation was conducted using an \textit{Exponential}, with \textit{Nugget} component, covariogram with a range parameter of 70\% of the grid size. The underlying data follows a \textit{Gaussian} distribution.}
\label{tab:complete_gaussian-nugget-03-v2_70_phi}
\begin{tabular}{llccccccc}
\toprule
\textbf{Model} & \textbf{Metric}  & \textbf{20\%} & \textbf{30\%} & \textbf{40\%} & \textbf{50\%} & \textbf{60\%} & \textbf{70\%} & \textbf{80\%}  \\
\midrule
\multirow{2}{*}{Kriging} & RMSE &  0.694 $\pm$ 0.08 &  \textbf{0.680 $\pm$ 0.07} &  \textbf{0.675 $\pm$ 0.05} &  \textbf{0.662 $\pm$ 0.04} &  \textbf{0.656 $\pm$ 0.04} &  \textbf{0.660 $\pm$ 0.03} &  \textbf{0.656 $\pm$ 0.03} \\
 & MI-RMSE &  0.284 $\pm$ 0.10 &  \textbf{0.275 $\pm$ 0.07} &  \textbf{0.253 $\pm$ 0.06} &  \textbf{0.239 $\pm$ 0.04} &  \textbf{0.214 $\pm$ 0.04} &  0.188 $\pm$ 0.03 &  0.161 $\pm$ 0.03 \\
\midrule
\multirow{2}{*}{ML Base} & RMSE &  0.993 $\pm$ 0.19 &  0.978 $\pm$ 0.17 &  0.975 $\pm$ 0.16 &  0.992 $\pm$ 0.16 &  0.986 $\pm$ 0.15 &  1.006 $\pm$ 0.17 &  1.008 $\pm$ 0.17 \\
 & MI-RMSE &  \textbf{0.272 $\pm$ 0.14} &  0.297 $\pm$ 0.13 &  0.260 $\pm$ 0.12 &  0.275 $\pm$ 0.09 &  0.269 $\pm$ 0.09 &  0.255 $\pm$ 0.07 &  0.219 $\pm$ 0.06 \\
\midrule
\multirow{2}{*}{ML VSL} & RMSE &  \textbf{0.683 $\pm$ 0.07} &  0.685 $\pm$ 0.06 &  0.677 $\pm$ 0.05 &  0.662 $\pm$ 0.04 &  0.659 $\pm$ 0.04 &  0.667 $\pm$ 0.04 &  0.664 $\pm$ 0.05 \\
 & MI-RMSE &  0.364 $\pm$ 0.08 &  0.327 $\pm$ 0.06 &  0.291 $\pm$ 0.05 &  0.253 $\pm$ 0.05 &  0.219 $\pm$ 0.04 &  \textbf{0.183 $\pm$ 0.03} &  \textbf{0.148 $\pm$ 0.03} \\
\bottomrule
\end{tabular}
\end{table}

\begin{table}[tb]
\centering
\caption{\textit{Exponential$\,+$Nugget} --- Comparison of model performance across varying percentages of known points (20\%, 30\%, 40\%, 50\%, 60\%, 70\% and 80\%). Values represent the mean ($\pm$ standard deviation) computed over 100 independent runs. The spatial interpolation was conducted using an \textit{Exponential}, with \textit{Nugget} component, covariogram with a range parameter of 80\% of the grid size. The underlying data follows a \textit{Gaussian} distribution.}
\label{tab:complete_gaussian-nugget-03-v2_80_phi}
\begin{tabular}{llccccccc}
\toprule
\textbf{Model} & \textbf{Metric}  & \textbf{20\%} & \textbf{30\%} & \textbf{40\%} & \textbf{50\%} & \textbf{60\%} & \textbf{70\%} & \textbf{80\%}  \\
\midrule
\multirow{2}{*}{Kriging} & RMSE &  0.684 $\pm$ 0.07 &  \textbf{0.675 $\pm$ 0.07} &  \textbf{0.670 $\pm$ 0.05} &  \textbf{0.656 $\pm$ 0.04} &  \textbf{0.650 $\pm$ 0.04} &  \textbf{0.654 $\pm$ 0.03} &  \textbf{0.651 $\pm$ 0.03} \\
 & MI-RMSE &  0.284 $\pm$ 0.10 &  \textbf{0.271 $\pm$ 0.08} &  \textbf{0.253 $\pm$ 0.06} &  \textbf{0.239 $\pm$ 0.05} &  \textbf{0.214 $\pm$ 0.04} &  \textbf{0.188 $\pm$ 0.03} &  0.162 $\pm$ 0.03 \\
\midrule
\multirow{2}{*}{ML Base} & RMSE &  0.975 $\pm$ 0.20 &  0.957 $\pm$ 0.17 &  0.984 $\pm$ 0.18 &  0.985 $\pm$ 0.17 &  0.971 $\pm$ 0.15 &  0.994 $\pm$ 0.17 &  0.985 $\pm$ 0.17 \\
 & MI-RMSE &  \textbf{0.256 $\pm$ 0.14} &  0.277 $\pm$ 0.13 &  0.262 $\pm$ 0.10 &  0.269 $\pm$ 0.09 &  0.261 $\pm$ 0.09 &  0.248 $\pm$ 0.08 &  0.213 $\pm$ 0.06 \\
\midrule
\multirow{2}{*}{ML VSL} & RMSE &  \textbf{0.673 $\pm$ 0.07} &  0.680 $\pm$ 0.06 &  0.672 $\pm$ 0.05 &  0.659 $\pm$ 0.04 &  0.655 $\pm$ 0.04 &  0.663 $\pm$ 0.05 &  0.660 $\pm$ 0.04 \\
 & MI-RMSE &  0.381 $\pm$ 0.08 &  0.339 $\pm$ 0.07 &  0.301 $\pm$ 0.05 &  0.263 $\pm$ 0.05 &  0.226 $\pm$ 0.04 &  0.189 $\pm$ 0.03 &  \textbf{0.151 $\pm$ 0.03} \\
\bottomrule
\end{tabular}
\end{table}

\end{landscape}

\subsection{Non-stationary and Anisotropic Fields}
\label{ap:subsec:non-stationary}

\begin{figure}[t]
\centering
\includegraphics[width=\textwidth]{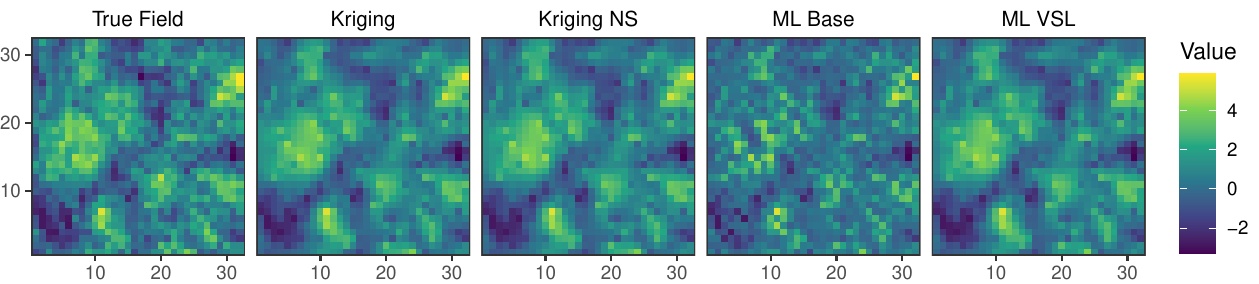}
\caption{Anisotropic \textit{Matérn} --- Visualisation of spatial interpolation methods for a representative realisation of an anisotropic Gaussian random field with \textit{Matérn} covariance. The panels show the True field alongside predictions from Ordinary Kriging, Ordinary Kriging with non-stationary covariance, ML Base, and the proposed ML VSL method.}
\label{fig:ns-aniso-tilt-20-phi-20-run_true_ok_krige-uns_ml-001_ml-004_002_plot}
\end{figure}

\paragraph{Intermediate Setting} Prior to the full non-stationary and anisotropic setting reported in \Cref{subsec:non-stationary} of the main text, a simpler intermediate case was considered in which only anisotropy was introduced (\Cref{fig:ns-aniso-tilt-20-phi-20-run_true_ok_krige-uns_ml-001_ml-004_002_plot}).
This preliminary experiment served to isolate the effect of directional dependence from that of spatially varying correlation structure, and its results are included here for completeness.

Moving beyond stationarity and isotropy, we report results for two progressively complex settings, first anisotropy alone, then the full non-stationary and anisotropic case described in \Cref{subsec:non-stationary} of the main text.
Spatial fields are simulated from a non-stationary Matérn covariance function (\ref{eq:cov_ns}) with smoothness parameter $\nu=1/2$ and variance $\sigma^2=1$, with spatially varying range $\phi$, anisotropy, and directional tilt.
Kriging predictions are obtained under both a misspecified stationary Exponential covariance model and the correctly specified non-stationary model, with known parameters in both cases.

\Crefrange{tab:complete_ns-aniso-tilt_10_phi}{tab:complete_ns-aniso-tilt_80_phi} report the full results across all considered values of $\phi$.
The qualitative behaviour is consistent across range parameters where non-stationary Kriging attains the lowest prediction error when correctly specified, while the proposed ML VSL approach remains competitive throughout, outperforming stationary Kriging across most sparsity levels.
These results confirm that the partial convolutional architecture is capable of implicitly adapting to spatially heterogeneous dependence structures without requiring explicit covariance model specification.

Across all range parameters, except at the sparsest case, Kriging NS consistently attains the lowest prediction error, as expected given correct covariance specification.
The performance gap between Kriging NS and ML VSL is modest throughout, with ML VSL remaining competitive despite relying on neither covariance model.
Both Kriging variants and ML VSL improve substantially as $\phi$ increases, reflecting the greater regularity of longer-range fields.
ML Base performs considerably worse across all settings, with RMSE values roughly twice those of the other methods, underscoring the importance of the spatial inductive biases embedded in the partial convolutional architecture.

Notably, the performance gap between Kriging NS and ML VSL widens slightly at shorter range parameters, where the anisotropic dependence structure is more spatially heterogeneous and harder to capture implicitly.
Nevertheless, ML VSL consistently outperforms stationary Kriging across most $\phi$ values and sparsity levels, suggesting that the network adapts more effectively to local variations in correlation structure than a globally misspecified linear predictor.

\newgeometry{top=20mm,left=20mm,right=20mm}
\fancyfootoffset{0pt}
\begin{landscape}
\begin{table}[tb]
\centering
\caption{Anisotropic \textit{Matérn} --- Comparison of model performance across varying percentages of known points (20\%, 30\%, 40\%, 50\%, 60\%, 70\% and 80\%). Values represent the mean ($\pm$ standard deviation) computed over 100 independent runs. The spatial interpolation was conducted using a non-stationary \textit{Matérn} covariogram with a range parameter of 10\% of the grid size. The underlying data follows a \textit{Gaussian} distribution.}
\label{tab:complete_ns-aniso-tilt_10_phi}
\begin{tabular}{llccccccc}
\toprule
\textbf{Model} & \textbf{Metric}  & \textbf{20\%} & \textbf{30\%} & \textbf{40\%} & \textbf{50\%} & \textbf{60\%} & \textbf{70\%} & \textbf{80\%}  \\
\midrule
\multirow{2}{*}{Kriging} & RMSE &  \textbf{1.390 $\pm$ 0.13} &  1.330 $\pm$ 0.12 &  1.316 $\pm$ 0.10 &  1.268 $\pm$ 0.08 &  1.235 $\pm$ 0.07 &  1.212 $\pm$ 0.06 &  1.200 $\pm$ 0.05 \\
 & MI-RMSE &  0.246 $\pm$ 0.07 &  0.236 $\pm$ 0.04 &  0.208 $\pm$ 0.03 &  0.187 $\pm$ 0.03 &  0.165 $\pm$ 0.02 &  0.140 $\pm$ 0.02 &  0.122 $\pm$ 0.02 \\
\midrule
\multirow{2}{*}{Kriging NS} & RMSE &  1.423 $\pm$ 0.15 &  \textbf{1.320 $\pm$ 0.12} &  \textbf{1.303 $\pm$ 0.10} &  \textbf{1.255 $\pm$ 0.08} &  \textbf{1.222 $\pm$ 0.07} &  \textbf{1.194 $\pm$ 0.06} &  \textbf{1.177 $\pm$ 0.05} \\
 & MI-RMSE &  \textbf{0.223 $\pm$ 0.07} &  \textbf{0.224 $\pm$ 0.04} &  \textbf{0.201 $\pm$ 0.03} &  \textbf{0.181 $\pm$ 0.03} &  \textbf{0.158 $\pm$ 0.02} &  0.135 $\pm$ 0.02 &  0.116 $\pm$ 0.02 \\
\midrule
\multirow{2}{*}{ML Base} & RMSE &  1.643 $\pm$ 0.15 &  1.636 $\pm$ 0.15 &  1.652 $\pm$ 0.13 &  1.650 $\pm$ 0.10 &  1.642 $\pm$ 0.10 &  1.643 $\pm$ 0.10 &  1.634 $\pm$ 0.10 \\
 & MI-RMSE &  0.383 $\pm$ 0.08 &  0.390 $\pm$ 0.06 &  0.357 $\pm$ 0.04 &  0.327 $\pm$ 0.03 &  0.303 $\pm$ 0.03 &  0.262 $\pm$ 0.03 &  0.214 $\pm$ 0.03 \\
\midrule
\multirow{2}{*}{ML VSL} & RMSE &  1.410 $\pm$ 0.14 &  1.350 $\pm$ 0.12 &  1.331 $\pm$ 0.10 &  1.280 $\pm$ 0.09 &  1.249 $\pm$ 0.07 &  1.226 $\pm$ 0.06 &  1.212 $\pm$ 0.06 \\
 & MI-RMSE &  0.314 $\pm$ 0.03 &  0.267 $\pm$ 0.03 &  0.226 $\pm$ 0.03 &  0.192 $\pm$ 0.02 &  0.162 $\pm$ 0.02 &  \textbf{0.132 $\pm$ 0.02} &  \textbf{0.109 $\pm$ 0.02} \\
\bottomrule
\end{tabular}
\end{table}

\begin{table}[tb]
\centering
\caption{Anisotropic \textit{Matérn} --- Comparison of model performance across varying percentages of known points (20\%, 30\%, 40\%, 50\%, 60\%, 70\% and 80\%). Values represent the mean ($\pm$ standard deviation) computed over 100 independent runs. The spatial interpolation was conducted using a non-stationary \textit{Matérn} covariogram with a range parameter of 20\% of the grid size. The underlying data follows a \textit{Gaussian} distribution.}
\label{tab:complete_ns-aniso-tilt_20_phi}
\begin{tabular}{llccccccc}
\toprule
\textbf{Model} & \textbf{Metric}  & \textbf{20\%} & \textbf{30\%} & \textbf{40\%} & \textbf{50\%} & \textbf{60\%} & \textbf{70\%} & \textbf{80\%}  \\
\midrule
\multirow{2}{*}{Kriging} & RMSE &  \textbf{1.075 $\pm$ 0.11} &  1.025 $\pm$ 0.10 &  1.003 $\pm$ 0.08 &  0.949 $\pm$ 0.07 &  0.921 $\pm$ 0.06 &  0.895 $\pm$ 0.04 &  0.888 $\pm$ 0.04 \\
 & MI-RMSE &  0.188 $\pm$ 0.03 &  0.170 $\pm$ 0.02 &  0.150 $\pm$ 0.02 &  0.133 $\pm$ 0.02 &  0.117 $\pm$ 0.02 &  0.100 $\pm$ 0.01 &  0.087 $\pm$ 0.01 \\
\midrule
\multirow{2}{*}{Kriging NS} & RMSE &  1.078 $\pm$ 0.12 &  \textbf{1.011 $\pm$ 0.10} &  \textbf{0.985 $\pm$ 0.08} &  \textbf{0.936 $\pm$ 0.06} &  \textbf{0.908 $\pm$ 0.06} &  \textbf{0.878 $\pm$ 0.04} &  \textbf{0.867 $\pm$ 0.04} \\
 & MI-RMSE &  \textbf{0.178 $\pm$ 0.03} &  \textbf{0.163 $\pm$ 0.03} &  0.145 $\pm$ 0.02 &  0.128 $\pm$ 0.02 &  0.112 $\pm$ 0.02 &  0.096 $\pm$ 0.02 &  0.083 $\pm$ 0.01 \\
\midrule
\multirow{2}{*}{ML Base} & RMSE &  1.591 $\pm$ 0.19 &  1.580 $\pm$ 0.17 &  1.580 $\pm$ 0.17 &  1.591 $\pm$ 0.15 &  1.570 $\pm$ 0.14 &  1.565 $\pm$ 0.14 &  1.554 $\pm$ 0.12 \\
 & MI-RMSE &  0.481 $\pm$ 0.10 &  0.483 $\pm$ 0.07 &  0.421 $\pm$ 0.08 &  0.400 $\pm$ 0.06 &  0.365 $\pm$ 0.06 &  0.317 $\pm$ 0.05 &  0.262 $\pm$ 0.04 \\
\midrule
\multirow{2}{*}{ML VSL} & RMSE &  1.086 $\pm$ 0.11 &  1.042 $\pm$ 0.10 &  1.018 $\pm$ 0.09 &  0.960 $\pm$ 0.07 &  0.934 $\pm$ 0.06 &  0.911 $\pm$ 0.04 &  0.897 $\pm$ 0.04 \\
 & MI-RMSE &  0.191 $\pm$ 0.03 &  0.166 $\pm$ 0.03 &  \textbf{0.145 $\pm$ 0.02} &  \textbf{0.124 $\pm$ 0.02} &  \textbf{0.106 $\pm$ 0.02} &  \textbf{0.088 $\pm$ 0.01} &  \textbf{0.075 $\pm$ 0.01} \\
\bottomrule
\end{tabular}
\end{table}

\begin{table}[tb]
\centering
\caption{Anisotropic \textit{Matérn} --- Comparison of model performance across varying percentages of known points (20\%, 30\%, 40\%, 50\%, 60\%, 70\% and 80\%). Values represent the mean ($\pm$ standard deviation) computed over 100 independent runs. The spatial interpolation was conducted using a non-stationary \textit{Matérn} covariogram with a range parameter of 30\% of the grid size. The underlying data follows a \textit{Gaussian} distribution.}
\label{tab:complete_ns-aniso-tilt_30_phi}
\begin{tabular}{llccccccc}
\toprule
\textbf{Model} & \textbf{Metric}  & \textbf{20\%} & \textbf{30\%} & \textbf{40\%} & \textbf{50\%} & \textbf{60\%} & \textbf{70\%} & \textbf{80\%}  \\
\midrule
\multirow{2}{*}{Kriging} & RMSE &  0.897 $\pm$ 0.09 &  0.856 $\pm$ 0.08 &  0.834 $\pm$ 0.07 &  0.785 $\pm$ 0.06 &  0.761 $\pm$ 0.05 &  0.737 $\pm$ 0.04 &  0.732 $\pm$ 0.04 \\
 & MI-RMSE &  0.147 $\pm$ 0.03 &  0.132 $\pm$ 0.02 &  0.117 $\pm$ 0.02 &  0.103 $\pm$ 0.02 &  0.091 $\pm$ 0.02 &  0.078 $\pm$ 0.01 &  0.068 $\pm$ 0.01 \\
\midrule
\multirow{2}{*}{Kriging NS} & RMSE &  \textbf{0.893 $\pm$ 0.10} &  \textbf{0.844 $\pm$ 0.08} &  \textbf{0.819 $\pm$ 0.07} &  \textbf{0.773 $\pm$ 0.05} &  \textbf{0.749 $\pm$ 0.05} &  \textbf{0.722 $\pm$ 0.04} &  \textbf{0.715 $\pm$ 0.03} \\
 & MI-RMSE &  \textbf{0.141 $\pm$ 0.03} &  0.127 $\pm$ 0.03 &  0.113 $\pm$ 0.02 &  0.099 $\pm$ 0.02 &  0.087 $\pm$ 0.02 &  0.075 $\pm$ 0.01 &  0.065 $\pm$ 0.01 \\
\midrule
\multirow{2}{*}{ML Base} & RMSE &  1.527 $\pm$ 0.23 &  1.509 $\pm$ 0.20 &  1.512 $\pm$ 0.20 &  1.509 $\pm$ 0.17 &  1.456 $\pm$ 0.17 &  1.467 $\pm$ 0.18 &  1.451 $\pm$ 0.17 \\
 & MI-RMSE &  0.484 $\pm$ 0.13 &  0.476 $\pm$ 0.11 &  0.419 $\pm$ 0.11 &  0.395 $\pm$ 0.08 &  0.340 $\pm$ 0.10 &  0.309 $\pm$ 0.08 &  0.256 $\pm$ 0.07 \\
\midrule
\multirow{2}{*}{ML VSL} & RMSE &  0.911 $\pm$ 0.09 &  0.879 $\pm$ 0.09 &  0.856 $\pm$ 0.08 &  0.801 $\pm$ 0.06 &  0.780 $\pm$ 0.05 &  0.760 $\pm$ 0.04 &  0.746 $\pm$ 0.04 \\
 & MI-RMSE &  0.143 $\pm$ 0.03 &  \textbf{0.125 $\pm$ 0.03} &  \textbf{0.110 $\pm$ 0.02} &  \textbf{0.094 $\pm$ 0.02} &  \textbf{0.081 $\pm$ 0.02} &  \textbf{0.068 $\pm$ 0.01} &  \textbf{0.057 $\pm$ 0.01} \\
\bottomrule
\end{tabular}
\end{table}

\begin{table}[tb]
\centering
\caption{Anisotropic \textit{Matérn} --- Comparison of model performance across varying percentages of known points (20\%, 30\%, 40\%, 50\%, 60\%, 70\% and 80\%). Values represent the mean ($\pm$ standard deviation) computed over 100 independent runs. The spatial interpolation was conducted using a non-stationary \textit{Matérn} covariogram with a range parameter of 40\% of the grid size. The underlying data follows a \textit{Gaussian} distribution.}
\label{tab:complete_ns-aniso-tilt_40_phi}
\begin{tabular}{llccccccc}
\toprule
\textbf{Model} & \textbf{Metric}  & \textbf{20\%} & \textbf{30\%} & \textbf{40\%} & \textbf{50\%} & \textbf{60\%} & \textbf{70\%} & \textbf{80\%}  \\
\midrule
\multirow{2}{*}{Kriging} & RMSE &  0.786 $\pm$ 0.08 &  0.748 $\pm$ 0.07 &  0.728 $\pm$ 0.06 &  0.683 $\pm$ 0.05 &  0.662 $\pm$ 0.04 &  0.641 $\pm$ 0.03 &  0.637 $\pm$ 0.03 \\
 & MI-RMSE &  0.124 $\pm$ 0.03 &  0.111 $\pm$ 0.02 &  0.099 $\pm$ 0.02 &  0.087 $\pm$ 0.02 &  0.076 $\pm$ 0.02 &  0.066 $\pm$ 0.01 &  0.057 $\pm$ 0.01 \\
\midrule
\multirow{2}{*}{Kriging NS} & RMSE &  \textbf{0.785 $\pm$ 0.11} &  \textbf{0.738 $\pm$ 0.07} &  \textbf{0.714 $\pm$ 0.06} &  \textbf{0.672 $\pm$ 0.05} &  \textbf{0.651 $\pm$ 0.04} &  \textbf{0.627 $\pm$ 0.03} &  \textbf{0.621 $\pm$ 0.03} \\
 & MI-RMSE &  0.122 $\pm$ 0.04 &  0.107 $\pm$ 0.03 &  0.095 $\pm$ 0.02 &  0.083 $\pm$ 0.02 &  0.073 $\pm$ 0.02 &  0.063 $\pm$ 0.01 &  0.055 $\pm$ 0.01 \\
\midrule
\multirow{2}{*}{ML Base} & RMSE &  1.459 $\pm$ 0.24 &  1.406 $\pm$ 0.21 &  1.415 $\pm$ 0.21 &  1.425 $\pm$ 0.17 &  1.359 $\pm$ 0.17 &  1.361 $\pm$ 0.20 &  1.334 $\pm$ 0.19 \\
 & MI-RMSE &  0.466 $\pm$ 0.14 &  0.426 $\pm$ 0.15 &  0.383 $\pm$ 0.14 &  0.369 $\pm$ 0.11 &  0.317 $\pm$ 0.12 &  0.282 $\pm$ 0.10 &  0.235 $\pm$ 0.09 \\
\midrule
\multirow{2}{*}{ML VSL} & RMSE &  0.802 $\pm$ 0.08 &  0.779 $\pm$ 0.08 &  0.758 $\pm$ 0.07 &  0.704 $\pm$ 0.05 &  0.687 $\pm$ 0.04 &  0.670 $\pm$ 0.04 &  0.654 $\pm$ 0.04 \\
 & MI-RMSE &  \textbf{0.118 $\pm$ 0.03} &  \textbf{0.103 $\pm$ 0.03} &  \textbf{0.091 $\pm$ 0.02} &  \textbf{0.078 $\pm$ 0.02} &  \textbf{0.067 $\pm$ 0.02} &  \textbf{0.056 $\pm$ 0.01} &  \textbf{0.047 $\pm$ 0.01} \\
\bottomrule
\end{tabular}
\end{table}

\begin{table}[tb]
\centering
\caption{Anisotropic \textit{Matérn} --- Comparison of model performance across varying percentages of known points (20\%, 30\%, 40\%, 50\%, 60\%, 70\% and 80\%). Values represent the mean ($\pm$ standard deviation) computed over 100 independent runs. The spatial interpolation was conducted using a non-stationary \textit{Matérn} covariogram with a range parameter of 50\% of the grid size. The underlying data follows a \textit{Gaussian} distribution.}
\label{tab:complete_ns-aniso-tilt_50_phi}
\begin{tabular}{llccccccc}
\toprule
\textbf{Model} & \textbf{Metric}  & \textbf{20\%} & \textbf{30\%} & \textbf{40\%} & \textbf{50\%} & \textbf{60\%} & \textbf{70\%} & \textbf{80\%}  \\
\midrule
\multirow{2}{*}{Kriging} & RMSE &  0.703 $\pm$ 0.07 &  0.674 $\pm$ 0.06 &  0.653 $\pm$ 0.06 &  0.612 $\pm$ 0.04 &  0.593 $\pm$ 0.04 &  0.574 $\pm$ 0.03 &  0.570 $\pm$ 0.03 \\
 & MI-RMSE &  0.110 $\pm$ 0.03 &  0.098 $\pm$ 0.02 &  0.087 $\pm$ 0.02 &  0.077 $\pm$ 0.02 &  0.068 $\pm$ 0.02 &  0.058 $\pm$ 0.01 &  0.051 $\pm$ 0.01 \\
\midrule
\multirow{2}{*}{Kriging NS} & RMSE &  \textbf{0.697 $\pm$ 0.08} &  \textbf{0.664 $\pm$ 0.06} &  \textbf{0.642 $\pm$ 0.06} &  \textbf{0.602 $\pm$ 0.04} &  \textbf{0.583 $\pm$ 0.04} &  \textbf{0.562 $\pm$ 0.03} &  \textbf{0.556 $\pm$ 0.03} \\
 & MI-RMSE &  0.106 $\pm$ 0.03 &  0.094 $\pm$ 0.02 &  0.084 $\pm$ 0.02 &  0.074 $\pm$ 0.02 &  0.065 $\pm$ 0.02 &  0.056 $\pm$ 0.01 &  0.049 $\pm$ 0.01 \\
\midrule
\multirow{2}{*}{ML Base} & RMSE &  1.380 $\pm$ 0.22 &  1.335 $\pm$ 0.21 &  1.328 $\pm$ 0.22 &  1.297 $\pm$ 0.18 &  1.259 $\pm$ 0.18 &  1.228 $\pm$ 0.22 &  1.224 $\pm$ 0.21 \\
 & MI-RMSE &  0.447 $\pm$ 0.15 &  0.396 $\pm$ 0.16 &  0.348 $\pm$ 0.17 &  0.318 $\pm$ 0.14 &  0.287 $\pm$ 0.13 &  0.242 $\pm$ 0.12 &  0.213 $\pm$ 0.10 \\
\midrule
\multirow{2}{*}{ML VSL} & RMSE &  0.721 $\pm$ 0.08 &  0.711 $\pm$ 0.07 &  0.690 $\pm$ 0.07 &  0.636 $\pm$ 0.05 &  0.622 $\pm$ 0.04 &  0.609 $\pm$ 0.04 &  0.591 $\pm$ 0.03 \\
 & MI-RMSE &  \textbf{0.103 $\pm$ 0.03} &  \textbf{0.089 $\pm$ 0.02} &  \textbf{0.080 $\pm$ 0.02} &  \textbf{0.068 $\pm$ 0.02} &  \textbf{0.059 $\pm$ 0.01} &  \textbf{0.049 $\pm$ 0.01} &  \textbf{0.041 $\pm$ 0.01} \\
\bottomrule
\end{tabular}
\end{table}

\begin{table}[tb]
\centering
\caption{Anisotropic \textit{Matérn} --- Comparison of model performance across varying percentages of known points (20\%, 30\%, 40\%, 50\%, 60\%, 70\% and 80\%). Values represent the mean ($\pm$ standard deviation) computed over 100 independent runs. The spatial interpolation was conducted using a non-stationary \textit{Matérn} covariogram with a range parameter of 60\% of the grid size. The underlying data follows a \textit{Gaussian} distribution.}
\label{tab:complete_ns-aniso-tilt_60_phi}
\begin{tabular}{llccccccc}
\toprule
\textbf{Model} & \textbf{Metric}  & \textbf{20\%} & \textbf{30\%} & \textbf{40\%} & \textbf{50\%} & \textbf{60\%} & \textbf{70\%} & \textbf{80\%}  \\
\midrule
\multirow{2}{*}{Kriging} & RMSE &  0.643 $\pm$ 0.07 &  0.617 $\pm$ 0.06 &  0.596 $\pm$ 0.05 &  0.559 $\pm$ 0.04 &  0.541 $\pm$ 0.04 &  0.524 $\pm$ 0.03 &  0.521 $\pm$ 0.03 \\
 & MI-RMSE &  0.101 $\pm$ 0.02 &  0.090 $\pm$ 0.02 &  0.080 $\pm$ 0.02 &  0.071 $\pm$ 0.02 &  0.062 $\pm$ 0.02 &  0.054 $\pm$ 0.01 &  0.047 $\pm$ 0.01 \\
\midrule
\multirow{2}{*}{Kriging NS} & RMSE &  \textbf{0.637 $\pm$ 0.07} &  \textbf{0.608 $\pm$ 0.06} &  \textbf{0.586 $\pm$ 0.05} &  \textbf{0.551 $\pm$ 0.04} &  \textbf{0.531 $\pm$ 0.04} &  \textbf{0.513 $\pm$ 0.03} &  \textbf{0.508 $\pm$ 0.03} \\
 & MI-RMSE &  0.097 $\pm$ 0.03 &  0.086 $\pm$ 0.02 &  0.077 $\pm$ 0.02 &  0.068 $\pm$ 0.02 &  0.060 $\pm$ 0.02 &  0.051 $\pm$ 0.01 &  0.045 $\pm$ 0.01 \\
\midrule
\multirow{2}{*}{ML Base} & RMSE &  1.295 $\pm$ 0.21 &  1.265 $\pm$ 0.21 &  1.275 $\pm$ 0.23 &  1.212 $\pm$ 0.19 &  1.177 $\pm$ 0.20 &  1.159 $\pm$ 0.22 &  1.119 $\pm$ 0.21 \\
 & MI-RMSE &  0.414 $\pm$ 0.17 &  0.393 $\pm$ 0.17 &  0.352 $\pm$ 0.17 &  0.291 $\pm$ 0.15 &  0.266 $\pm$ 0.13 &  0.240 $\pm$ 0.13 &  0.189 $\pm$ 0.11 \\
\midrule
\multirow{2}{*}{ML VSL} & RMSE &  0.663 $\pm$ 0.07 &  0.659 $\pm$ 0.07 &  0.638 $\pm$ 0.07 &  0.587 $\pm$ 0.04 &  0.573 $\pm$ 0.04 &  0.563 $\pm$ 0.04 &  0.544 $\pm$ 0.03 \\
 & MI-RMSE &  \textbf{0.093 $\pm$ 0.03} &  \textbf{0.081 $\pm$ 0.02} &  \textbf{0.072 $\pm$ 0.02} &  \textbf{0.061 $\pm$ 0.02} &  \textbf{0.053 $\pm$ 0.01} &  \textbf{0.044 $\pm$ 0.01} &  \textbf{0.037 $\pm$ 0.01} \\
\bottomrule
\end{tabular}
\end{table}

\begin{table}[tb]
\centering
\caption{Anisotropic \textit{Matérn} --- Comparison of model performance across varying percentages of known points (20\%, 30\%, 40\%, 50\%, 60\%, 70\% and 80\%). Values represent the mean ($\pm$ standard deviation) computed over 100 independent runs. The spatial interpolation was conducted using a non-stationary \textit{Matérn} covariogram with a range parameter of 70\% of the grid size. The underlying data follows a \textit{Gaussian} distribution.}
\label{tab:complete_ns-aniso-tilt_70_phi}
\begin{tabular}{llccccccc}
\toprule
\textbf{Model} & \textbf{Metric}  & \textbf{20\%} & \textbf{30\%} & \textbf{40\%} & \textbf{50\%} & \textbf{60\%} & \textbf{70\%} & \textbf{80\%}  \\
\midrule
\multirow{2}{*}{Kriging} & RMSE &  0.596 $\pm$ 0.06 &  0.573 $\pm$ 0.05 &  0.552 $\pm$ 0.05 &  0.519 $\pm$ 0.04 &  0.501 $\pm$ 0.03 &  0.485 $\pm$ 0.02 &  0.481 $\pm$ 0.03 \\
 & MI-RMSE &  0.093 $\pm$ 0.02 &  0.085 $\pm$ 0.02 &  0.075 $\pm$ 0.02 &  0.066 $\pm$ 0.02 &  0.059 $\pm$ 0.02 &  0.050 $\pm$ 0.01 &  0.044 $\pm$ 0.01 \\
\midrule
\multirow{2}{*}{Kriging NS} & RMSE &  \textbf{0.590 $\pm$ 0.07} &  \textbf{0.565 $\pm$ 0.05} &  \textbf{0.542 $\pm$ 0.05} &  \textbf{0.511 $\pm$ 0.03} &  \textbf{0.492 $\pm$ 0.03} &  \textbf{0.474 $\pm$ 0.02} &  \textbf{0.469 $\pm$ 0.02} \\
 & MI-RMSE &  0.090 $\pm$ 0.03 &  0.082 $\pm$ 0.02 &  0.072 $\pm$ 0.02 &  0.063 $\pm$ 0.02 &  0.056 $\pm$ 0.02 &  0.048 $\pm$ 0.01 &  0.042 $\pm$ 0.01 \\
\midrule
\multirow{2}{*}{ML Base} & RMSE &  1.251 $\pm$ 0.21 &  1.204 $\pm$ 0.20 &  1.181 $\pm$ 0.22 &  1.159 $\pm$ 0.21 &  1.104 $\pm$ 0.20 &  1.073 $\pm$ 0.22 &  1.032 $\pm$ 0.21 \\
 & MI-RMSE &  0.400 $\pm$ 0.18 &  0.368 $\pm$ 0.17 &  0.302 $\pm$ 0.17 &  0.289 $\pm$ 0.16 &  0.251 $\pm$ 0.15 &  0.216 $\pm$ 0.12 &  0.171 $\pm$ 0.11 \\
\midrule
\multirow{2}{*}{ML VSL} & RMSE &  0.618 $\pm$ 0.07 &  0.618 $\pm$ 0.07 &  0.599 $\pm$ 0.07 &  0.549 $\pm$ 0.04 &  0.536 $\pm$ 0.04 &  0.525 $\pm$ 0.04 &  0.506 $\pm$ 0.03 \\
 & MI-RMSE &  \textbf{0.085 $\pm$ 0.03} &  \textbf{0.076 $\pm$ 0.02} &  \textbf{0.067 $\pm$ 0.02} &  \textbf{0.057 $\pm$ 0.02} &  \textbf{0.049 $\pm$ 0.01} &  \textbf{0.041 $\pm$ 0.01} &  \textbf{0.034 $\pm$ 0.01} \\
\bottomrule
\end{tabular}
\end{table}

\begin{table}[tb]
\centering
\caption{Anisotropic \textit{Matérn} --- Comparison of model performance across varying percentages of known points (20\%, 30\%, 40\%, 50\%, 60\%, 70\% and 80\%). Values represent the mean ($\pm$ standard deviation) computed over 100 independent runs. The spatial interpolation was conducted using a non-stationary \textit{Matérn} covariogram with a range parameter of 80\% of the grid size. The underlying data follows a \textit{Gaussian} distribution.}
\label{tab:complete_ns-aniso-tilt_80_phi}
\begin{tabular}{llccccccc}
\toprule
\textbf{Model} & \textbf{Metric}  & \textbf{20\%} & \textbf{30\%} & \textbf{40\%} & \textbf{50\%} & \textbf{60\%} & \textbf{70\%} & \textbf{80\%}  \\
\midrule
\multirow{2}{*}{Kriging} & RMSE &  0.560 $\pm$ 0.06 &  0.537 $\pm$ 0.05 &  0.517 $\pm$ 0.04 &  0.486 $\pm$ 0.03 &  0.468 $\pm$ 0.03 &  0.454 $\pm$ 0.02 &  0.450 $\pm$ 0.02 \\
 & MI-RMSE &  0.090 $\pm$ 0.02 &  0.081 $\pm$ 0.02 &  0.071 $\pm$ 0.02 &  0.063 $\pm$ 0.02 &  0.056 $\pm$ 0.01 &  0.049 $\pm$ 0.01 &  0.043 $\pm$ 0.01 \\
\midrule
\multirow{2}{*}{Kriging NS} & RMSE &  \textbf{0.554 $\pm$ 0.06} &  \textbf{0.531 $\pm$ 0.05} &  \textbf{0.508 $\pm$ 0.04} &  \textbf{0.478 $\pm$ 0.03} &  \textbf{0.460 $\pm$ 0.03} &  \textbf{0.444 $\pm$ 0.02} &  \textbf{0.439 $\pm$ 0.02} \\
 & MI-RMSE &  0.087 $\pm$ 0.02 &  0.077 $\pm$ 0.02 &  0.068 $\pm$ 0.02 &  0.061 $\pm$ 0.02 &  0.053 $\pm$ 0.01 &  0.047 $\pm$ 0.01 &  0.041 $\pm$ 0.01 \\
\midrule
\multirow{2}{*}{ML Base} & RMSE &  1.187 $\pm$ 0.21 &  1.142 $\pm$ 0.19 &  1.120 $\pm$ 0.21 &  1.099 $\pm$ 0.18 &  1.043 $\pm$ 0.20 &  1.021 $\pm$ 0.21 &  0.976 $\pm$ 0.20 \\
 & MI-RMSE &  0.385 $\pm$ 0.18 &  0.355 $\pm$ 0.19 &  0.281 $\pm$ 0.17 &  0.271 $\pm$ 0.15 &  0.236 $\pm$ 0.15 &  0.212 $\pm$ 0.13 &  0.168 $\pm$ 0.11 \\
\midrule
\multirow{2}{*}{ML VSL} & RMSE &  0.583 $\pm$ 0.06 &  0.586 $\pm$ 0.07 &  0.567 $\pm$ 0.06 &  0.519 $\pm$ 0.04 &  0.505 $\pm$ 0.04 &  0.497 $\pm$ 0.04 &  0.477 $\pm$ 0.03 \\
 & MI-RMSE &  \textbf{0.081 $\pm$ 0.02} &  \textbf{0.071 $\pm$ 0.02} &  \textbf{0.062 $\pm$ 0.02} &  \textbf{0.053 $\pm$ 0.02} &  \textbf{0.046 $\pm$ 0.01} &  \textbf{0.038 $\pm$ 0.01} &  \textbf{0.033 $\pm$ 0.01} \\
\bottomrule
\end{tabular}
\end{table}

\end{landscape}
\restoregeometry

Following the anisotropic setting, we next present the complete results for the non-stationary and anisotropic case in which an additional spatially varying mean component is introduced, as described in the main text.
The covariance structure remains identical to the preceding experiment, with all parameters assumed known, while the mean function introduces an additional source of non-stationarity.
\Crefrange{tab:complete_ns-aniso-tilt-mean_10_phi}{tab:complete_ns-aniso-tilt-mean_80_phi} report the full results across all considered values of $\phi$.

The results are broadly consistent with those observed in the preceding anisotropic setting.
Kriging NS attains the lowest RMSE across virtually all $\phi$ values and sparsity levels, with stationary Kriging performing comparably or only marginally worse.
ML VSL remains competitive throughout, tracking closely the performance of both Kriging variants and substantially outperforming ML Base across all settings.

Compared to the anisotropic only experiment, the introduction of a spatially varying mean component does not materially alter the relative ordering of methods, nor does it widen the gap between ML VSL and the Kriging approaches.
This stability suggests that the partial convolutional architecture is able to implicitly absorb low-frequency mean variation as part of its learned spatial representation, without requiring an explicit trend model.
ML Base continues to perform substantially worse, with RMSE values approximately twice those of the remaining methods, further reinforcing the conclusion that architectural design and loss function choices are the primary drivers of performance in this class of interpolation problems.

As the non-stationary and anisotropic setting subsumes the anisotropy only case and provides a more comprehensive characterisation of spatially heterogeneous dependence structures, it was selected as the representative experiment reported in the main text.

\begin{figure}[t]
\centering
\includegraphics[width=\textwidth]{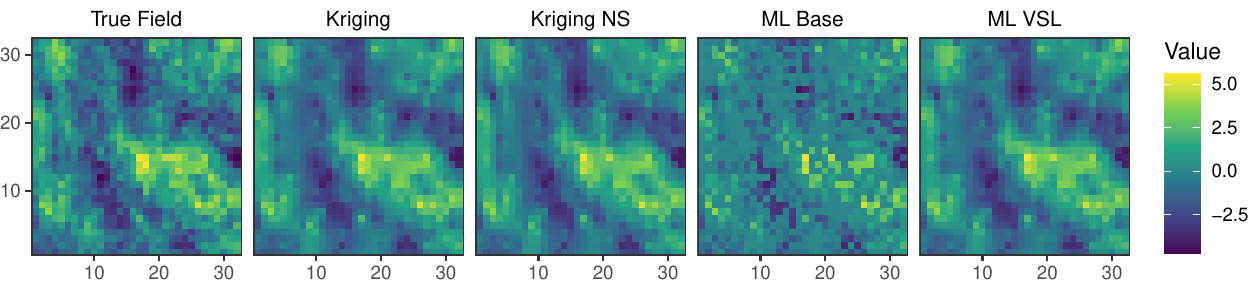}
\caption{Non-stationary \textit{Matérn} --- Visualisation of spatial interpolation methods for a representative realisation of a non-stationary, anisotropic Gaussian random field with \textit{Matérn} covariance. The panels show the True field alongside predictions from Ordinary Kriging, Ordinary Kriging with non-stationary covariance, ML Base, and the proposed ML VSL method.}
\label{fig:ns-aniso-tilt-mean-20-phi-20-run_true_ok_krige-uns_ml-001_ml-004_003_plot}
\end{figure}

\newgeometry{top=20mm,left=20mm,right=20mm}
\fancyfootoffset{0pt}
\begin{landscape}
\begin{table}[tb]
\centering
\caption{Non-stationary \textit{Matérn} --- Comparison of model performance across varying percentages of known points (20\%, 30\%, 40\%, 50\%, 60\%, 70\% and 80\%). Values represent the mean ($\pm$ standard deviation) computed over 100 independent runs. The spatial interpolation was conducted using a non-stationary \textit{Matérn} covariogram with a range parameter of 10\% of the grid size. The underlying data follows a \textit{Gaussian} distribution.}
\label{tab:complete_ns-aniso-tilt-mean_10_phi}
\begin{tabular}{llccccccc}
\toprule
\textbf{Model} & \textbf{Metric}  & \textbf{20\%} & \textbf{30\%} & \textbf{40\%} & \textbf{50\%} & \textbf{60\%} & \textbf{70\%} & \textbf{80\%}  \\
\midrule
\multirow{2}{*}{Kriging} & RMSE &  \textbf{1.398 $\pm$ 0.14} &  1.338 $\pm$ 0.12 &  1.320 $\pm$ 0.10 &  1.270 $\pm$ 0.08 &  1.238 $\pm$ 0.07 &  1.214 $\pm$ 0.06 &  1.202 $\pm$ 0.05 \\
 & MI-RMSE &  0.239 $\pm$ 0.06 &  0.227 $\pm$ 0.04 &  0.203 $\pm$ 0.03 &  0.182 $\pm$ 0.02 &  0.161 $\pm$ 0.02 &  0.137 $\pm$ 0.02 &  0.119 $\pm$ 0.01 \\
\midrule
\multirow{2}{*}{Kriging NS} & RMSE &  1.426 $\pm$ 0.15 &  \textbf{1.325 $\pm$ 0.12} &  \textbf{1.306 $\pm$ 0.10} &  \textbf{1.256 $\pm$ 0.08} &  \textbf{1.222 $\pm$ 0.07} &  \textbf{1.195 $\pm$ 0.06} &  \textbf{1.177 $\pm$ 0.05} \\
 & MI-RMSE &  \textbf{0.227 $\pm$ 0.06} &  \textbf{0.222 $\pm$ 0.04} &  \textbf{0.198 $\pm$ 0.03} &  \textbf{0.178 $\pm$ 0.02} &  \textbf{0.156 $\pm$ 0.02} &  0.133 $\pm$ 0.02 &  0.114 $\pm$ 0.02 \\
\midrule
\multirow{2}{*}{ML Base} & RMSE &  1.686 $\pm$ 0.15 &  1.676 $\pm$ 0.16 &  1.704 $\pm$ 0.14 &  1.697 $\pm$ 0.10 &  1.684 $\pm$ 0.11 &  1.686 $\pm$ 0.11 &  1.688 $\pm$ 0.10 \\
 & MI-RMSE &  0.407 $\pm$ 0.08 &  0.399 $\pm$ 0.06 &  0.371 $\pm$ 0.04 &  0.338 $\pm$ 0.04 &  0.309 $\pm$ 0.04 &  0.268 $\pm$ 0.03 &  0.224 $\pm$ 0.03 \\
\midrule
\multirow{2}{*}{ML VSL} & RMSE &  1.411 $\pm$ 0.14 &  1.359 $\pm$ 0.12 &  1.335 $\pm$ 0.10 &  1.283 $\pm$ 0.09 &  1.253 $\pm$ 0.07 &  1.229 $\pm$ 0.06 &  1.214 $\pm$ 0.06 \\
 & MI-RMSE &  0.298 $\pm$ 0.03 &  0.256 $\pm$ 0.03 &  0.218 $\pm$ 0.03 &  0.186 $\pm$ 0.02 &  0.157 $\pm$ 0.02 &  \textbf{0.128 $\pm$ 0.02} &  \textbf{0.107 $\pm$ 0.02} \\
\bottomrule
\end{tabular}
\end{table}

\begin{table}[tb]
\centering
\caption{Non-stationary \textit{Matérn} --- Comparison of model performance across varying percentages of known points (20\%, 30\%, 40\%, 50\%, 60\%, 70\% and 80\%). Values represent the mean ($\pm$ standard deviation) computed over 100 independent runs. The spatial interpolation was conducted using a non-stationary \textit{Matérn} covariogram with a range parameter of 20\% of the grid size. The underlying data follows a \textit{Gaussian} distribution.}
\label{tab:complete_ns-aniso-tilt-mean_20_phi}
\begin{tabular}{llccccccc}
\toprule
\textbf{Model} & \textbf{Metric}  & \textbf{20\%} & \textbf{30\%} & \textbf{40\%} & \textbf{50\%} & \textbf{60\%} & \textbf{70\%} & \textbf{80\%}  \\
\midrule
\multirow{2}{*}{Kriging} & RMSE &  \textbf{1.076 $\pm$ 0.11} &  1.028 $\pm$ 0.10 &  1.004 $\pm$ 0.08 &  0.949 $\pm$ 0.07 &  0.922 $\pm$ 0.06 &  0.895 $\pm$ 0.04 &  0.889 $\pm$ 0.04 \\
 & MI-RMSE &  0.177 $\pm$ 0.03 &  0.161 $\pm$ 0.03 &  0.143 $\pm$ 0.02 &  0.127 $\pm$ 0.02 &  0.111 $\pm$ 0.02 &  0.095 $\pm$ 0.02 &  0.083 $\pm$ 0.01 \\
\midrule
\multirow{2}{*}{Kriging NS} & RMSE &  1.080 $\pm$ 0.12 &  \textbf{1.014 $\pm$ 0.09} &  \textbf{0.987 $\pm$ 0.08} &  \textbf{0.937 $\pm$ 0.06} &  \textbf{0.908 $\pm$ 0.06} &  \textbf{0.878 $\pm$ 0.04} &  \textbf{0.868 $\pm$ 0.04} \\
 & MI-RMSE &  \textbf{0.169 $\pm$ 0.03} &  \textbf{0.155 $\pm$ 0.03} &  0.138 $\pm$ 0.02 &  0.122 $\pm$ 0.02 &  0.107 $\pm$ 0.02 &  0.092 $\pm$ 0.02 &  0.079 $\pm$ 0.01 \\
\midrule
\multirow{2}{*}{ML Base} & RMSE &  1.638 $\pm$ 0.19 &  1.624 $\pm$ 0.19 &  1.637 $\pm$ 0.19 &  1.643 $\pm$ 0.16 &  1.616 $\pm$ 0.15 &  1.616 $\pm$ 0.17 &  1.612 $\pm$ 0.14 \\
 & MI-RMSE &  0.490 $\pm$ 0.09 &  0.479 $\pm$ 0.07 &  0.425 $\pm$ 0.07 &  0.408 $\pm$ 0.06 &  0.366 $\pm$ 0.05 &  0.321 $\pm$ 0.05 &  0.269 $\pm$ 0.04 \\
\midrule
\multirow{2}{*}{ML VSL} & RMSE &  1.089 $\pm$ 0.11 &  1.051 $\pm$ 0.10 &  1.024 $\pm$ 0.08 &  0.963 $\pm$ 0.07 &  0.939 $\pm$ 0.06 &  0.914 $\pm$ 0.05 &  0.901 $\pm$ 0.05 \\
 & MI-RMSE &  0.179 $\pm$ 0.03 &  0.156 $\pm$ 0.03 &  \textbf{0.137 $\pm$ 0.02} &  \textbf{0.118 $\pm$ 0.02} &  \textbf{0.101 $\pm$ 0.02} &  \textbf{0.084 $\pm$ 0.01} &  \textbf{0.071 $\pm$ 0.01} \\
\bottomrule
\end{tabular}
\end{table}

\begin{table}[tb]
\centering
\caption{Non-stationary \textit{Matérn} --- Comparison of model performance across varying percentages of known points (20\%, 30\%, 40\%, 50\%, 60\%, 70\% and 80\%). Values represent the mean ($\pm$ standard deviation) computed over 100 independent runs. The spatial interpolation was conducted using a non-stationary \textit{Matérn} covariogram with a range parameter of 30\% of the grid size. The underlying data follows a \textit{Gaussian} distribution.}
\label{tab:complete_ns-aniso-tilt-mean_30_phi}
\begin{tabular}{llccccccc}
\toprule
\textbf{Model} & \textbf{Metric}  & \textbf{20\%} & \textbf{30\%} & \textbf{40\%} & \textbf{50\%} & \textbf{60\%} & \textbf{70\%} & \textbf{80\%}  \\
\midrule
\multirow{2}{*}{Kriging} & RMSE &  0.898 $\pm$ 0.09 &  0.857 $\pm$ 0.08 &  0.835 $\pm$ 0.07 &  0.785 $\pm$ 0.06 &  0.762 $\pm$ 0.05 &  0.738 $\pm$ 0.04 &  0.733 $\pm$ 0.04 \\
 & MI-RMSE &  0.138 $\pm$ 0.03 &  0.124 $\pm$ 0.03 &  0.111 $\pm$ 0.02 &  0.098 $\pm$ 0.02 &  0.086 $\pm$ 0.02 &  0.074 $\pm$ 0.02 &  0.064 $\pm$ 0.01 \\
\midrule
\multirow{2}{*}{Kriging NS} & RMSE &  \textbf{0.895 $\pm$ 0.10} &  \textbf{0.846 $\pm$ 0.08} &  \textbf{0.820 $\pm$ 0.07} &  \textbf{0.773 $\pm$ 0.05} &  \textbf{0.749 $\pm$ 0.05} &  \textbf{0.722 $\pm$ 0.04} &  \textbf{0.715 $\pm$ 0.04} \\
 & MI-RMSE &  \textbf{0.132 $\pm$ 0.03} &  0.120 $\pm$ 0.03 &  0.106 $\pm$ 0.02 &  0.094 $\pm$ 0.02 &  0.082 $\pm$ 0.02 &  0.071 $\pm$ 0.02 &  0.061 $\pm$ 0.01 \\
\midrule
\multirow{2}{*}{ML Base} & RMSE &  1.564 $\pm$ 0.24 &  1.559 $\pm$ 0.21 &  1.557 $\pm$ 0.24 &  1.557 $\pm$ 0.19 &  1.502 $\pm$ 0.20 &  1.500 $\pm$ 0.19 &  1.489 $\pm$ 0.17 \\
 & MI-RMSE &  0.478 $\pm$ 0.13 &  0.479 $\pm$ 0.10 &  0.413 $\pm$ 0.11 &  0.395 $\pm$ 0.08 &  0.339 $\pm$ 0.10 &  0.304 $\pm$ 0.08 &  0.254 $\pm$ 0.07 \\
\midrule
\multirow{2}{*}{ML VSL} & RMSE &  0.914 $\pm$ 0.09 &  0.889 $\pm$ 0.09 &  0.864 $\pm$ 0.08 &  0.805 $\pm$ 0.06 &  0.786 $\pm$ 0.05 &  0.765 $\pm$ 0.04 &  0.750 $\pm$ 0.04 \\
 & MI-RMSE &  0.133 $\pm$ 0.03 &  \textbf{0.117 $\pm$ 0.03} &  \textbf{0.103 $\pm$ 0.02} &  \textbf{0.089 $\pm$ 0.02} &  \textbf{0.076 $\pm$ 0.02} &  \textbf{0.064 $\pm$ 0.01} &  \textbf{0.054 $\pm$ 0.01} \\
\bottomrule
\end{tabular}
\end{table}

\begin{table}[tb]
\centering
\caption{Non-stationary \textit{Matérn} --- Comparison of model performance across varying percentages of known points (20\%, 30\%, 40\%, 50\%, 60\%, 70\% and 80\%). Values represent the mean ($\pm$ standard deviation) computed over 100 independent runs. The spatial interpolation was conducted using a non-stationary \textit{Matérn} covariogram with a range parameter of 40\% of the grid size. The underlying data follows a \textit{Gaussian} distribution.}
\label{tab:complete_ns-aniso-tilt-mean_40_phi}
\begin{tabular}{llccccccc}
\toprule
\textbf{Model} & \textbf{Metric}  & \textbf{20\%} & \textbf{30\%} & \textbf{40\%} & \textbf{50\%} & \textbf{60\%} & \textbf{70\%} & \textbf{80\%}  \\
\midrule
\multirow{2}{*}{Kriging} & RMSE &  0.785 $\pm$ 0.08 &  0.750 $\pm$ 0.07 &  0.728 $\pm$ 0.06 &  0.683 $\pm$ 0.05 &  0.662 $\pm$ 0.04 &  0.641 $\pm$ 0.03 &  0.637 $\pm$ 0.03 \\
 & MI-RMSE &  0.116 $\pm$ 0.03 &  0.104 $\pm$ 0.03 &  0.093 $\pm$ 0.02 &  0.082 $\pm$ 0.02 &  0.072 $\pm$ 0.02 &  0.062 $\pm$ 0.02 &  0.054 $\pm$ 0.01 \\
\midrule
\multirow{2}{*}{Kriging NS} & RMSE &  \textbf{0.780 $\pm$ 0.09} &  \textbf{0.740 $\pm$ 0.07} &  \textbf{0.715 $\pm$ 0.06} &  \textbf{0.673 $\pm$ 0.05} &  \textbf{0.651 $\pm$ 0.04} &  \textbf{0.627 $\pm$ 0.03} &  \textbf{0.621 $\pm$ 0.03} \\
 & MI-RMSE &  0.111 $\pm$ 0.03 &  0.100 $\pm$ 0.03 &  0.089 $\pm$ 0.02 &  0.078 $\pm$ 0.02 &  0.069 $\pm$ 0.02 &  0.059 $\pm$ 0.01 &  0.051 $\pm$ 0.01 \\
\midrule
\multirow{2}{*}{ML Base} & RMSE &  1.497 $\pm$ 0.25 &  1.462 $\pm$ 0.23 &  1.459 $\pm$ 0.25 &  1.455 $\pm$ 0.20 &  1.394 $\pm$ 0.18 &  1.369 $\pm$ 0.21 &  1.357 $\pm$ 0.23 \\
 & MI-RMSE &  0.459 $\pm$ 0.15 &  0.438 $\pm$ 0.14 &  0.370 $\pm$ 0.14 &  0.357 $\pm$ 0.12 &  0.310 $\pm$ 0.11 &  0.268 $\pm$ 0.10 &  0.224 $\pm$ 0.10 \\
\midrule
\multirow{2}{*}{ML VSL} & RMSE &  0.805 $\pm$ 0.08 &  0.790 $\pm$ 0.08 &  0.767 $\pm$ 0.07 &  0.708 $\pm$ 0.05 &  0.694 $\pm$ 0.04 &  0.676 $\pm$ 0.04 &  0.659 $\pm$ 0.04 \\
 & MI-RMSE &  \textbf{0.109 $\pm$ 0.03} &  \textbf{0.096 $\pm$ 0.03} &  \textbf{0.085 $\pm$ 0.02} &  \textbf{0.073 $\pm$ 0.02} &  \textbf{0.063 $\pm$ 0.02} &  \textbf{0.053 $\pm$ 0.01} &  \textbf{0.044 $\pm$ 0.01} \\
\bottomrule
\end{tabular}
\end{table}

\begin{table}[tb]
\centering
\caption{Non-stationary \textit{Matérn} --- Comparison of model performance across varying percentages of known points (20\%, 30\%, 40\%, 50\%, 60\%, 70\% and 80\%). Values represent the mean ($\pm$ standard deviation) computed over 100 independent runs. The spatial interpolation was conducted using a non-stationary \textit{Matérn} covariogram with a range parameter of 50\% of the grid size. The underlying data follows a \textit{Gaussian} distribution.}
\label{tab:complete_ns-aniso-tilt-mean_50_phi}
\begin{tabular}{llccccccc}
\toprule
\textbf{Model} & \textbf{Metric}  & \textbf{20\%} & \textbf{30\%} & \textbf{40\%} & \textbf{50\%} & \textbf{60\%} & \textbf{70\%} & \textbf{80\%}  \\
\midrule
\multirow{2}{*}{Kriging} & RMSE &  0.703 $\pm$ 0.07 &  0.674 $\pm$ 0.07 &  0.652 $\pm$ 0.06 &  0.612 $\pm$ 0.04 &  0.592 $\pm$ 0.04 &  0.574 $\pm$ 0.03 &  0.570 $\pm$ 0.03 \\
 & MI-RMSE &  0.103 $\pm$ 0.03 &  0.092 $\pm$ 0.02 &  0.082 $\pm$ 0.02 &  0.073 $\pm$ 0.02 &  0.065 $\pm$ 0.02 &  0.055 $\pm$ 0.01 &  0.048 $\pm$ 0.01 \\
\midrule
\multirow{2}{*}{Kriging NS} & RMSE &  \textbf{0.697 $\pm$ 0.08} &  \textbf{0.665 $\pm$ 0.06} &  \textbf{0.641 $\pm$ 0.06} &  \textbf{0.603 $\pm$ 0.04} &  \textbf{0.582 $\pm$ 0.04} &  \textbf{0.562 $\pm$ 0.03} &  \textbf{0.556 $\pm$ 0.03} \\
 & MI-RMSE &  0.099 $\pm$ 0.03 &  0.089 $\pm$ 0.02 &  0.079 $\pm$ 0.02 &  0.070 $\pm$ 0.02 &  0.062 $\pm$ 0.02 &  0.053 $\pm$ 0.01 &  0.046 $\pm$ 0.01 \\
\midrule
\multirow{2}{*}{ML Base} & RMSE &  1.419 $\pm$ 0.24 &  1.379 $\pm$ 0.21 &  1.370 $\pm$ 0.25 &  1.320 $\pm$ 0.19 &  1.273 $\pm$ 0.20 &  1.238 $\pm$ 0.22 &  1.216 $\pm$ 0.24 \\
 & MI-RMSE &  0.440 $\pm$ 0.16 &  0.413 $\pm$ 0.16 &  0.343 $\pm$ 0.16 &  0.304 $\pm$ 0.15 &  0.278 $\pm$ 0.13 &  0.235 $\pm$ 0.13 &  0.198 $\pm$ 0.11 \\
\midrule
\multirow{2}{*}{ML VSL} & RMSE &  0.725 $\pm$ 0.08 &  0.722 $\pm$ 0.08 &  0.697 $\pm$ 0.07 &  0.641 $\pm$ 0.05 &  0.627 $\pm$ 0.04 &  0.614 $\pm$ 0.05 &  0.595 $\pm$ 0.04 \\
 & MI-RMSE &  \textbf{0.095 $\pm$ 0.03} &  \textbf{0.084 $\pm$ 0.02} &  \textbf{0.074 $\pm$ 0.02} &  \textbf{0.064 $\pm$ 0.02} &  \textbf{0.056 $\pm$ 0.02} &  \textbf{0.046 $\pm$ 0.01} &  \textbf{0.038 $\pm$ 0.01} \\
\bottomrule
\end{tabular}
\end{table}

\begin{table}[tb]
\centering
\caption{Non-stationary \textit{Matérn} --- Comparison of model performance across varying percentages of known points (20\%, 30\%, 40\%, 50\%, 60\%, 70\% and 80\%). Values represent the mean ($\pm$ standard deviation) computed over 100 independent runs. The spatial interpolation was conducted using a non-stationary \textit{Matérn} covariogram with a range parameter of 60\% of the grid size. The underlying data follows a \textit{Gaussian} distribution.}
\label{tab:complete_ns-aniso-tilt-mean_60_phi}
\begin{tabular}{llccccccc}
\toprule
\textbf{Model} & \textbf{Metric}  & \textbf{20\%} & \textbf{30\%} & \textbf{40\%} & \textbf{50\%} & \textbf{60\%} & \textbf{70\%} & \textbf{80\%}  \\
\midrule
\multirow{2}{*}{Kriging} & RMSE &  0.643 $\pm$ 0.06 &  0.618 $\pm$ 0.06 &  0.595 $\pm$ 0.05 &  0.560 $\pm$ 0.04 &  0.541 $\pm$ 0.04 &  0.524 $\pm$ 0.03 &  0.520 $\pm$ 0.03 \\
 & MI-RMSE &  0.095 $\pm$ 0.02 &  0.085 $\pm$ 0.02 &  0.076 $\pm$ 0.02 &  0.067 $\pm$ 0.02 &  0.059 $\pm$ 0.02 &  0.051 $\pm$ 0.01 &  0.045 $\pm$ 0.01 \\
\midrule
\multirow{2}{*}{Kriging NS} & RMSE &  \textbf{0.638 $\pm$ 0.07} &  \textbf{0.610 $\pm$ 0.06} &  \textbf{0.585 $\pm$ 0.05} &  \textbf{0.551 $\pm$ 0.04} &  \textbf{0.532 $\pm$ 0.04} &  \textbf{0.513 $\pm$ 0.03} &  \textbf{0.507 $\pm$ 0.03} \\
 & MI-RMSE &  0.092 $\pm$ 0.02 &  0.081 $\pm$ 0.02 &  0.073 $\pm$ 0.02 &  0.065 $\pm$ 0.02 &  0.057 $\pm$ 0.02 &  0.049 $\pm$ 0.01 &  0.043 $\pm$ 0.01 \\
\midrule
\multirow{2}{*}{ML Base} & RMSE &  1.352 $\pm$ 0.21 &  1.298 $\pm$ 0.22 &  1.299 $\pm$ 0.25 &  1.218 $\pm$ 0.20 &  1.184 $\pm$ 0.21 &  1.127 $\pm$ 0.21 &  1.135 $\pm$ 0.22 \\
 & MI-RMSE &  0.434 $\pm$ 0.16 &  0.382 $\pm$ 0.17 &  0.330 $\pm$ 0.15 &  0.280 $\pm$ 0.15 &  0.255 $\pm$ 0.14 &  0.204 $\pm$ 0.12 &  0.189 $\pm$ 0.11 \\
\midrule
\multirow{2}{*}{ML VSL} & RMSE &  0.667 $\pm$ 0.07 &  0.672 $\pm$ 0.07 &  0.647 $\pm$ 0.07 &  0.591 $\pm$ 0.04 &  0.580 $\pm$ 0.04 &  0.569 $\pm$ 0.05 &  0.547 $\pm$ 0.04 \\
 & MI-RMSE &  \textbf{0.087 $\pm$ 0.03} &  \textbf{0.075 $\pm$ 0.02} &  \textbf{0.067 $\pm$ 0.02} &  \textbf{0.058 $\pm$ 0.02} &  \textbf{0.051 $\pm$ 0.02} &  \textbf{0.042 $\pm$ 0.01} &  \textbf{0.035 $\pm$ 0.01} \\
\bottomrule
\end{tabular}
\end{table}

\begin{table}[tb]
\centering
\caption{Non-stationary \textit{Matérn} --- Comparison of model performance across varying percentages of known points (20\%, 30\%, 40\%, 50\%, 60\%, 70\% and 80\%). Values represent the mean ($\pm$ standard deviation) computed over 100 independent runs. The spatial interpolation was conducted using a non-stationary \textit{Matérn} covariogram with a range parameter of 70\% of the grid size. The underlying data follows a \textit{Gaussian} distribution.}
\label{tab:complete_ns-aniso-tilt-mean_70_phi}
\begin{tabular}{llccccccc}
\toprule
\textbf{Model} & \textbf{Metric}  & \textbf{20\%} & \textbf{30\%} & \textbf{40\%} & \textbf{50\%} & \textbf{60\%} & \textbf{70\%} & \textbf{80\%}  \\
\midrule
\multirow{2}{*}{Kriging} & RMSE &  0.598 $\pm$ 0.06 &  0.575 $\pm$ 0.05 &  0.552 $\pm$ 0.05 &  0.518 $\pm$ 0.04 &  0.501 $\pm$ 0.03 &  0.486 $\pm$ 0.02 &  0.482 $\pm$ 0.03 \\
 & MI-RMSE &  0.089 $\pm$ 0.02 &  0.080 $\pm$ 0.02 &  0.071 $\pm$ 0.02 &  0.063 $\pm$ 0.02 &  0.055 $\pm$ 0.02 &  0.048 $\pm$ 0.01 &  0.041 $\pm$ 0.01 \\
\midrule
\multirow{2}{*}{Kriging NS} & RMSE &  \textbf{0.591 $\pm$ 0.06} &  \textbf{0.567 $\pm$ 0.05} &  \textbf{0.542 $\pm$ 0.05} &  \textbf{0.510 $\pm$ 0.03} &  \textbf{0.493 $\pm$ 0.03} &  \textbf{0.475 $\pm$ 0.02} &  \textbf{0.470 $\pm$ 0.02} \\
 & MI-RMSE &  0.086 $\pm$ 0.02 &  0.077 $\pm$ 0.02 &  0.068 $\pm$ 0.02 &  0.060 $\pm$ 0.02 &  0.053 $\pm$ 0.02 &  0.045 $\pm$ 0.01 &  0.039 $\pm$ 0.01 \\
\midrule
\multirow{2}{*}{ML Base} & RMSE &  1.280 $\pm$ 0.20 &  1.215 $\pm$ 0.19 &  1.199 $\pm$ 0.23 &  1.154 $\pm$ 0.21 &  1.101 $\pm$ 0.20 &  1.070 $\pm$ 0.22 &  1.068 $\pm$ 0.23 \\
 & MI-RMSE &  0.416 $\pm$ 0.17 &  0.351 $\pm$ 0.18 &  0.294 $\pm$ 0.17 &  0.268 $\pm$ 0.15 &  0.232 $\pm$ 0.14 &  0.199 $\pm$ 0.12 &  0.177 $\pm$ 0.11 \\
\midrule
\multirow{2}{*}{ML VSL} & RMSE &  0.623 $\pm$ 0.06 &  0.633 $\pm$ 0.07 &  0.610 $\pm$ 0.07 &  0.552 $\pm$ 0.04 &  0.543 $\pm$ 0.04 &  0.534 $\pm$ 0.05 &  0.513 $\pm$ 0.03 \\
 & MI-RMSE &  \textbf{0.081 $\pm$ 0.02} &  \textbf{0.070 $\pm$ 0.02} &  \textbf{0.062 $\pm$ 0.02} &  \textbf{0.053 $\pm$ 0.02} &  \textbf{0.047 $\pm$ 0.02} &  \textbf{0.039 $\pm$ 0.01} &  \textbf{0.032 $\pm$ 0.01} \\
\bottomrule
\end{tabular}
\end{table}

\begin{table}[tb]
\centering
\caption{Non-stationary \textit{Matérn} --- Comparison of model performance across varying percentages of known points (20\%, 30\%, 40\%, 50\%, 60\%, 70\% and 80\%). Values represent the mean ($\pm$ standard deviation) computed over 100 independent runs. The spatial interpolation was conducted using a non-stationary \textit{Matérn} covariogram with a range parameter of 80\% of the grid size. The underlying data follows a \textit{Gaussian} distribution.}
\label{tab:complete_ns-aniso-tilt-mean_80_phi}
\begin{tabular}{llccccccc}
\toprule
\textbf{Model} & \textbf{Metric}  & \textbf{20\%} & \textbf{30\%} & \textbf{40\%} & \textbf{50\%} & \textbf{60\%} & \textbf{70\%} & \textbf{80\%}  \\
\midrule
\multirow{2}{*}{Kriging} & RMSE &  \textbf{0.558 $\pm$ 0.06} &  0.539 $\pm$ 0.05 &  0.517 $\pm$ 0.04 &  0.486 $\pm$ 0.03 &  0.469 $\pm$ 0.03 &  0.454 $\pm$ 0.02 &  0.452 $\pm$ 0.02 \\
 & MI-RMSE &  0.084 $\pm$ 0.02 &  0.075 $\pm$ 0.02 &  0.067 $\pm$ 0.02 &  0.059 $\pm$ 0.02 &  0.052 $\pm$ 0.01 &  0.045 $\pm$ 0.01 &  0.040 $\pm$ 0.01 \\
\midrule
\multirow{2}{*}{Kriging NS} & RMSE &  0.560 $\pm$ 0.09 &  \textbf{0.532 $\pm$ 0.05} &  \textbf{0.508 $\pm$ 0.04} &  \textbf{0.478 $\pm$ 0.03} &  \textbf{0.461 $\pm$ 0.03} &  \textbf{0.444 $\pm$ 0.02} &  \textbf{0.441 $\pm$ 0.02} \\
 & MI-RMSE &  0.085 $\pm$ 0.04 &  0.072 $\pm$ 0.02 &  0.064 $\pm$ 0.02 &  0.057 $\pm$ 0.02 &  0.050 $\pm$ 0.01 &  0.043 $\pm$ 0.01 &  0.038 $\pm$ 0.01 \\
\midrule
\multirow{2}{*}{ML Base} & RMSE &  1.228 $\pm$ 0.21 &  1.166 $\pm$ 0.20 &  1.144 $\pm$ 0.22 &  1.099 $\pm$ 0.21 &  1.032 $\pm$ 0.20 &  1.021 $\pm$ 0.23 &  1.011 $\pm$ 0.23 \\
 & MI-RMSE &  0.381 $\pm$ 0.16 &  0.343 $\pm$ 0.18 &  0.279 $\pm$ 0.17 &  0.254 $\pm$ 0.15 &  0.215 $\pm$ 0.14 &  0.198 $\pm$ 0.13 &  0.173 $\pm$ 0.11 \\
\midrule
\multirow{2}{*}{ML VSL} & RMSE &  0.586 $\pm$ 0.06 &  0.602 $\pm$ 0.07 &  0.580 $\pm$ 0.06 &  0.522 $\pm$ 0.04 &  0.514 $\pm$ 0.04 &  0.506 $\pm$ 0.05 &  0.484 $\pm$ 0.03 \\
 & MI-RMSE &  \textbf{0.075 $\pm$ 0.02} &  \textbf{0.065 $\pm$ 0.02} &  \textbf{0.058 $\pm$ 0.02} &  \textbf{0.050 $\pm$ 0.02} &  \textbf{0.044 $\pm$ 0.01} &  \textbf{0.036 $\pm$ 0.01} &  \textbf{0.030 $\pm$ 0.01} \\
\bottomrule
\end{tabular}
\end{table}

\end{landscape}
\restoregeometry

\subsection{Composite Fields}
\label{ap:subsec:composite_fields}

\begin{figure}[t]
\centering
\includegraphics[width=\textwidth]{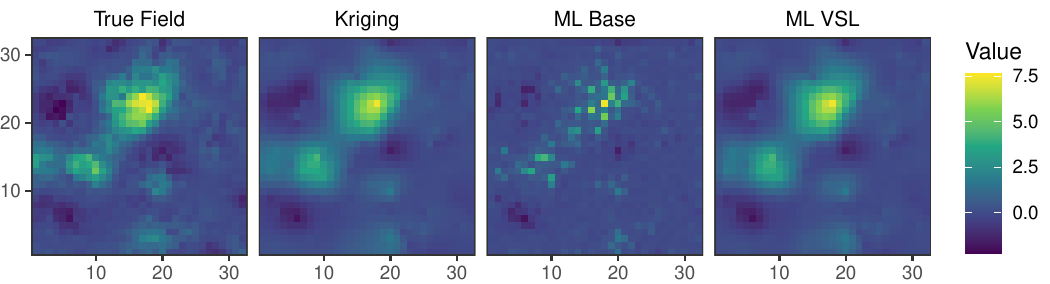}
\caption{\textit{Exponential$\times$Wave} --- Visualisation of spatial interpolation methods for a representative realisation of an isotropic Gaussian random field with \textit{Exponential$\:\times$Wave} covariance. The panels show the True field alongside predictions from Ordinary Kriging, ML Base, and the proposed ML VSL method.}
\label{fig:exponential-times-wave-20-phi-20-run_true_ok_krige-uns_ml-001_ml-004_016_plot}
\end{figure}

This section presents the complete quantitative results for the composite field setting described in \Cref{subsec:composite_fields}.
Spatial fields are constructed by combining a smooth Gaussian random field with an Exponential covariance function (\ref{eq:cov_exp}) and an oscillatory component governed by a Wave covariance function (\ref{eq:cov_wave}), producing fields with qualitatively distinct spatial regimes and sharp transitions (\Cref{fig:exponential-times-wave-20-phi-20-run_true_ok_krige-uns_ml-001_ml-004_016_plot}).
As noted in the main text, such fields are poorly described by a single global covariance model, placing classical Kriging at a structural disadvantage.
\Crefrange{tab:complete_exponential-times-wave_10_phi}{tab:complete_exponential-times-wave_80_phi} report the full results across all considered values of $\phi$.

The composite field setting produces a distinct pattern of results relative to the preceding experiments.
Ordinary Kriging attains the lowest overall RMSE across the majority of $\phi$ values and sparsity levels, with ML VSL competitive primarily at 20\% observed locations, where it matches or slightly outperforms Kriging.
At higher observation densities, Kriging consistently achieves lower RMSE, reflecting its ability to exploit the smooth Exponential component of the field when sufficient observations are available.
Nevertheless, the margin between Kriging and ML VSL remains modest throughout, and both methods substantially outperform ML Base, whose errors are roughly twice as large across all settings.

The MI-RMSE metric, however, tells a more nuanced story, as ML VSL achieves lower MI-RMSE than Kriging across almost all settings, indicating superior reconstruction of the unobserved regions.
This is consistent with the structural limitation of Kriging in this setting, where a single globally fitted covariance model tends to over-smooth oscillatory regions or introduce artefacts at the transition between the smooth and periodic components, whereas the partial convolutional architecture better preserves both the smooth background structure and the localised periodic patterns simultaneously.
ML Base again exhibits high variance and little sensitivity to observation density, in contrast to both Kriging and ML VSL, which show consistent error reduction as the proportion of observed locations increases.

\begin{landscape}
\begin{table}[tb]
\centering
\caption{\textit{Exponential$\times$Wave} --- Comparison of model performance across varying percentages of known points (20\%, 30\%, 40\%, 50\%, 60\%, 70\% and 80\%). Values represent the mean ($\pm$ standard deviation) computed over 100 independent runs. The spatial interpolation was conducted using an \textit{Exponential$\:\times$Wave} covariogram with a range parameter of 10\% of the grid size. The underlying data follows a \textit{Gaussian} distribution.}
\label{tab:complete_exponential-times-wave_10_phi}
\begin{tabular}{llccccccc}
\toprule
\textbf{Model} & \textbf{Metric}  & \textbf{20\%} & \textbf{30\%} & \textbf{40\%} & \textbf{50\%} & \textbf{60\%} & \textbf{70\%} & \textbf{80\%}  \\
\midrule
\multirow{2}{*}{Kriging} & RMSE &  0.734 $\pm$ 0.17 &  \textbf{0.682 $\pm$ 0.14} &  \textbf{0.666 $\pm$ 0.14} &  0.621 $\pm$ 0.11 &  \textbf{0.592 $\pm$ 0.09} &  \textbf{0.582 $\pm$ 0.09} &  \textbf{0.569 $\pm$ 0.08} \\
 & MI-RMSE &  \textbf{0.177 $\pm$ 0.05} &  \textbf{0.159 $\pm$ 0.04} &  0.142 $\pm$ 0.03 &  0.126 $\pm$ 0.03 &  0.109 $\pm$ 0.03 &  0.095 $\pm$ 0.02 &  0.081 $\pm$ 0.02 \\
\midrule
\multirow{2}{*}{ML Base} & RMSE &  1.037 $\pm$ 0.31 &  1.010 $\pm$ 0.26 &  1.029 $\pm$ 0.25 &  1.007 $\pm$ 0.24 &  1.000 $\pm$ 0.22 &  1.004 $\pm$ 0.25 &  0.989 $\pm$ 0.21 \\
 & MI-RMSE &  0.490 $\pm$ 0.11 &  0.464 $\pm$ 0.08 &  0.417 $\pm$ 0.07 &  0.371 $\pm$ 0.07 &  0.350 $\pm$ 0.06 &  0.300 $\pm$ 0.06 &  0.252 $\pm$ 0.05 \\
\midrule
\multirow{2}{*}{ML VSL} & RMSE &  \textbf{0.729 $\pm$ 0.15} &  0.687 $\pm$ 0.14 &  0.674 $\pm$ 0.14 &  \textbf{0.621 $\pm$ 0.11} &  0.596 $\pm$ 0.09 &  0.588 $\pm$ 0.10 &  0.575 $\pm$ 0.08 \\
 & MI-RMSE &  0.189 $\pm$ 0.05 &  0.160 $\pm$ 0.04 &  \textbf{0.138 $\pm$ 0.03} &  \textbf{0.117 $\pm$ 0.03} &  \textbf{0.099 $\pm$ 0.03} &  \textbf{0.084 $\pm$ 0.02} &  \textbf{0.070 $\pm$ 0.02} \\
\bottomrule
\end{tabular}
\end{table}

\begin{table}[tb]
\centering
\caption{\textit{Exponential$\times$Wave} --- Comparison of model performance across varying percentages of known points (20\%, 30\%, 40\%, 50\%, 60\%, 70\% and 80\%). Values represent the mean ($\pm$ standard deviation) computed over 100 independent runs. The spatial interpolation was conducted using an \textit{Exponential$\:\times$Wave} covariogram with a range parameter of 20\% of the grid size. The underlying data follows a \textit{Gaussian} distribution.}
\label{tab:complete_exponential-times-wave_20_phi}
\begin{tabular}{llccccccc}
\toprule
\textbf{Model} & \textbf{Metric}  & \textbf{20\%} & \textbf{30\%} & \textbf{40\%} & \textbf{50\%} & \textbf{60\%} & \textbf{70\%} & \textbf{80\%}  \\
\midrule
\multirow{2}{*}{Kriging} & RMSE &  0.562 $\pm$ 0.13 &  \textbf{0.525 $\pm$ 0.11} &  \textbf{0.503 $\pm$ 0.11} &  \textbf{0.458 $\pm$ 0.08} &  \textbf{0.434 $\pm$ 0.07} &  \textbf{0.426 $\pm$ 0.07} &  \textbf{0.414 $\pm$ 0.06} \\
 & MI-RMSE &  0.110 $\pm$ 0.04 &  0.098 $\pm$ 0.03 &  0.087 $\pm$ 0.03 &  0.077 $\pm$ 0.03 &  0.067 $\pm$ 0.02 &  0.058 $\pm$ 0.02 &  0.050 $\pm$ 0.02 \\
\midrule
\multirow{2}{*}{ML Base} & RMSE &  1.035 $\pm$ 0.36 &  1.006 $\pm$ 0.30 &  1.002 $\pm$ 0.27 &  0.960 $\pm$ 0.25 &  0.942 $\pm$ 0.22 &  0.938 $\pm$ 0.25 &  0.918 $\pm$ 0.20 \\
 & MI-RMSE &  0.531 $\pm$ 0.10 &  0.491 $\pm$ 0.11 &  0.422 $\pm$ 0.11 &  0.377 $\pm$ 0.10 &  0.340 $\pm$ 0.10 &  0.288 $\pm$ 0.10 &  0.245 $\pm$ 0.08 \\
\midrule
\multirow{2}{*}{ML VSL} & RMSE &  \textbf{0.556 $\pm$ 0.12} &  0.538 $\pm$ 0.11 &  0.519 $\pm$ 0.12 &  0.465 $\pm$ 0.09 &  0.448 $\pm$ 0.07 &  0.442 $\pm$ 0.08 &  0.428 $\pm$ 0.06 \\
 & MI-RMSE &  \textbf{0.105 $\pm$ 0.04} &  \textbf{0.090 $\pm$ 0.03} &  \textbf{0.079 $\pm$ 0.03} &  \textbf{0.068 $\pm$ 0.02} &  \textbf{0.058 $\pm$ 0.02} &  \textbf{0.049 $\pm$ 0.02} &  \textbf{0.042 $\pm$ 0.02} \\
\bottomrule
\end{tabular}
\end{table}

\begin{table}[tb]
\centering
\caption{\textit{Exponential$\times$Wave} --- Comparison of model performance across varying percentages of known points (20\%, 30\%, 40\%, 50\%, 60\%, 70\% and 80\%). Values represent the mean ($\pm$ standard deviation) computed over 100 independent runs. The spatial interpolation was conducted using an \textit{Exponential$\:\times$Wave} covariogram with a range parameter of 30\% of the grid size. The underlying data follows a \textit{Gaussian} distribution.}
\label{tab:complete_exponential-times-wave_30_phi}
\begin{tabular}{llccccccc}
\toprule
\textbf{Model} & \textbf{Metric}  & \textbf{20\%} & \textbf{30\%} & \textbf{40\%} & \textbf{50\%} & \textbf{60\%} & \textbf{70\%} & \textbf{80\%}  \\
\midrule
\multirow{2}{*}{Kriging} & RMSE &  0.479 $\pm$ 0.11 &  \textbf{0.448 $\pm$ 0.10} &  \textbf{0.424 $\pm$ 0.09} &  \textbf{0.381 $\pm$ 0.07} &  \textbf{0.360 $\pm$ 0.06} &  \textbf{0.354 $\pm$ 0.06} &  \textbf{0.342 $\pm$ 0.05} \\
 & MI-RMSE &  0.083 $\pm$ 0.03 &  0.074 $\pm$ 0.03 &  0.066 $\pm$ 0.03 &  0.058 $\pm$ 0.02 &  0.051 $\pm$ 0.02 &  0.044 $\pm$ 0.02 &  0.038 $\pm$ 0.02 \\
\midrule
\multirow{2}{*}{ML Base} & RMSE &  1.013 $\pm$ 0.38 &  0.973 $\pm$ 0.30 &  0.956 $\pm$ 0.27 &  0.890 $\pm$ 0.24 &  0.879 $\pm$ 0.23 &  0.864 $\pm$ 0.24 &  0.831 $\pm$ 0.19 \\
 & MI-RMSE &  0.536 $\pm$ 0.10 &  0.488 $\pm$ 0.12 &  0.408 $\pm$ 0.12 &  0.349 $\pm$ 0.12 &  0.323 $\pm$ 0.11 &  0.268 $\pm$ 0.11 &  0.218 $\pm$ 0.09 \\
\midrule
\multirow{2}{*}{ML VSL} & RMSE &  \textbf{0.472 $\pm$ 0.10} &  0.469 $\pm$ 0.11 &  0.447 $\pm$ 0.11 &  0.394 $\pm$ 0.08 &  0.381 $\pm$ 0.07 &  0.377 $\pm$ 0.08 &  0.361 $\pm$ 0.06 \\
 & MI-RMSE &  \textbf{0.076 $\pm$ 0.04} &  \textbf{0.066 $\pm$ 0.03} &  \textbf{0.058 $\pm$ 0.03} &  \textbf{0.049 $\pm$ 0.02} &  \textbf{0.042 $\pm$ 0.02} &  \textbf{0.036 $\pm$ 0.02} &  \textbf{0.031 $\pm$ 0.01} \\
\bottomrule
\end{tabular}
\end{table}

\begin{table}[tb]
\centering
\caption{\textit{Exponential$\times$Wave} --- Comparison of model performance across varying percentages of known points (20\%, 30\%, 40\%, 50\%, 60\%, 70\% and 80\%). Values represent the mean ($\pm$ standard deviation) computed over 100 independent runs. The spatial interpolation was conducted using an \textit{Exponential$\:\times$Wave} covariogram with a range parameter of 40\% of the grid size. The underlying data follows a \textit{Gaussian} distribution.}
\label{tab:complete_exponential-times-wave_40_phi}
\begin{tabular}{llccccccc}
\toprule
\textbf{Model} & \textbf{Metric}  & \textbf{20\%} & \textbf{30\%} & \textbf{40\%} & \textbf{50\%} & \textbf{60\%} & \textbf{70\%} & \textbf{80\%}  \\
\midrule
\multirow{2}{*}{Kriging} & RMSE &  0.428 $\pm$ 0.10 &  \textbf{0.401 $\pm$ 0.09} &  \textbf{0.376 $\pm$ 0.09} &  \textbf{0.334 $\pm$ 0.06} &  \textbf{0.316 $\pm$ 0.05} &  \textbf{0.311 $\pm$ 0.05} &  \textbf{0.299 $\pm$ 0.04} \\
 & MI-RMSE &  0.069 $\pm$ 0.03 &  0.061 $\pm$ 0.03 &  0.055 $\pm$ 0.03 &  0.049 $\pm$ 0.02 &  0.042 $\pm$ 0.02 &  0.037 $\pm$ 0.02 &  0.032 $\pm$ 0.02 \\
\midrule
\multirow{2}{*}{ML Base} & RMSE &  0.989 $\pm$ 0.39 &  0.944 $\pm$ 0.31 &  0.912 $\pm$ 0.28 &  0.839 $\pm$ 0.24 &  0.810 $\pm$ 0.22 &  0.790 $\pm$ 0.23 &  0.760 $\pm$ 0.18 \\
 & MI-RMSE &  0.541 $\pm$ 0.12 &  0.479 $\pm$ 0.13 &  0.384 $\pm$ 0.13 &  0.324 $\pm$ 0.13 &  0.290 $\pm$ 0.13 &  0.241 $\pm$ 0.11 &  0.196 $\pm$ 0.09 \\
\midrule
\multirow{2}{*}{ML VSL} & RMSE &  \textbf{0.421 $\pm$ 0.10} &  0.426 $\pm$ 0.11 &  0.403 $\pm$ 0.11 &  0.351 $\pm$ 0.07 &  0.341 $\pm$ 0.07 &  0.338 $\pm$ 0.08 &  0.321 $\pm$ 0.06 \\
 & MI-RMSE &  \textbf{0.061 $\pm$ 0.03} &  \textbf{0.053 $\pm$ 0.03} &  \textbf{0.047 $\pm$ 0.03} &  \textbf{0.040 $\pm$ 0.02} &  \textbf{0.034 $\pm$ 0.02} &  \textbf{0.030 $\pm$ 0.02} &  \textbf{0.025 $\pm$ 0.01} \\
\bottomrule
\end{tabular}
\end{table}

\begin{table}[tb]
\centering
\caption{\textit{Exponential$\times$Wave} --- Comparison of model performance across varying percentages of known points (20\%, 30\%, 40\%, 50\%, 60\%, 70\% and 80\%). Values represent the mean ($\pm$ standard deviation) computed over 100 independent runs. The spatial interpolation was conducted using an \textit{Exponential$\:\times$Wave} covariogram with a range parameter of 50\% of the grid size. The underlying data follows a \textit{Gaussian} distribution.}
\label{tab:complete_exponential-times-wave_50_phi}
\begin{tabular}{llccccccc}
\toprule
\textbf{Model} & \textbf{Metric}  & \textbf{20\%} & \textbf{30\%} & \textbf{40\%} & \textbf{50\%} & \textbf{60\%} & \textbf{70\%} & \textbf{80\%}  \\
\midrule
\multirow{2}{*}{Kriging} & RMSE &  0.394 $\pm$ 0.09 &  \textbf{0.369 $\pm$ 0.09} &  \textbf{0.343 $\pm$ 0.08} &  \textbf{0.302 $\pm$ 0.06} &  \textbf{0.285 $\pm$ 0.05} &  \textbf{0.282 $\pm$ 0.05} &  \textbf{0.269 $\pm$ 0.04} \\
 & MI-RMSE &  0.060 $\pm$ 0.03 &  0.054 $\pm$ 0.03 &  0.048 $\pm$ 0.03 &  0.042 $\pm$ 0.02 &  0.037 $\pm$ 0.02 &  0.032 $\pm$ 0.02 &  0.028 $\pm$ 0.02 \\
\midrule
\multirow{2}{*}{ML Base} & RMSE &  0.973 $\pm$ 0.40 &  0.923 $\pm$ 0.33 &  0.870 $\pm$ 0.28 &  0.797 $\pm$ 0.24 &  0.753 $\pm$ 0.20 &  0.740 $\pm$ 0.22 &  0.693 $\pm$ 0.17 \\
 & MI-RMSE &  0.550 $\pm$ 0.10 &  0.472 $\pm$ 0.14 &  0.371 $\pm$ 0.14 &  0.312 $\pm$ 0.13 &  0.269 $\pm$ 0.13 &  0.221 $\pm$ 0.11 &  0.176 $\pm$ 0.10 \\
\midrule
\multirow{2}{*}{ML VSL} & RMSE &  \textbf{0.386 $\pm$ 0.09} &  0.397 $\pm$ 0.11 &  0.373 $\pm$ 0.11 &  0.322 $\pm$ 0.07 &  0.313 $\pm$ 0.07 &  0.311 $\pm$ 0.08 &  0.294 $\pm$ 0.05 \\
 & MI-RMSE &  \textbf{0.052 $\pm$ 0.03} &  \textbf{0.046 $\pm$ 0.03} &  \textbf{0.040 $\pm$ 0.02} &  \textbf{0.034 $\pm$ 0.02} &  \textbf{0.029 $\pm$ 0.02} &  \textbf{0.026 $\pm$ 0.02} &  \textbf{0.021 $\pm$ 0.01} \\
\bottomrule
\end{tabular}
\end{table}

\begin{table}[tb]
\centering
\caption{\textit{Exponential$\times$Wave} --- Comparison of model performance across varying percentages of known points (20\%, 30\%, 40\%, 50\%, 60\%, 70\% and 80\%). Values represent the mean ($\pm$ standard deviation) computed over 100 independent runs. The spatial interpolation was conducted using an \textit{Exponential$\:\times$Wave} covariogram with a range parameter of 60\% of the grid size. The underlying data follows a \textit{Gaussian} distribution.}
\label{tab:complete_exponential-times-wave_60_phi}
\begin{tabular}{llccccccc}
\toprule
\textbf{Model} & \textbf{Metric}  & \textbf{20\%} & \textbf{30\%} & \textbf{40\%} & \textbf{50\%} & \textbf{60\%} & \textbf{70\%} & \textbf{80\%}  \\
\midrule
\multirow{2}{*}{Kriging} & RMSE &  0.368 $\pm$ 0.09 &  \textbf{0.345 $\pm$ 0.08} &  \textbf{0.319 $\pm$ 0.08} &  \textbf{0.279 $\pm$ 0.05} &  \textbf{0.263 $\pm$ 0.04} &  \textbf{0.260 $\pm$ 0.05} &  \textbf{0.247 $\pm$ 0.04} \\
 & MI-RMSE &  0.054 $\pm$ 0.03 &  0.048 $\pm$ 0.03 &  0.044 $\pm$ 0.02 &  0.038 $\pm$ 0.02 &  0.033 $\pm$ 0.02 &  0.029 $\pm$ 0.02 &  0.025 $\pm$ 0.01 \\
\midrule
\multirow{2}{*}{ML Base} & RMSE &  0.954 $\pm$ 0.41 &  0.908 $\pm$ 0.34 &  0.844 $\pm$ 0.27 &  0.756 $\pm$ 0.23 &  0.706 $\pm$ 0.19 &  0.674 $\pm$ 0.21 &  0.642 $\pm$ 0.15 \\
 & MI-RMSE &  0.545 $\pm$ 0.11 &  0.469 $\pm$ 0.13 &  0.365 $\pm$ 0.14 &  0.291 $\pm$ 0.13 &  0.251 $\pm$ 0.13 &  0.191 $\pm$ 0.11 &  0.158 $\pm$ 0.10 \\
\midrule
\multirow{2}{*}{ML VSL} & RMSE &  \textbf{0.360 $\pm$ 0.09} &  0.375 $\pm$ 0.11 &  0.351 $\pm$ 0.10 &  0.300 $\pm$ 0.07 &  0.293 $\pm$ 0.07 &  0.292 $\pm$ 0.08 &  0.273 $\pm$ 0.05 \\
 & MI-RMSE &  \textbf{0.046 $\pm$ 0.03} &  \textbf{0.040 $\pm$ 0.03} &  \textbf{0.036 $\pm$ 0.02} &  \textbf{0.030 $\pm$ 0.02} &  \textbf{0.026 $\pm$ 0.02} &  \textbf{0.023 $\pm$ 0.01} &  \textbf{0.019 $\pm$ 0.01} \\
\bottomrule
\end{tabular}
\end{table}

\begin{table}[tb]
\centering
\caption{\textit{Exponential$\times$Wave} --- Comparison of model performance across varying percentages of known points (20\%, 30\%, 40\%, 50\%, 60\%, 70\% and 80\%). Values represent the mean ($\pm$ standard deviation) computed over 100 independent runs. The spatial interpolation was conducted using an \textit{Exponential$\:\times$Wave} covariogram with a range parameter of 70\% of the grid size. The underlying data follows a \textit{Gaussian} distribution.}
\label{tab:complete_exponential-times-wave_70_phi}
\begin{tabular}{llccccccc}
\toprule
\textbf{Model} & \textbf{Metric}  & \textbf{20\%} & \textbf{30\%} & \textbf{40\%} & \textbf{50\%} & \textbf{60\%} & \textbf{70\%} & \textbf{80\%}  \\
\midrule
\multirow{2}{*}{Kriging} & RMSE &  0.347 $\pm$ 0.08 &  \textbf{0.326 $\pm$ 0.08} &  \textbf{0.300 $\pm$ 0.07} &  \textbf{0.260 $\pm$ 0.05} &  \textbf{0.245 $\pm$ 0.04} &  \textbf{0.243 $\pm$ 0.04} &  \textbf{0.230 $\pm$ 0.03} \\
 & MI-RMSE &  0.050 $\pm$ 0.03 &  0.044 $\pm$ 0.03 &  0.040 $\pm$ 0.02 &  0.035 $\pm$ 0.02 &  0.030 $\pm$ 0.02 &  0.027 $\pm$ 0.02 &  0.023 $\pm$ 0.01 \\
\midrule
\multirow{2}{*}{ML Base} & RMSE &  0.939 $\pm$ 0.42 &  0.887 $\pm$ 0.34 &  0.820 $\pm$ 0.28 &  0.734 $\pm$ 0.23 &  0.678 $\pm$ 0.18 &  0.645 $\pm$ 0.20 &  0.599 $\pm$ 0.16 \\
 & MI-RMSE &  0.545 $\pm$ 0.10 &  0.476 $\pm$ 0.13 &  0.356 $\pm$ 0.14 &  0.289 $\pm$ 0.13 &  0.242 $\pm$ 0.12 &  0.181 $\pm$ 0.11 &  0.145 $\pm$ 0.10 \\
\midrule
\multirow{2}{*}{ML VSL} & RMSE &  \textbf{0.339 $\pm$ 0.08} &  0.358 $\pm$ 0.11 &  0.334 $\pm$ 0.10 &  0.283 $\pm$ 0.06 &  0.278 $\pm$ 0.07 &  0.277 $\pm$ 0.08 &  0.258 $\pm$ 0.05 \\
 & MI-RMSE &  \textbf{0.042 $\pm$ 0.03} &  \textbf{0.036 $\pm$ 0.02} &  \textbf{0.032 $\pm$ 0.02} &  \textbf{0.027 $\pm$ 0.02} &  \textbf{0.024 $\pm$ 0.02} &  \textbf{0.021 $\pm$ 0.01} &  \textbf{0.017 $\pm$ 0.01} \\
\bottomrule
\end{tabular}
\end{table}

\begin{table}[tb]
\centering
\caption{\textit{Exponential$\times$Wave} --- Comparison of model performance across varying percentages of known points (20\%, 30\%, 40\%, 50\%, 60\%, 70\% and 80\%). Values represent the mean ($\pm$ standard deviation) computed over 100 independent runs. The spatial interpolation was conducted using an \textit{Exponential$\:\times$Wave} covariogram with a range parameter of 80\% of the grid size. The underlying data follows a \textit{Gaussian} distribution.}
\label{tab:complete_exponential-times-wave_80_phi}
\begin{tabular}{llccccccc}
\toprule
\textbf{Model} & \textbf{Metric}  & \textbf{20\%} & \textbf{30\%} & \textbf{40\%} & \textbf{50\%} & \textbf{60\%} & \textbf{70\%} & \textbf{80\%}  \\
\midrule
\multirow{2}{*}{Kriging} & RMSE &  0.331 $\pm$ 0.08 &  \textbf{0.310 $\pm$ 0.08} &  \textbf{0.285 $\pm$ 0.07} &  \textbf{0.246 $\pm$ 0.05} &  \textbf{0.231 $\pm$ 0.04} &  \textbf{0.230 $\pm$ 0.04} &  \textbf{0.217 $\pm$ 0.03} \\
 & MI-RMSE &  0.046 $\pm$ 0.03 &  0.041 $\pm$ 0.02 &  0.037 $\pm$ 0.02 &  0.032 $\pm$ 0.02 &  0.028 $\pm$ 0.02 &  0.025 $\pm$ 0.02 &  0.021 $\pm$ 0.01 \\
\midrule
\multirow{2}{*}{ML Base} & RMSE &  0.925 $\pm$ 0.43 &  0.873 $\pm$ 0.34 &  0.783 $\pm$ 0.28 &  0.719 $\pm$ 0.24 &  0.634 $\pm$ 0.19 &  0.609 $\pm$ 0.19 &  0.566 $\pm$ 0.15 \\
 & MI-RMSE &  0.549 $\pm$ 0.11 &  0.465 $\pm$ 0.13 &  0.331 $\pm$ 0.15 &  0.280 $\pm$ 0.13 &  0.221 $\pm$ 0.13 &  0.165 $\pm$ 0.11 &  0.138 $\pm$ 0.10 \\
\midrule
\multirow{2}{*}{ML VSL} & RMSE &  \textbf{0.323 $\pm$ 0.08} &  0.344 $\pm$ 0.11 &  0.320 $\pm$ 0.10 &  0.270 $\pm$ 0.06 &  0.265 $\pm$ 0.07 &  0.265 $\pm$ 0.08 &  0.245 $\pm$ 0.05 \\
 & MI-RMSE &  \textbf{0.038 $\pm$ 0.03} &  \textbf{0.033 $\pm$ 0.02} &  \textbf{0.030 $\pm$ 0.02} &  \textbf{0.025 $\pm$ 0.02} &  \textbf{0.022 $\pm$ 0.02} &  \textbf{0.019 $\pm$ 0.01} &  \textbf{0.016 $\pm$ 0.01} \\
\bottomrule
\end{tabular}
\end{table}

\end{landscape}

\subsection{Conclusion}
\label{ap:subsec:num_conclusion}

The numerical experiments presented in this appendix provide a comprehensive characterisation of the proposed visual spatial learning method across a range of synthetic spatial settings, from the idealized stationary isotropic case to increasingly challenging non-stationary, anisotropic, and structurally composite fields.
Taken together, the results support the view that the proposed method provides a robust and flexible alternative to Kriging, as it achieves near-optimal performance under conditions that strictly favour classical geostatistical approaches, while attaining superior or competitive performance in more complex settings where global covariance model misspecification, nugget contamination, or structural heterogeneity limits the effectiveness of linear predictors.
Across all experiments, ML Base consistently underperforms, underscoring that the gains of ML VSL are attributable to the specific architectural and training design choices rather than to the use of a learning-based approach alone.

\section*{Code Availability}
The code for reproducing our experiments is publicly available at \url{https://github.com/dbtnc/visual_spatial_learning}.

\end{document}